\documentclass[9.5pt,journal,compsoc]{IEEEtran}
\usepackage[hidelinks,breaklinks=true,bookmarks=false,linkcolor={black},citecolor={black},urlcolor={black}]{hyperref}
\usepackage{xcolor}
\usepackage{graphicx}
\usepackage[table]{colortbl}
\usepackage{enumitem}
\usepackage{amsmath}
\usepackage{amssymb}
\usepackage{multirow}
\usepackage{amsthm}
\usepackage{mathrsfs}
\usepackage{booktabs}
\usepackage{bbding}
\usepackage{bbm}
\usepackage{subfigure}
\usepackage{ragged2e}
\usepackage{tabularx} 
\usepackage{hyperref}
\usepackage{svg}
\svgpath{{fig/}} 
\usepackage{amsmath}
\usepackage{url}
\usepackage{makecell}
\renewcommand{\paragraph}[1]{\vspace{2mm}\noindent \textbf{#1}}
\newcommand{\myeg}{\textit{e.g.,}~}

\definecolor{gray}{HTML}{efefef}
\definecolor{revise}{rgb}{0.0,0.0,0.0}
\ifCLASSOPTIONcompsoc
  \usepackage[nocompress]{cite}
\else
  \usepackage{cite}
\fi

\begin{document}

\title{High-quality Pseudo-labeling for Point Cloud Segmentation with Scene-level Annotation}
\author{Lunhao~Duan,
        Shanshan~Zhao,
        Xingxing~Weng,
        Jing~Zhang,
        Gui-Song~Xia
\IEEEcompsocitemizethanks{
\IEEEcompsocthanksitem Lunhao Duan, Xingxing Weng, and Jing Zhang are with the School of Computer Science, Wuhan University, Wuhan 430072, China (email: lhduan@whu.edu.cn; xingxingw@whu.edu.cn; jingzhang.cv@gmail.com).
\IEEEcompsocthanksitem Shanshan Zhao is with JD Explore Academy, Beijing 100176, China (email: sshan.zhao00@gmail.com).
\IEEEcompsocthanksitem Gui-Song Xia is with the School of Computer Science and the School of Artificial Intelligence, Wuhan University, Wuhan 430072, China (email: guisong.xia@whu.edu.cn).
}
\thanks{(Corresponding author: Shanshan Zhao and Gui-Song~Xia.)}
}

\IEEEtitleabstractindextext{
\justify
\begin{abstract}

This paper investigates indoor point cloud semantic segmentation under scene-level annotation, which is less explored compared to methods relying on sparse point-level labels. 
In the absence of precise point-level labels, current methods first generate point-level pseudo-labels, which are then used to train segmentation models.
However, generating accurate pseudo-labels for each point solely based on scene-level annotations poses a considerable challenge, substantially affecting segmentation performance.
Consequently, to enhance accuracy,
this paper proposes a high-quality pseudo-label generation framework by exploring contemporary multi-modal information and region-point semantic consistency.
Specifically, with a cross-modal feature guidance module, our method utilizes 2D-3D correspondences to align point cloud features with corresponding 2D image pixels, thereby assisting point cloud feature learning.
To further alleviate the challenge presented by the scene-level annotation, we introduce a region-point semantic consistency module. 
It produces regional semantics through a region-voting strategy derived from point-level semantics, which are subsequently employed to guide the point-level semantic predictions. 
Leveraging the aforementioned modules, our method can rectify inaccurate point-level semantic predictions during training and obtain high-quality pseudo-labels.
Significant improvements over previous works on ScanNet v2 and S3DIS datasets under scene-level annotation can demonstrate the effectiveness.
Additionally, comprehensive ablation studies validate the contributions of our approach's individual components.
The code is available at \href{https://github.com/LHDuan/WSegPC}{https://github.com/LHDuan/WSegPC}.

\end{abstract}

}
\maketitle

\IEEEdisplaynontitleabstractindextext
\IEEEpeerreviewmaketitle

\IEEEraisesectionheading{\section{Introduction}\label{sec:introduction}}
3D point cloud segmentation is a foundational vision task, attracting significant research focus due to its pivotal role across diverse domains, such as robotics and augmented reality.
Recently, this task has achieved remarkable performance relying on various efficient point cloud operations~\cite{thomas2019kpconv,hu2021learning, zhu2021cylindrical,zhao2021point,ptv2,wang2023octformer,duan2024condaformer} that follow the seminal work~\cite{qi2017pointnet,qi2017pointnet++}. 
However, high-quality point-level annotations are generally required to train these models for high performance, which are high-cost to obtain. 
To reduce the labeling costs, weakly supervised 3D semantic segmentation has been investigated~\cite{xu2020weakly,hu2022sqn,liu2021one,zhang2021perturbed,yang2022mil,liu2023cpcm, tang2024all}. 
These methods focus on developing 3D point cloud semantic segmentation models under the setting where only a few points in each point cloud sample are manually labeled.
Beyond sparse annotations, utilizing scene-level classification annotations for labeling point clouds offers a more efficient method.
However, this weakly-supervised segmentation significantly lags behind its fully-supervised counterpart.
Consequently, this study focuses on this challenging setting with the goal of achieving substantial performance improvements.

\begin{figure}[tbp]
    \centering
	\small
	\resizebox{0.94\linewidth}{!}{
            \begin{tabular}{cc}
                \includegraphics[width=0.45\linewidth]{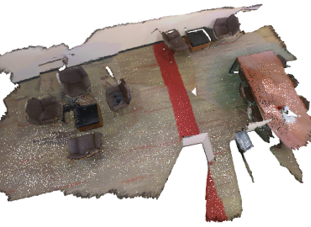}&
                \includegraphics[width=0.45\linewidth]{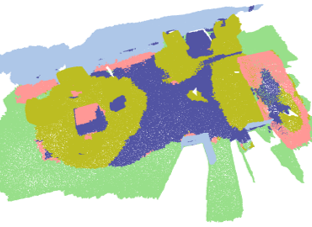}\\
                    Input Point Cloud & Pseudo-Labels of PCAM\\
                \includegraphics[width=0.45\linewidth]{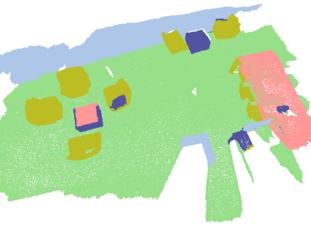}&
                \includegraphics[width=0.45\linewidth]{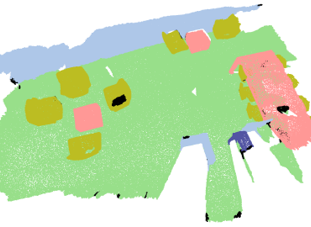}\\
                    Pseudo-Labels of Ours & Ground Truth Labels \\
            \end{tabular}
        }
    \caption{Comparisons between pseudo-labels generated by PCAM~\cite{wei2020multi} and our method. Without precise point-level labels, PCAM can only activate the approximate location of each category, while our method can obtain more accurate segmentation results by introducing cross-modal feature guidance and region-point semantic consistency.}
    \label{fig:teaser}
\end{figure}

Achieving weakly supervised 3D segmentation with scene-level annotation typically involves two stages.
In the initial stage, a weak supervision method is employed to generate pseudo-labels for individual points based on scene-level annotations. 
Subsequently, in the second stage, the semantic segmentation model is trained with the point-wise pseudo-labels in a supervised manner, which is then utilized for inference.
Thus, the quality of the pseudo-labels significantly influences the performance of the ultimate segmentation result.
However, due to the lack of point-level annotations to guide the learning process, generating high-quality pseudo-labels for each point presents a significant challenge when relying exclusively on scene-level annotation.
As an example, each scene from the widely utilized ScanNet dataset~\cite{dai2017scannet} features an average of 148,000 points and contains 8 categories on average across 20 categories, presenting a challenge in determining category assignments for individual points within this large-scale dataset.
Illustrated in Fig.~\ref{fig:teaser}, the pseudo-labels generated by PCAM~(Point Class Activation Map)~\cite{wei2020multi}, a previous algorithm based on Class Activation Map~\cite{zhou2016learning} for weakly supervised point cloud semantic segmentation, contains a lot of errors.

To improve pseudo-label quality, WyPR~\cite{ren20213d} introduces an unsupervised 3D proposal generation algorithm. 
This algorithm employs cross-task and cross-transformation consistency losses to address both semantic segmentation and object detection under scene-level annotation. 
Acknowledging the presence of dense RGB information in indoor point cloud data, some methods~\cite{kweon2022joint, yang20232d} utilize associated 2D images for cross-modal knowledge distillation or feature fusion.
Kweon et al.~\cite{kweon2022joint} focus on the scenarios with image-level annotation, which is equivalent to having annotations for each sub-cloud associated with an image, and utilize 2D class activation maps as supplementary supervision for the 3D branch.
However, in the context of the more challenging scene-level annotation, sharing a unified classification label for multi-view images from the same scene leads to inaccurate class activation maps for each image, which results in inaccurate supervision for the 3D branch.
MIT~\cite{yang20232d} introduces a multi-modal interlaced transformer model aiming at fusing 2D and 3D features.
It averages point features within regions, classifying each region as a token.
After training the transformer model, MIT assigns the pseudo-label to each point based on the predicted category of its corresponding region, subsequently using these pseudo-labels to train the segmentation network.
However, region partitions can be inaccurate and points within the same region may actually belong to different categories. 
Directly using regional pseudo-labels may lead to inaccurate point-level pseudo-labels for corresponding points within each region if the regional pseudo-labels are incorrect.

Inspired by the preceding analysis, this study aims to address the challenge of pseudo-labeling by more efficiently leveraging contemporary image information and introducing a region-point semantic consistency constraint. 
Specifically, we first develop a cross-modal feature guidance based on 2D-3D correspondences.
{
For both 2D-to-3D and 3D-to-2D guidance, based on systematic ablation studies conducted specifically for this weakly supervised task, we employ a point-wise contrastive distillation to align the features of 3D points with their corresponding 2D pixels via a contrastive loss~\cite{hadsell2006dimensionality,oord2018representation}. 
}
In comparison with~\cite{kweon2022joint}, our method effectively guides feature learning even when class activation maps are inaccurate under scene-level annotation. 

Subsequently, we propose a region-point semantic consistency module. 
{
Operating within a teacher-student framework, this module leverages unsupervised region partitioning~\cite{lafarge2012creating} to enforce semantic consistency between derived local region semantics and individual point predictions. 
Unlike prior work like MIT~\cite{yang20232d} that directly uses regional predictions and, unlike point-to-point distillation objectives, our module explicitly addresses potential noise in partitions and weak predictions using a dynamic category threshold strategy. This allows robust extension of coarse scene-level tags to reliable local guidance. 
}
Specifically, local regions are extracted using an unsupervised shape extraction algorithm~\cite{lafarge2012creating}, and regional semantic outputs are derived by averaging point-level outputs of the teacher model within each region.
The selected regional outputs from the teacher model then serve as guidance to supervise the point-level outputs from the student model, achieving the semantic consistency between point-level outputs and regional outputs.
During network training, the student model's point-level predictions are continuously enhanced by the region-point semantic consistency module, leading to improved regional predictions for guidance.
This iterative process will eventually lead to a significant improvement in the quality of the pseudo-labels.

To demonstrate the effectiveness of our method, we conduct experiments under scene-level annotation on ScanNet v2~\cite{dai2017scannet} and S3DIS~\cite{s3dis} datasets, achieving a significant improvement compared to prior works and establishing a new state-of-the-art. 
In summary, the contributions of this paper are three-fold:
\begin{itemize}
    \item This paper studies the 3D weakly supervised point cloud semantic segmentation with scene-level annotation by developing a high-quality pseudo-label generation framework.
    \item The proposed method first achieves cross-modal feature guidance by constructing a point-wise contrastive distillation loss, and then introduces a region-point semantic consistency module to improve the point-level semantic predictions.
    \item Experiments were conducted on ScanNet v2 and S3DIS datasets, demonstrating our method's effectiveness. Ablation studies were also performed to provide a thorough analysis.
\end{itemize}

\section{Related Work}
In this section, we review previous methods for weakly supervised point cloud semantic segmentation and cross-modal knowledge distillation.
\begin{figure*}[tbp]
    \centering
    \includegraphics[width=0.85\textwidth]{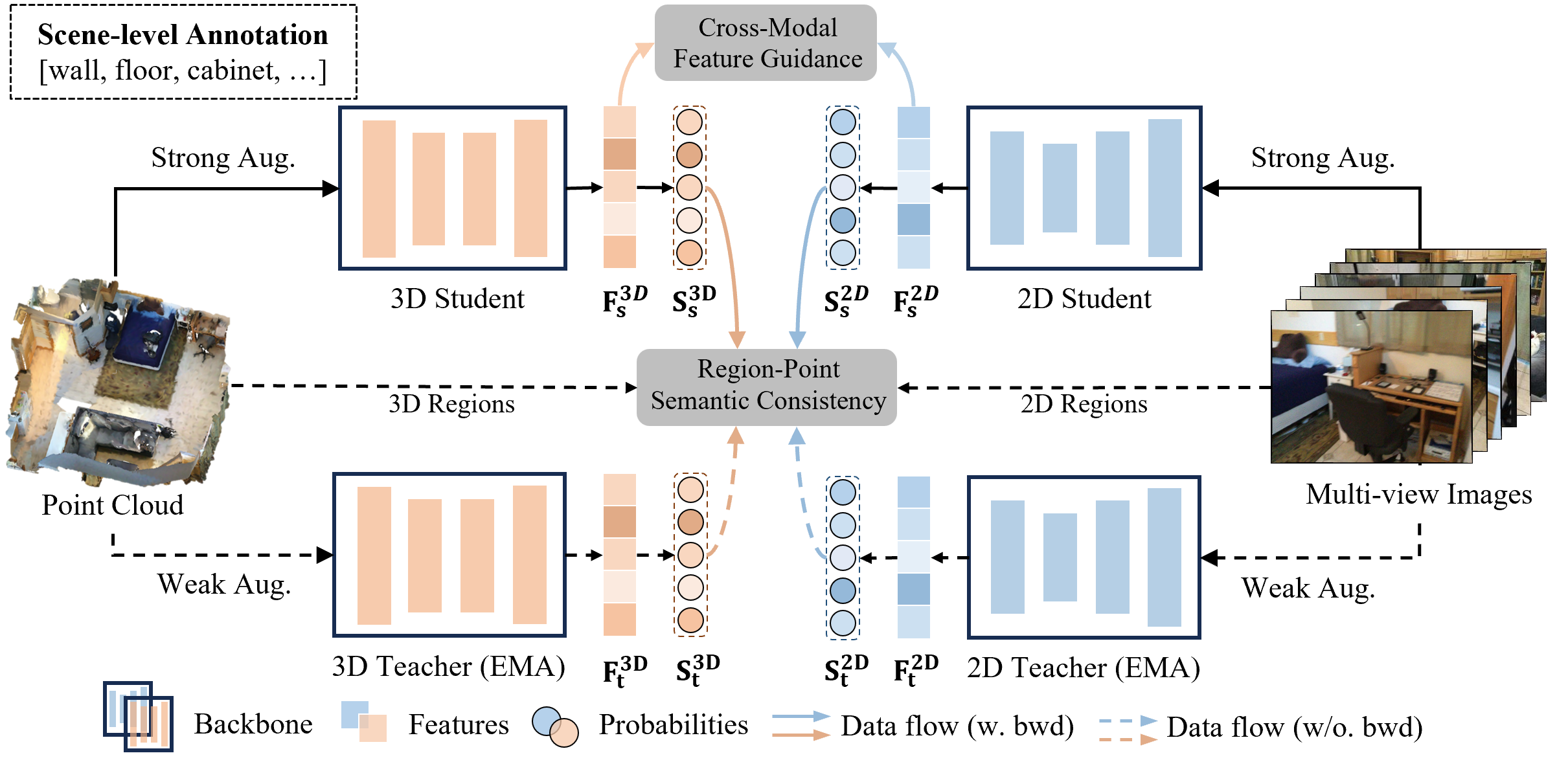}
    \caption{Overview of the proposed framework: The point cloud and multi-view images are initially processed by the 3D and 2D branches to extract corresponding features and classification probabilities, respectively. Both the 3D and 2D branches comprise a teacher network and a student network. Subsequently, the features from the student networks of the 3D and 2D branches are aligned using the cross-modal feature guidance module. The point-wise classification probabilities from the student network are aligned with the regional classification probabilities from the teacher network using the region-point semantic consistency module. For brevity, the multi-label classification losses between scene-level annotation and classification probabilities are not presented. }
    \label{fig:model}
\end{figure*}

\paragraph{Weakly Supervised Point Cloud Segmentation.}
Weakly supervised point cloud semantic segmentation has been extensively studied in recent years.
According to the type of weak labels, this task can be divided into three categories: sparsely labeled points, subcloud-level labels, and scene-level labels.
Among them, 3D semantic segmentation with the supervision of sparsely labeled points attracts the most attention from the community.
For example, OTOC~\cite{liu2021one} and PSD~\cite{zhang2021weakly} utilize graph propagation to propagate the limited point labels to the remaining unlabeled points.
There are some methods~\cite{zhang2021perturbed,li2022hybridcr,yang2022mil,wu2022dual} leveraging predictive consistency under various perturbations to provide more constraints.
In comparison, only very few works study the setting with subcloud-level or scene-level labels.
The seminal work, MPRM~\cite{wei2020multi}, proposes a multi-path region mining module with various attention mechanisms to produce discriminative features from different aspects, basically following the CAM strategy in 2D images.
WyPR~\cite{ren20213d} develops an unsupervised 3D proposal generation algorithm to construct a cross-task consistency loss between segmentation and object detection tasks.
WyPR also proposes a smoothness regularization loss to improve local smoothness, but with only marginal performance improvement.
Subsequently, MIL~\cite{yang2022mil} explores pair-wise cloud-level supervision to mine additional supervisory signals and achieves better segmentation results than MPRM and WyPR with subcloud-level labels.
Recently, a Multimodal Interlaced Transformer (MIT)~\cite{yang20232d} is introduced, which achieves 2D and 3D feature fusion for weakly supervised point cloud segmentation.
Xia et al.~\cite{xia2023densify} propose to allocate scene-level labels to over-segment point cloud clusters by bipartite matching.
However, these studies suffer from the low quality of pseudo-labels under scene-level annotation. 
In this paper, we propose to explore cross-modal feature guidance and local region semantic consistency to improve the quality of pseudo-labels for training segmentation network.

\vspace{-2mm}
\paragraph{Cross-modal Knowledge Distillation.}
Knowledge distillation (KD) ~\cite{hinton2015distilling} is proposed to transfer rich useful information from a teacher model to a student model.
The student network is usually constrained to learn some of the characteristics of the teacher network, \myeg intermediate features or output probabilities.
Many works have adapted KD techniques to distill knowledge from a 2D teacher network trained on images to a 3D student network~\cite{jaritz2022cross}.
For instance, paired RGB-Depth images are used in~\cite{gupta2016cross} to achieve knowledge transfer from a 2D image teacher network trained on ImageNet to a 3D student network by aligning mid-level semantic features of the networks.
Recently, PPKT~\cite{liu2021learning} and SLidR~\cite{sautier2022image} propose to utilize a pre-trained 2D network to help 3D point cloud model pre-training by applying point-pixel and superpoint-superpixel contrastive distillation, respectively. 
{Multi-modal knowledge distillation techniques are also explored in 2DPASS~\cite{yan20222dpass} for point cloud semantic segmentation. 
For joint optical and scene flow estimation, CamLiFlow~\cite{liu2022camliflow} introduces a module for bidirectional camera-LiDAR feature fusion. 
Additionally, transformer-based frameworks like BrT~\cite{wang2022bridged} aim to strengthen image-point correlations for 3D object detection through dedicated bridging mechanisms, such as conditional object queries and point-to-patch projection.}
In the weakly supervised 3D semantic segmentation scenario, Kweon~et al.~\cite{kweon2022joint} leverage the complementary benefits from the 2D image and 3D point cloud to jointly perform 2D-3D weakly supervised semantic segmentation.
Inspired by the previous works, in this paper we explore an efficient way to utilize the features from 2D images for providing point-level guidance under scene-level annotation.

\section{Method}
\subsection{Overview}
\paragraph{Problem Statement.}
The training dataset comprises $T$ samples, each including a point cloud $\cal P$, its corresponding multi-view images $\cal I$, and weak labels $\bf y$.
The point cloud ${\cal P} \in\mathbb{R}^{N\times 6}$ is composed of $N$ points, with each point characterized by spatial coordinates and RGB colors. 
The multi-view images ${\cal I}\in\mathbb{R}^{V\times H\times W\times3}$ comprise $V$ RGB images, each with a resolution of $H\times W$.
The scene-level weak label ${\bf y} \in\mathbb{R}^{C}$ denotes the presence of any of the $C$ categories within the point cloud.
Given this training set with only scene-level annotation, our objective is to train a point-wise segmentation model that can classify each point within the point cloud into one of the $C$ categories. 

\paragraph{Proposed Framework.}
The architecture of our method is illustrated in Fig.~\ref{fig:model}.
It comprises four main components: a 2D network, a 3D network, a cross-modal feature-guidance module, and a region-point semantic consistency module.
Both the 2D and 3D networks are comprised of a backbone network and classification layers.
Let ${\bf F}^{\textsc{2d}}\in \mathbb{R}^{V\times H\times W\times D} $ and ${\bf F}^{\textsc{3d}}\in \mathbb{R}^{N\times L} $ represent features extracted by the backbone networks, where $D$ and $L$ denote the dimensions of 2D and 3D features, respectively, and ${\bf S}^{\textsc{2d}}\in \mathbb{R}^{V\times H\times W\times C}$ and ${\bf S}^{\textsc{3d}}\in \mathbb{R}^{N\times C}$ represent classification predictions made by the classification layers.
The scene-level classification scores for 3D and 2D are obtained by applying average pooling on classification predictions of all points and all pixels across $V$ views.
Then, the multi-label classification losses $\mathcal{L}^{\textsc{3d}}_{\text{cls}}$ and $\mathcal{L}^{\textsc{2d}}_{\text{cls}}$ are calculated between the scene-level classification scores and label $\bf y$.
In the cross-modal feature-guidance module, representative features ${\bf F}^{\textsc{2d}}$ and ${\bf F}^{\textsc{3d}}$ are aligned using specifically developed cross-modal feature guidance constraints.
In the region-point semantic consistency module, high-quality regional semantic guidance is generated from the teacher networks, which are then used to supervise the student networks.
The details of our model are presented in the subsequent sections.

\subsection{Cross-Modal Feature Guidance}
Inspired by prior cross-modal knowledge distillation research~\cite{sautier2022image,kweon2022joint,yang20232d}, dense 2D images are employed to offer cross-modal feature guidance. 
For guidance from both 2D to 3D and 3D to 2D, widely-used point-wise contrastive distillation is introduced to distill knowledge between the image and point cloud networks. 

\paragraph{From 2D to 3D.}
Given that 2D and 3D features are extracted from distinct backbones, they often exhibit different channel dimensions ($D\neq L$). 
To align 3D features with corresponding 2D features, 3D features are projected to match the dimensions of 2D features using fully connected layers, and projected 3D features are denoted by ${\bf F}^{\textsc{3d}}_{\text{proj}}$.
For the $i$'th point, following~\cite{hu2021bidirectional}, the corresponding pixel coordinate $[v_i,h_i,w_i]$ are obtained using known camera parameters, where $v_i$ is the index of the view and $[h_i,w_i]$ are the 2D coordinate in $v_i$'th image.
Point-pixel feature pairs are sampled from ${\bf F}^{\textsc{3d}}_{\text{proj}}$ and ${\bf F}^{\textsc{2d}}$. Here, ${\bf F}^{\textsc{2d}}$ is detached to solely provide guidance without initiating gradient backpropagation generated by the subsequent loss.
And the sampled point index set is denoted as ${\mathcal G^{\textsc{3d}}}$.
The contrastive loss from 2D to 3D is calculated as follows:

\begin{equation}\label{eq:2dto3d}
   \mathcal{L}_{\text{con}}^{\textsc{3d}}
   = -\sum_{i\in{\mathcal G^{\textsc{3d}}}} \log 
   \frac{\exp({\bf f}^{\textsc{3d}}_i {{\bf f}^{\textsc{2d}}_i}^\top / \tau)}
   {\sum\limits_{j\in{\mathcal G^{\textsc{3d}}}}
   \exp({\bf f}^{\textsc{3d}}_i {{\bf f}^{\textsc{2d}}_j}^\top / \tau)},
\end{equation}
where ${\bf f}^{\textsc{3d}}_i = {\bf F}_{\text{proj}}^{\textsc{3d}}[i,\cdot]$, ${\bf f}^{\textsc{2d}}_i = {\bf F}^{\textsc{2d}}[v_i,h_i,w_i,\cdot]$, 
and $\tau > 0$ serves as a temperature parameter.
This formula guarantees that a point feature ${\bf f}^{\textsc{3d}}_i$ has a stronger correlation with its corresponding pixel feature ${\bf f}^{\textsc{2d}}_i$ than with any other pixel features ${\bf f}^{\textsc{2d}}_j$ ($j \neq i$).

\paragraph{From 3D to 2D.}
We achieve the distillation from 3D to 2D in a manner akin to the process from 2D to 3D: 1) 2D features are initially projected to align with the dimensions of 3D features; 2) point-pixel feature pairs are sampled to calculate the contrastive loss $\mathcal{L}_{\text{con}}^{\textsc{2d}}$ with a similar form to Eq.~\ref{eq:2dto3d}.

Unlike the previous work~\cite{kweon2022joint}, which uses guidance such as class activation map and feature correlation matrix, we directly project the features from one mode to another mode, and leverage contrastive loss to align the projected features to the original features of the other mode.
This contrastive loss reduces the distance between corresponding points and pixel features while increasing the distance between non-corresponding points and pixel features, thereby enhancing the feature's discriminability. 
We compare the performance of various guidance methods in the experimental section.

\subsection{Region-Point Semantic Consistency}
Given that point-level semantic outputs under scene-level annotations are inherently noisy, we employ region-voting within locally consistent regions to produce more reliable regional semantic guidance.
To this end, we adopt a teacher-student framework where the teacher model generates stable regional semantic outputs to supervise the student model and the teacher model’s parameters are updated from the student model via an exponential moving average (EMA) approach.
Additionally, acknowledging the category imbalances, an adaptive filtering threshold strategy is devised, establishing dynamically adjusted thresholds for each category contingent on the frequency of occurrences.

\paragraph{Regional Semantic Guidance Generation.}
Given the input point cloud $\cal P$, local region partitioning is performed to obtain a set of regions $\cal R$ using an unsupervised shape extraction algorithm~\cite{lafarge2012creating}.
For each local region, regional semantic outputs $\bar{\bf S}^{\textsc{3d}}_{\text{t}}\in \mathbb{R}^{M\times C}$ are obtained by averaging point-level semantic outputs ${\bf S}^{\textsc{3d}}_{\text{t}}$ from the teacher model, with $M$ representing the number of regions.
For a region ${\cal R}[m]$, the calculation process is expressed as follows:
\begin{equation}
   \bar{\bf S}^{\textsc{3d}}_{\text{t}}[m,\cdot] = \frac{1}{|{\cal R}[m]|}\sum_{{p} \in {\cal R}[m]}{{\bf S}^{\textsc{3d}}_{\text{t}}[p, \cdot]},
\end{equation}
where $|\cdot|$ represents the number of points belonging to ${\cal R}[m]$ and $p$ denotes the index of the point belonging to ${\cal R}[m]$.
Subsequently, the point-level semantic guidance for points $p \in {\cal R}[m]$ is derived from the regional semantic outputs of region ${\cal R}[m]$:
\begin{equation}
\begin{aligned}
    c &= \text{argmax}(\bar{\bf S}^{\textsc{3d}}_{\text{t}}[m,\cdot]);\\
    \hat{\bf Y}^{\textsc{3d}}_{\text{t}}[p,c] &= 1,~
    \hat{\bf S}^{\textsc{3d}}_{\text{t}}[p,\cdot] = \bar{\bf S}^{\textsc{3d}}_{\text{t}}[m,\cdot], \forall~ p \in {\cal R}[m],
\end{aligned}
\end{equation}
where $\hat{\bf Y}^{\textsc{3d}}_{\text{t}}$ is the category label for each point and initialized to all zeros, and $\hat{\bf S}^{\textsc{3d}}_{\text{t}}$ is the probability of each point belonging to each category. 
Finally, the region-point semantic consistency loss is calculated as follows:
\begin{equation}
    \mathcal{L}_{\text{rpc}}^{\textsc{3d}} 
    = -\frac{1}{N}\sum_{{p}=1}^{N}\sum_{c=1}^{C} 
    {\hat{\bf Y}^{\textsc{3d}}_{\text{t}}}[p,c])
    \log({{\bf S}^{\textsc{3d}}_{\text{s}}[p,c]}).
\end{equation}

\paragraph{Adaptive Filtering  Threshold.}
The aforementioned loss function incorporates all semantic outputs for consistency supervision, including noisy semantic outputs with low scores, i.e., $\hat{\bf S}^{\textsc{3d}}_{\text{t}}$ with minimal values.
To disregard these noisy outputs, a simple method involves establishing a fixed threshold $\eta$ to exclude points belonging to category $c$ with lower scores, i.e., $\hat{\bf S}^{\textsc{3d}}_{\text{t}}[{p},c] < \eta$.
However, applying a uniform threshold across all categories overlooks the disparity in point cloud quantities among categories, leading to a bias in selected predictions towards categories with a higher prevalence of point clouds. 
To mitigate this concern, an adaptive threshold ${\bf T}^{\textsc{3d}}\in \mathbb{R}^{C}$ is introduced, which adjusts inversely to the number of predictions per category, based on a base threshold $\eta$, drawing inspiration from FlexMatch~\cite{flexmatch}.
Specifically, an array ${\bf Q}^{\textsc{3d}}\in \mathbb{R}^{C}$ is utilized to enumerate the semantic outputs for each category: 
\begin{equation}
{\bf Q}^{\textsc{3d}}[c] = |\hat{\bf Y}^{\textsc{3d}}_{\text{t}}[\cdot,c] == 1|,
\end{equation}
where $|\cdot|$ means the number of points belonging to category $c$. 
The threshold for each category is calculated as follows:
\begin{equation}
{\bf T}^{\textsc{3d}}[c] = \frac{{\bf Q}^{\textsc{3d}}[c]}{\text{max}({\bf Q}^{\textsc{3d}})}\eta.
\end{equation}
Subsequently, the point selection mask ${\bf K}\in \mathbb{R}^{N}$ is derived as:
\begin{equation}
{\bf K}[{p}] = \mathbb{I}(\hat{\bf S}^{\textsc{3d}}_{\text{t}}[{p},c] > {\bf T}^{\textsc{3d}}[c]), \forall~ \hat{\bf Y}^{\textsc{3d}}_{\text{t}}[{p},c] = 1.
\end{equation}
Finally, the region-point semantic consistency loss is reformulated as:
\begin{equation}
    \mathcal{L}_{\text{rpc}}^{\textsc{3d}} 
    = -\frac{1}{N}\sum_{{p}=1}^{N}{\bf K}[{p}]\sum_{c=1}^{C} 
    {\hat{\bf Y}^{\textsc{3d}}_{\text{t}}}[{p},c])
    \log({{\bf S}^{\textsc{3d}}_{\text{s}}[{p},c]}).
\end{equation}
For the 2D image branch, 2D pixels to are first projected to 3D space using depth map and camera intrinsic, and local regions are extracted with the same algorithm used in 3D branch. 
Then the region-point semantic consistency loss $\mathcal{L}_{\text{rpc}}^{\textsc{2d}}$ of the 2D image branch can be calculated in the same way.

\paragraph{Mix3D Augmentations.}
The widely utilized indoor point cloud segmentation dataset, e.g. ScanNet v2~\cite{dai2017scannet}, comprises approximately 1200 training scenes, resulting in limited scene-level annotations for weakly supervised 3D semantic segmentation.
To diversify scene-level annotations, the Mix3D augmentation~\cite{nekrasov2021mix3d} is employed to randomly combine two scenes.
Specifically, for a scene-level point cloud, a scene is randomly selected from the remaining scenes and combined.
Consequently, the scene-level weak label for the mixed scene encompasses all categories present in both scenes.
Notably, the point cloud input is mixed solely for the student network, whereas the teacher network receives the original, separate point clouds.
Therefore, the Mix3D augmentation also acts as a strong data augmentation for the student network, enhancing its robustness.
\subsection{Final Objective}
The final loss function comprises three components: multi-label classification losses $\mathcal{L}^{\textsc{3d}}_{\text{cls}}$, $\mathcal{L}^{\textsc{2d}}_{\text{cls}}$, point-wise contrastive distillation losses $\mathcal{L}_{\text{con}}^{\textsc{3d}}$, $\mathcal{L}_{\text{con}}^{\textsc{2d}}$, and region-point semantic consistency losses $\mathcal{L}_{\text{rpc}}^{\textsc{3d}}$, $\mathcal{L}_{\text{rpc}}^{\textsc{2d}}$:
\begin{equation}
\mathcal{L} = \mathcal{L}^{\textsc{3d}}_{\text{cls}} + \mathcal{L}^{\textsc{2d}}_{\text{cls}} +\alpha \mathcal{L}_{\text{con}}^{\textsc{3d}} +\alpha \mathcal{L}_{\text{con}}^{\textsc{2d}}+ \beta \mathcal{L}_{\text{rpc}}^{\textsc{3d}} + \beta \mathcal{L}_{\text{rpc}}^{\textsc{2d}},
\end{equation}
where $\alpha$ and $\beta$ are the loss coefficients.

\section{Experiments}
This section presents experimental results on public datasets and ablations to show the effectiveness of our method and the associated components. 
Additional results can be found in the supplementary material.

\subsection{Datasets and Evaluation Metrics}

\paragraph{ScanNet v2.} The ScanNet v2 dataset~\cite{dai2017scannet} comprises 1513 training scenes and 100 test scenes, encompassing over 2.5 million RGB-D images.
The official train-validation split is utilized, featuring 1201 scenes for training and 312 scenes for validation.
Following MIT~\cite{yang20232d}, one image every 20 frames is sampled from the RGB-D video for use as the corresponding image input. 
The input point clouds and images are annotated with 20 semantic categories.

\paragraph{S3DIS.} The S3DIS dataset~\cite{s3dis} encompasses 271 room scenes, containing 90,496 RGB-D images from six areas. 
Following previous methods, scenes from Area 5 are reserved for validation, with the remaining scenes dedicated to training. 
The input point clouds and images are annotated with 13 semantic categories.

\paragraph{Evaluation Metrics.}
The evaluation metric is the mean intersection-over-union (mIoU) across all classes. 

\subsection{Implementation Details}
\vspace{-2mm}
\paragraph{Data Processing and Augmentation.} 
For each image, projection between 3D points and corresponding image pixels is computed using camera parameters (extrinsic and intrinsic), yielding point-pixel pair inputs for cross-modal feature guidance.
The classification label corresponds to the classes present in the scene-level point cloud.
Upon exploring the datasets, it was discovered that the categories of wall, ceiling, and floor are present in every scene of the S3DIS dataset, with a similar occurrence in the ScanNet dataset where walls and floor are prevalent in most scenes. 
Such phenomenon is also noted in~\cite{xia2023densify}.
This phenomenon results in a scarcity of negative classification samples for these categories during the calculation of the multi-label classification loss $\mathcal{L}^{\textsc{3d}}_{\text{cls}}$, meaning that in most scene-level classification labels, these categories are labeled as 1. 
Therefore, similar to~\cite{xia2023densify}, a data augmentation strategy based on the prior distribution of indoor point clouds is adopted. 
Specifically, in each scene viewed as a cuboid, the region with the most points within the bottom 0.5 meters is randomly removed, and the classification label for the floor is set to 0. 
For the ceiling category, the selection occurs from the region within the top 0.5 meters. 
For the wall category, four regions are chosen from each of the surrounding sides.
\begin{table}[t!]
    \begin{minipage}{.49\textwidth}
        \centering
        \small
        \caption{
            Quantitative results (mIoU) of compared methods with diverse supervisions settings on the ScanNet and S3DIS datasets.``Sup.'' indicates the type of supervision. ``$\mathcal{F.}$'',  ``$\mathcal{P.}$'', ``$\mathcal{S.}$'', and ``$\mathcal{I.}$'' denote full annotation, sparsely labeled points, scene-level annotation, and image-level annotation, respectively. {``$^\star$" indicates methods using a 2D image backbone.}
        }
        \label{tab:sota}
        \vspace{-2mm}
        \scalebox{0.9}{
            \begin{tabular}{lccccc}
                \toprule[1pt] 
                \multirow{2}{*}{Method} & \multirow{2}{*}{Publication} &  \multirow{2}{*}{Sup.} &  \multicolumn{2}{c}{ScanNet}  & {S3DIS} \\
                 &   &   & Val  &  Test  &  Test  \\
                 \midrule
                MinkUNet~\cite{choy20194d} &CVPR 19& $\mathcal{F.}$ & 72.2 & 73.6 & 65.4  \\ 
                \midrule
                MIL-Trans~\cite{yang2022mil} &CVPR 22& $\mathcal{P.}$ &  57.8 & 54.4 & 51.4 \\
                {MIT$^\star$}~\cite{yang20232d} &ICCV 23& $\mathcal{P.}$&  61.9 & - & - \\
                \midrule
                {Kweon~et al.$^\star$}~\cite{kweon2022joint} &NeurIPS 22& $\mathcal{S.+I.}$ & 49.6 & 47.4 & -\\ 
                {MIT$^\star$}~\cite{yang20232d} &ICCV 23& $\mathcal{S.+I.}$&  45.4 & - & - \\
                \midrule
                MPRM~\cite{wei2020multi} &CVPR 20& $\mathcal{S.}$  &  24.4 & - & 10.3 \\
                MIL-Trans~\cite{yang2022mil} &CVPR 22& $\mathcal{S.}$&  26.2 & - & 12.9 \\
                WYPR~\cite{ren20213d}&CVPR 21& $\mathcal{S.}$ &  29.6 & 24.0 & 22.3 \\
                {MIT$^\star$}~\cite{yang20232d} &ICCV 23& $\mathcal{S.}$&  35.8 & 31.7 & 27.7 \\ 
                Xia et al.~\cite{xia2023densify} &arXiv 23& $\mathcal{S.}$&  38.1 & 35.1 & 25.7 \\ 
                \cmidrule(lr){1-6}
                Baseline &-& $\mathcal{S.}$&  26.5 & - & 31.5\\ 
                {\textbf{Ours$^\star$}} &-& $\mathcal{S.}$&  \textbf{46.9} & \textbf{46.8} & \textbf{47.4} \\ 
                \bottomrule[1pt]
            \end{tabular}
        
        }
        \vspace{-3mm}
    \end{minipage}
\end{table}

\paragraph{Network Architecture.} 
During the training of the first stage for pseudo-label generation, following~\cite{yang20232d}, Mink U-Net 18a~\cite{choy20194d} is adopted as the 3D backbone network, and ResNet50~\cite{he2016deep}, initialized with ImageNet~\cite{deng2009imagenet} pre-trained weights, serves as the 2D backbone network.
In the feature guidance module, temperature $\tau$ is set to 0.07, and 512 point-pixel pairs are randomly selected for computing the contrastive loss.
The loss coefficients $\alpha$ and $\beta$ are set to 0.05 and 1.0, respectively.
Following~\cite{yang20232d}, Mink U-Net 18c~\cite{choy20194d} is selected as the 3D segmentation model for the second stage.

\paragraph{Training.} 
The networks in the first stage are trained using the SGD optimizer with a base learning rate of 0.1, weight decay of 0.0001, momentum of 0.9, and a batch size of 4.
The networks are trained for 300 epochs on ScanNet dataset and 1200 epochs for S3DIS dataset.
The poly learning rate decay scheduler, with a power of 0.9, is adopted.
Following the conventions of previous methods, point-level pseudo-labels are generated through inference on the training set.
These pseudo-labels are then utilized to train the segmentation network in the second stage.
\vspace{-3mm}
\subsection{Main Results}
In our experiments, the performance of our method is compared against state-of-the-art methods in 3D weakly-supervised semantic segmentation.
The baseline method is implemented with only multi-label classification losses.
In addition to methods with scene-level annotation, methods utilizing varied types of supervision are presented, including full annotation, sparsely labeled points, and image-level annotation.

\paragraph{Quantitative Results.}
Results from full annotation can be regarded as performance upper bounds for weakly supervised methods.
In the setting of sparsely labeled points, one point from each category in every scene is labeled. 
Unlike scene-level labeling, sparsely labeling points demands knowledge of the categories present in each scene and identification of a point's location within each category. 
According to~\cite{yang20232d}, sparsely labeling points requires 2 minutes per scene, over twice the time needed for scene-level labeling, which requires less than 1 minute.
As illustrated in Table~\ref{tab:sota}, the MIT~\cite{yang20232d} and MIL-Trans\cite{yang2022mil} methods significantly outperform their scene-level annotation results when using sparsely labeled points. 
This underscores the challenges and potential for improvement in 3D semantic segmentation under scene-level annotation, due to the absence of precise point-level label information.
Among all methods with scene-level annotation, our method achieves the best performance, surpassing the state-of-the-art method Xia et al.~\cite{xia2023densify} and MIT~\cite{yang20232d} by 8\% and 19\% mIoU on the ScanNet and S3DIS datasets, respectively.
Our method has significantly narrowed the performance gap between sparsely labeled points and scene-level annotation.
Additionally, our method's performance under scene-level annotation is on par with that of Kweon et al.\cite{kweon2022joint} and MIT~\cite{yang20232d} methods using extra image-level annotation.
However, for each scene, they extra use an average of 16 images for annotation, whereas our approach requires only a single scene-level annotation, significantly reducing the annotation cost. 
These results demonstrate the effectiveness of our method.

\begin{figure}[t!]
	\centering
	\small
	\resizebox{0.99\linewidth}{!}{
            \setlength{\tabcolsep}{0.5pt}
            \begin{tabular}{cccc}
            	Input & Ground truth & Baseline & Ours \\
            	\includegraphics[width=0.245\linewidth]{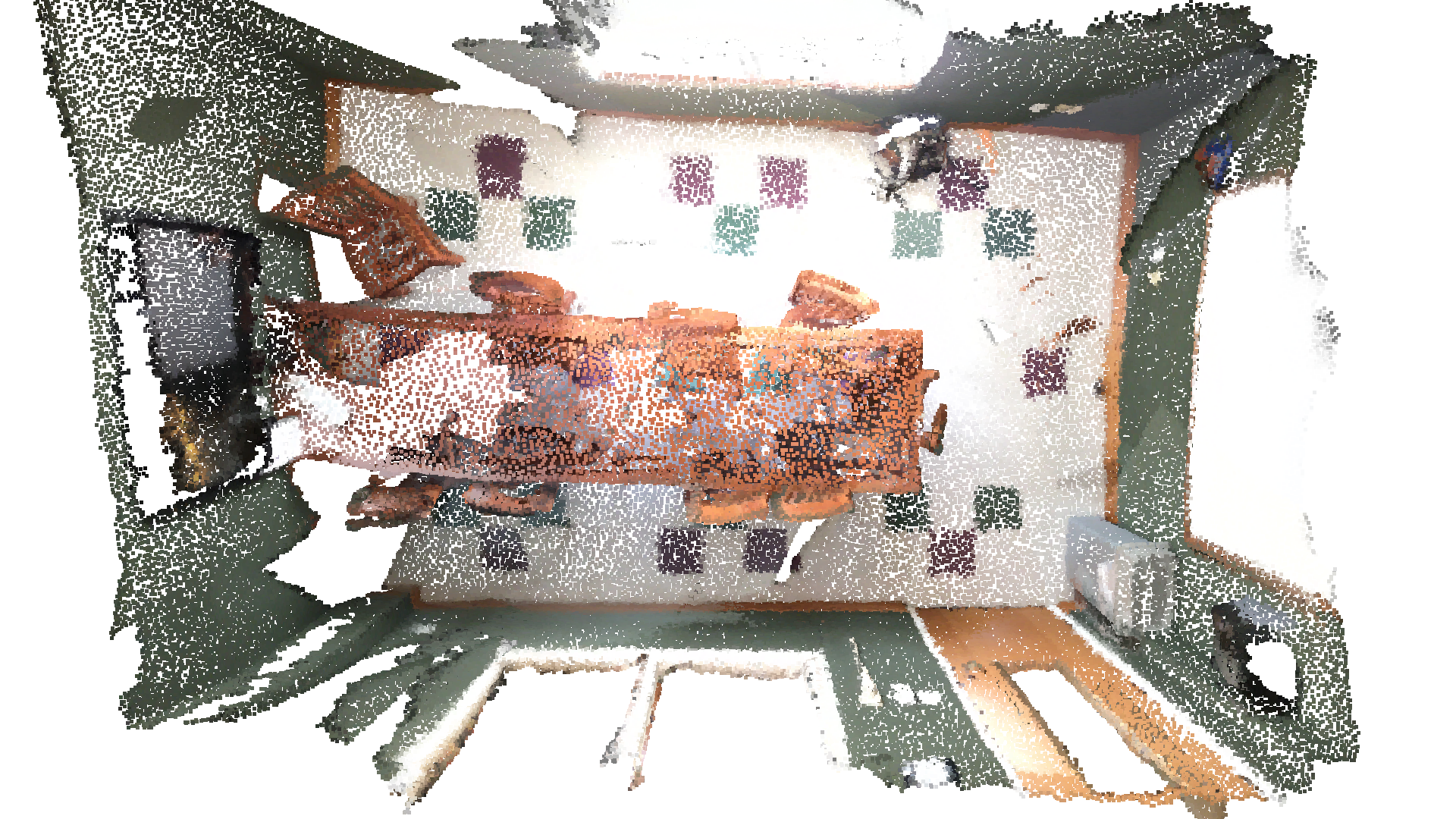}&
            	\includegraphics[width=0.245\linewidth]{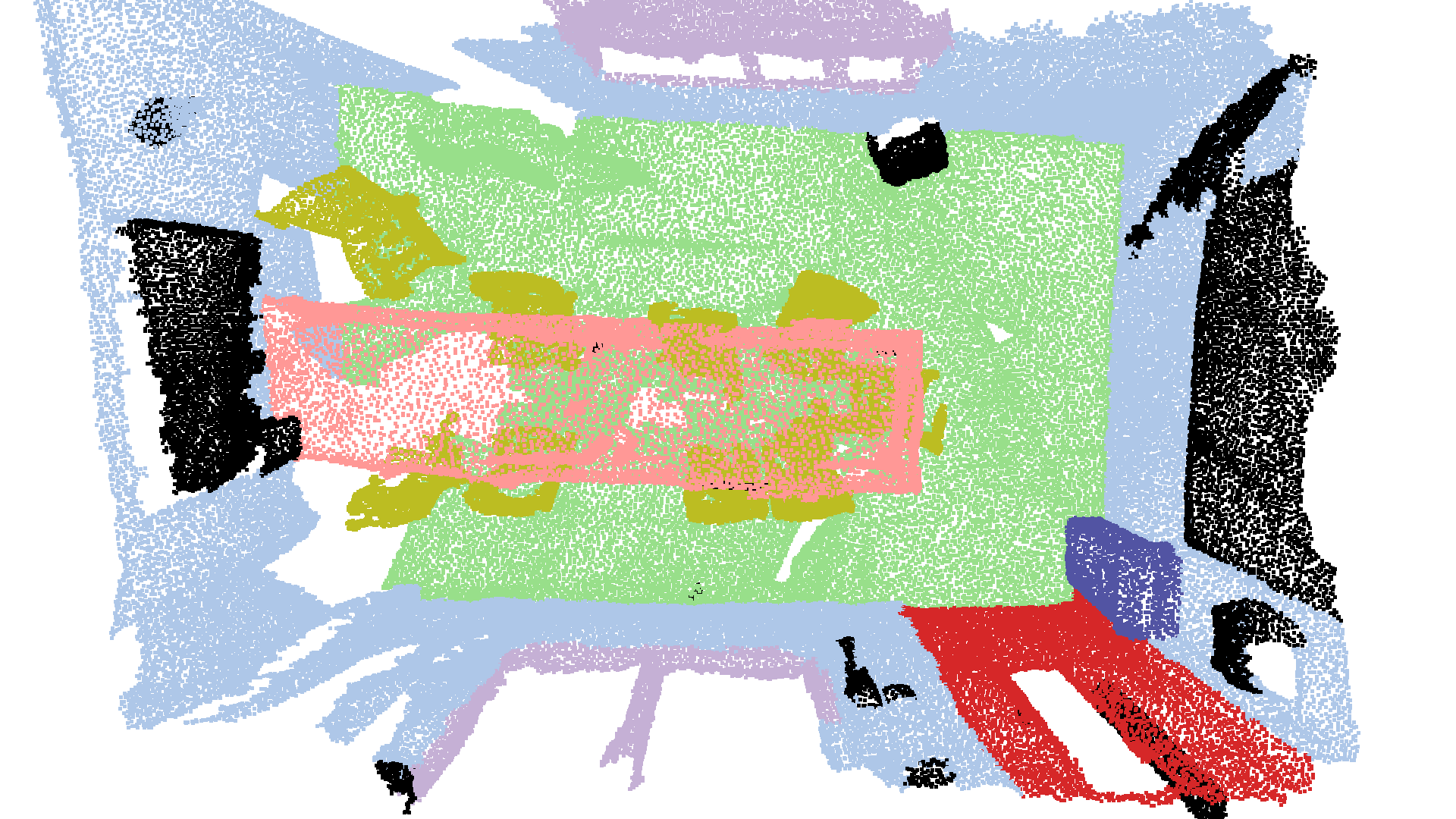}&
            	\includegraphics[width=0.245\linewidth]{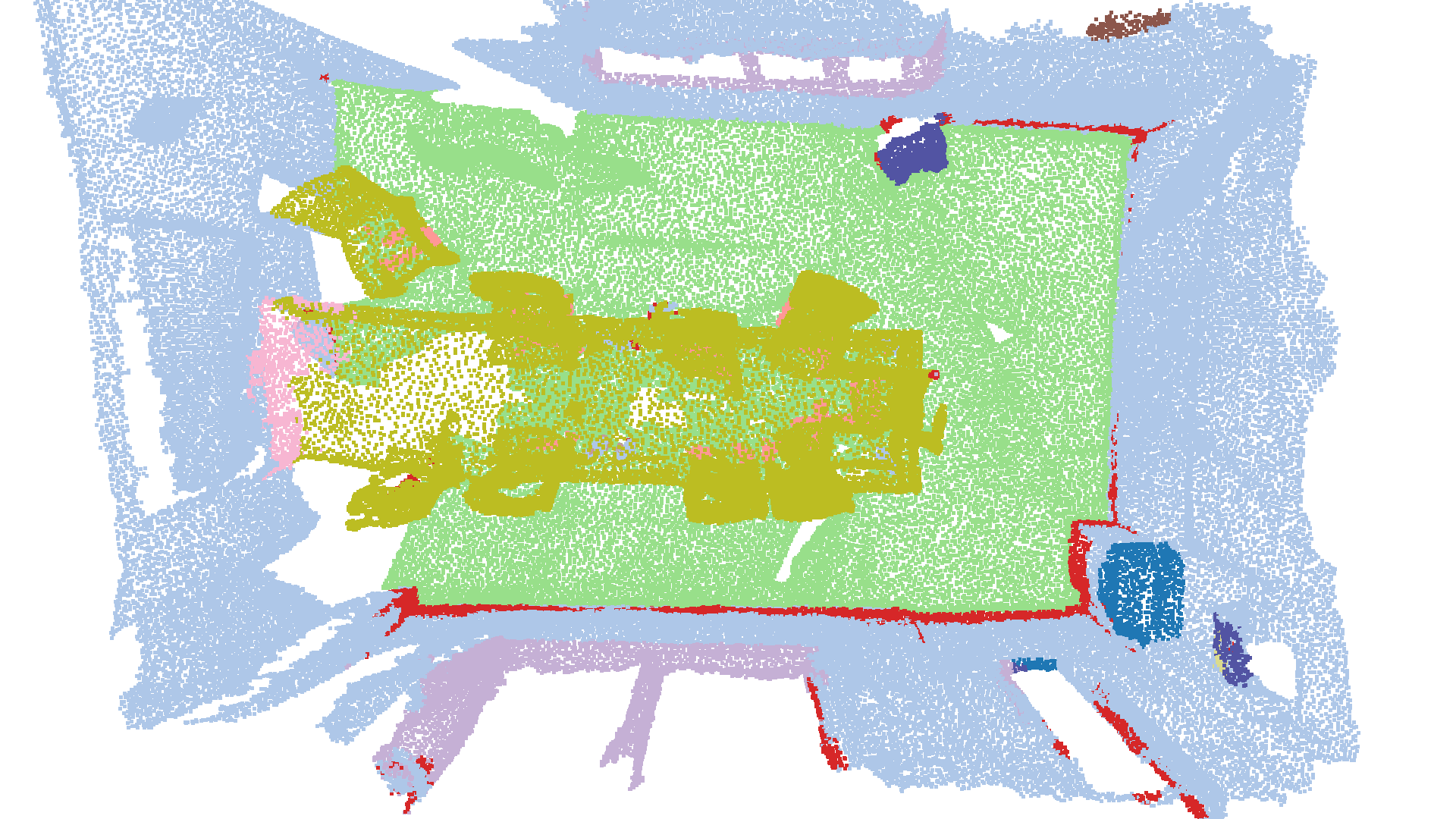}&
            	\includegraphics[width=0.245\linewidth]{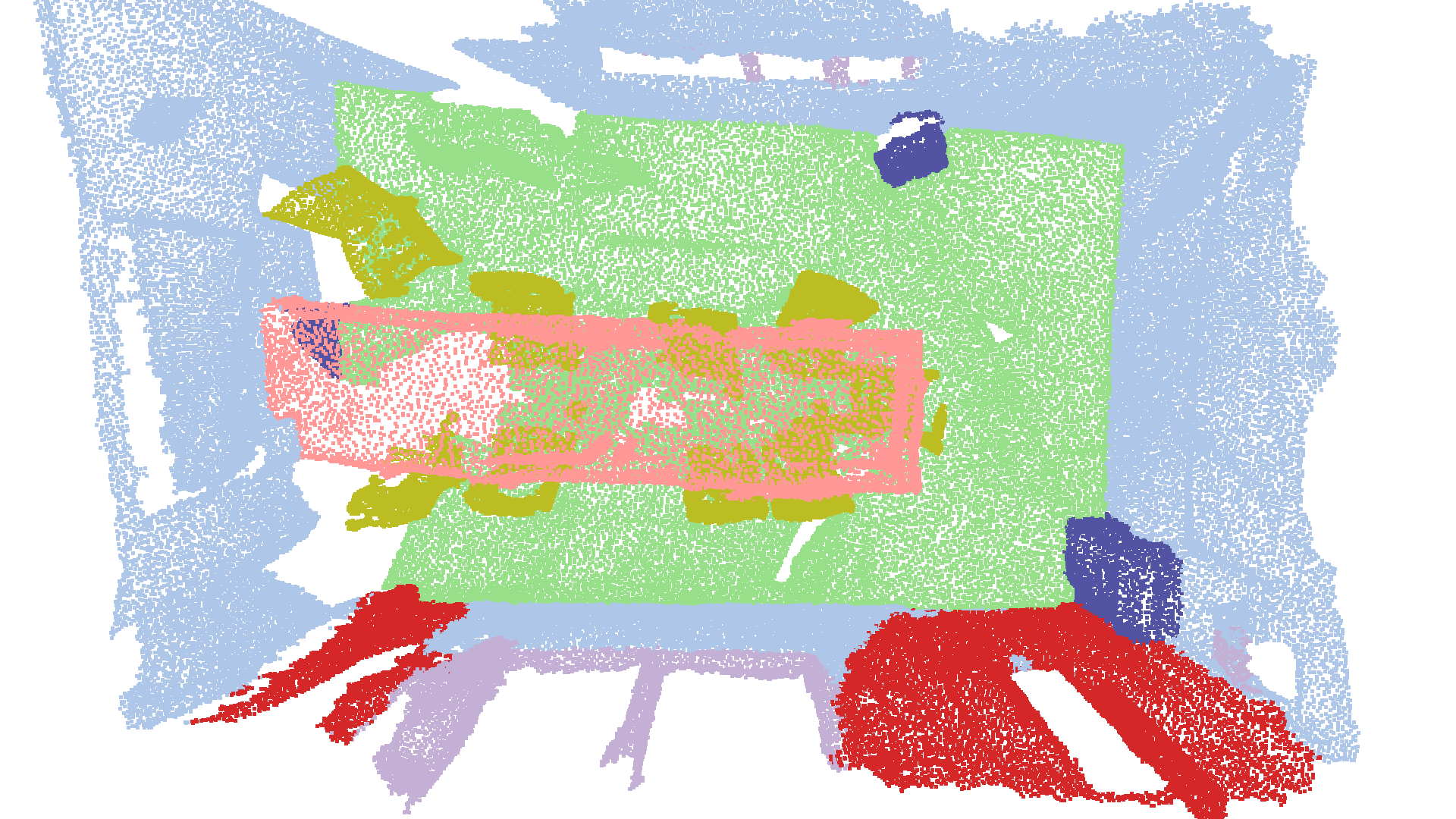}\\

            	\includegraphics[width=0.245\linewidth]{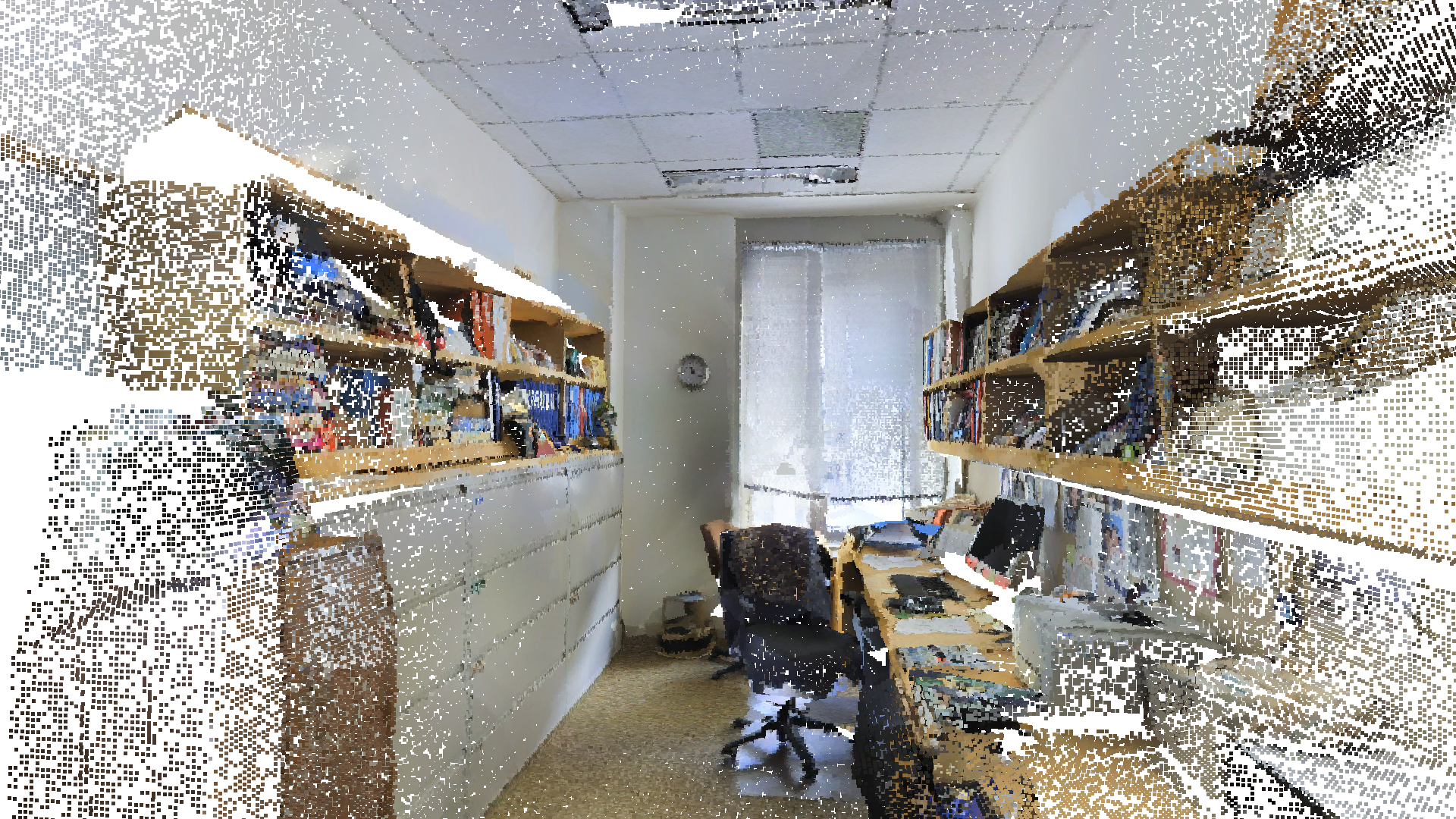}&
            	\includegraphics[width=0.245\linewidth]{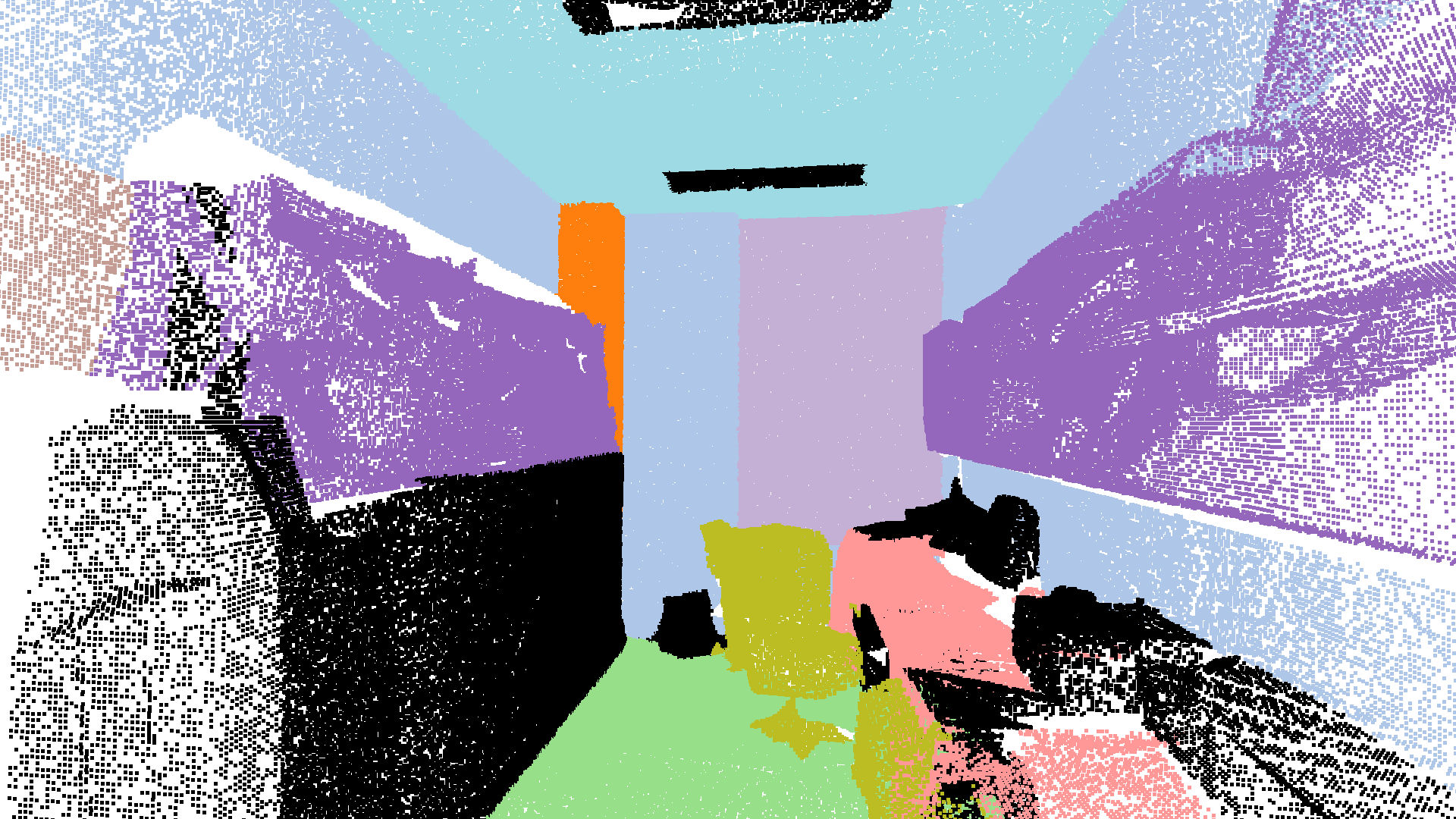}&
            	\includegraphics[width=0.245\linewidth]{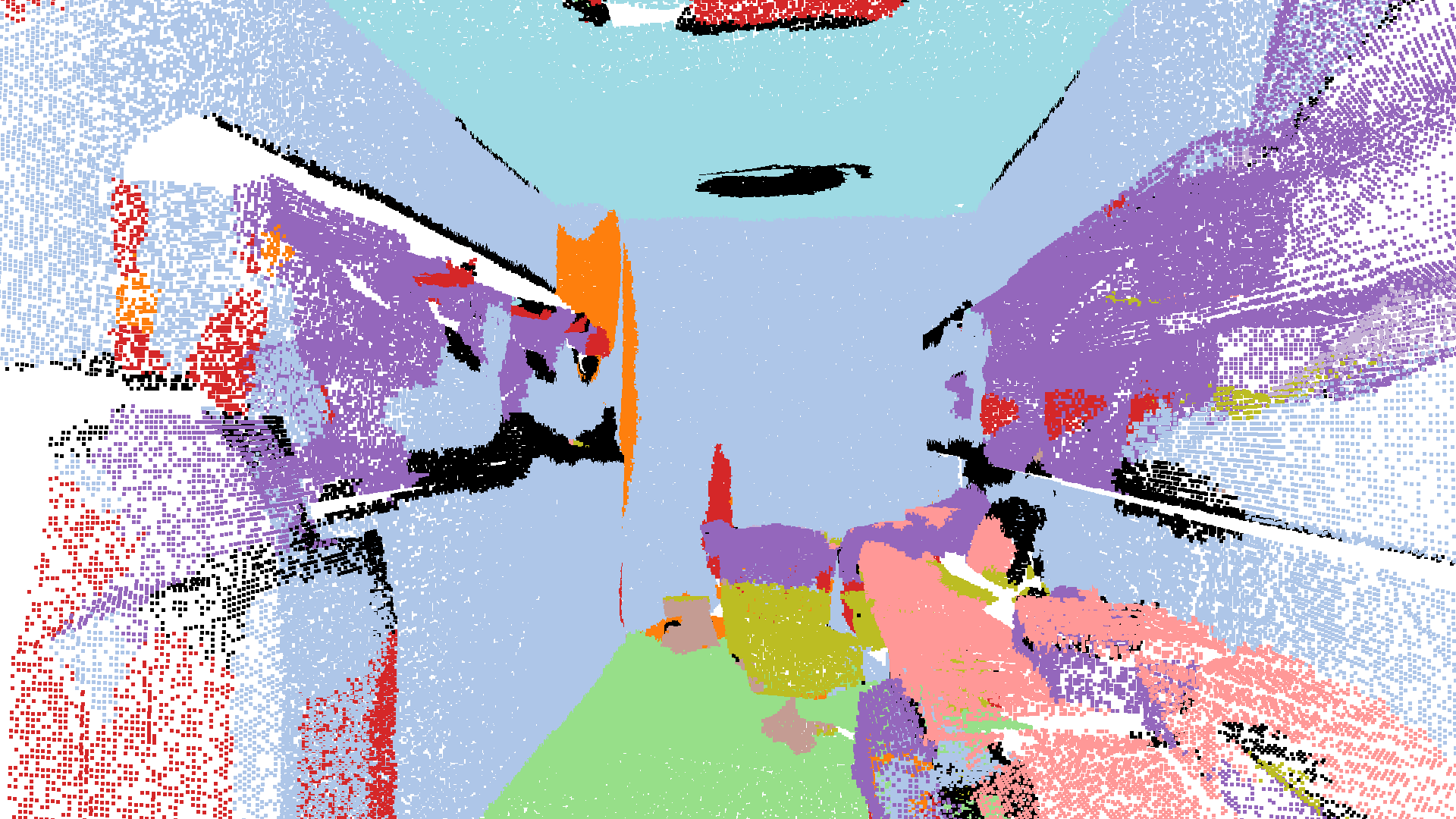}&
            	\includegraphics[width=0.245\linewidth]{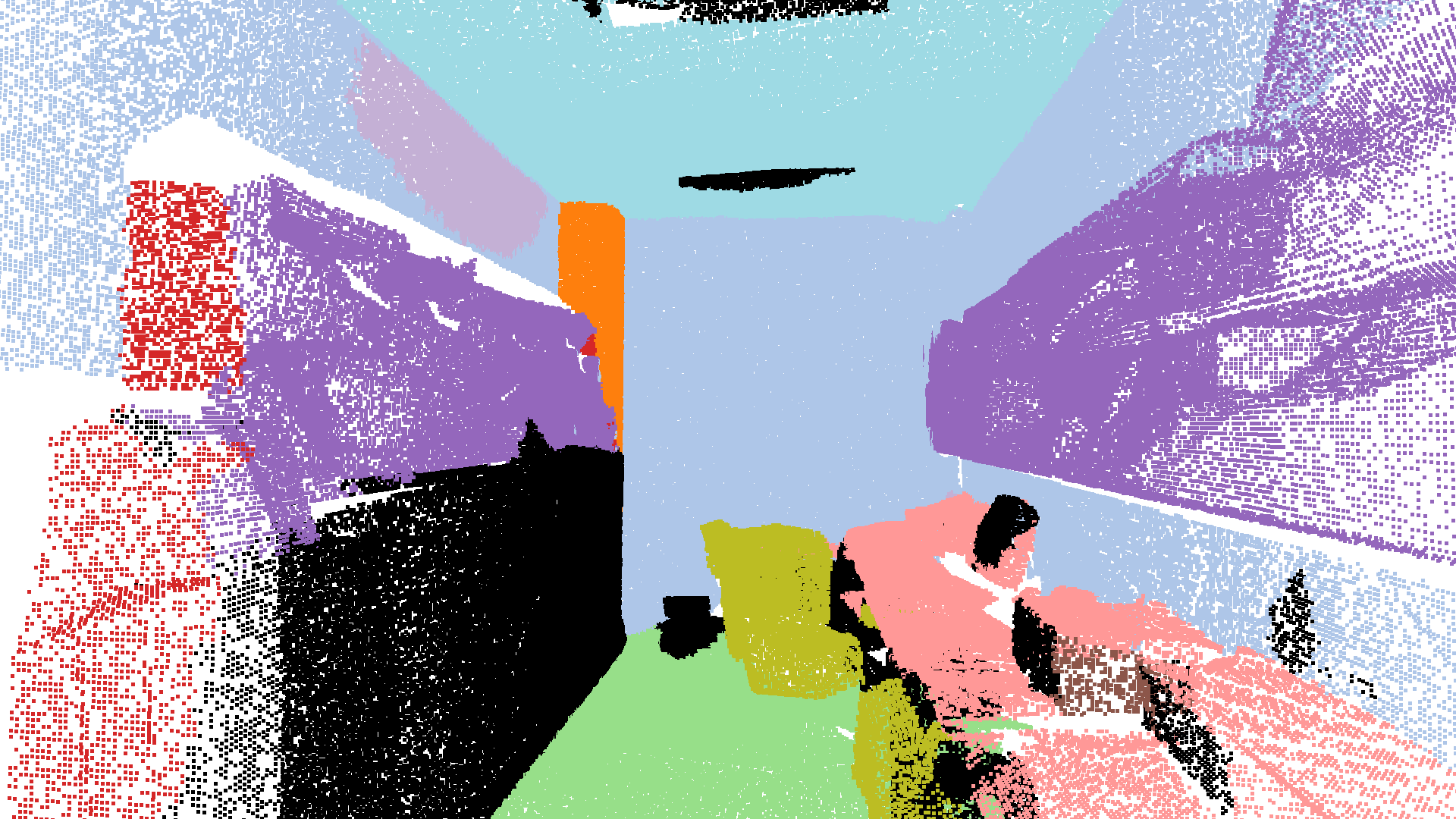}\\

            \end{tabular}
        }
        \vspace{-2mm}
	\caption{Visualization of semantic segmentation results on ScanNet v2 (first row) and S3DIS (second row) validation sets.}
	\label{fig:scannet}
	\vspace{-2mm}
\end{figure}

\paragraph{Qualitative Results.}
Fig.~\ref{fig:scannet} present the segmentation results of the baseline and our method on the ScanNet and S3DIS validation sets. As illustrated in the first row of Fig.~\ref{fig:scannet}, our method achieves more accurate segmentation than the baseline method, particularly for objects such as tables and doors. 
Furthermore, as depicted in the second row of  Fig.~\ref{fig:scannet}, our method produces significantly cleaner and regionally consistent segmentation results with fewer noisy predictions, highlighting the effectiveness of the proposed region-point consistency module.

\subsection{Ablation Studies}
Ablation experiments were conducted on the ScanNet and S3DIS datasets without Mix3D-augmentation to demonstrate the effectiveness of each component of the proposed method and analyze performance under various settings.
The performance is evaluated on the generated pseudo labels of the training set.

\begin{table}[t!]
    \begin{minipage}{.495\textwidth}
    \caption{{Ablation study on the main components. Performance is measured by the mIoU of generated pseudo-labels on the ScanNet and S3DIS training sets. CMG and RPC denote cross-modal feature guidance and region-point consistency, respectively.}
    }
    \centering
    \label{tab:main_abla}
    \vspace{-2mm}
    \resizebox{0.95\linewidth}{!}{
        \begin{tabular}{cc|cc|c|cc|cc}
            \toprule[1.0pt]
            \multicolumn{2}{c|}{CMG} & \multicolumn{2}{c|}{RPC} & \multirow{2}{*}{Mix3D} & \multicolumn{2}{c|}{ScanNet}  & \multicolumn{2}{c}{S3DIS} \\
             $\mathcal{L}_{\text{con}}^{\textsc{3d}}$ & $\mathcal{L}_{\text{con}}^{\textsc{2d}}$ & $\mathcal{L}_{\text{rpc}}^{\textsc{3d}}$ & $\mathcal{L}_{\text{rpc}}^{\textsc{2d}}$ & & 3D & 2D & 3D & 2D\\
            \midrule[0.2pt]
            & & & & & 25.1 & 15.2 & 33.9 & 10.8\\
            \midrule[0.2pt]
            \checkmark& & & & & 36.8 & 16.3 & 41.2 & 10.7\\
            \checkmark&\checkmark & & & & 36.8 & 24.3 & 40.1 & 13.9\\
             & &\checkmark &\checkmark& & 29.6 & 24.4 & 36.1 & 13.3\\
             \midrule[0.2pt]
            \checkmark&\checkmark &\checkmark &&  & 47.7 & 31.2 & 48.7 & 15.0\\
            \checkmark&\checkmark &\checkmark &\checkmark&  & 48.2 & 35.4 & 50.5 & 18.5\\
            \midrule[0.2pt]
            \checkmark&\checkmark &\checkmark &\checkmark& \checkmark & 50.1 & 35.4 & 51.6 & 18.0\\
            \bottomrule[1.0pt]
        \end{tabular}
    }
    \vspace{-2mm}
    \end{minipage}
\end{table}

\paragraph{Contributions of Main Components.}
To assess the effectiveness of each component, a baseline method based on PCAM was established, followed by testing various combinations of proposed components.
Results are detailed in Table~\ref{tab:main_abla}.
The comparison of results with and without cross-modal feature guidance (CMG) demonstrates that CMG significantly boosts performance, particularly in 3D segmentation, providing a solid basis for the region-point semantic consistency constraint.
Adding the region-point consistency (RPC) module to the baseline method without CMG also enhances performance but to a limited extent, due to the high noise level in semantic predictions.
Combining CMG and RPC results in a substantial improvement in performance, achieving an mIoU increase of 10\% in the 3D segmentation of both datasets.
These results suggest that the RPC module can further enhance performance based on improved semantic prediction.
Additionally, the use of Mix3D data augmentation provides a slight improvement in performance.
In conclusion, the results in Table~\ref{tab:main_abla} confirm the significant contribution of each method component to overall performance.

\paragraph{Analysis of Cross-modal Feature Guidance.}
To explore the effect of different distillation losses in the cross-modal feature guidance module, experiments were conducted with different forms of distillation losses from 2D to 3D and 3D to 2D.
And the results are presented on Table~\ref{tab:type}.
The ``MSE'' loss represents a direct minimization of the L2 distance between point-pixel feature pairs.
The ``Affinity'' loss minimizes the distance between affinity matrices derived from the image and the point cloud features, while the ``CAM'' loss utilizes the pseudo labels from class activation map as supervision information, which are both adopt in~\cite{kweon2022joint}.
The contrastive and affinity losses achieve comparable performance.
Compared to MSE loss, contrastive loss not only reduces the distance between corresponding points and pixel features but also increases the distance between non-corresponding points and pixel features, thereby enhancing feature representation.
Additionally, the CAM loss, as adopted in~\cite{kweon2022joint}, performs poorly due to CAM's inherent inaccuracy under scene-level annotation, leading to noisy guidance.

\paragraph{{Impact of Region Partitioning Strategy.}}
{
To assess the sensitivity of our region-point consistency module to the choice of partitioning, we conducted an ablation study comparing various strategies (Point-wise, VCCS~\cite{VCCS}, Region Growing~\cite{adams1994seeded}, VCCS+RG, WyPR~\cite{ren20213d}, and ground truth). We evaluated performance by using partitions for direct majority voting (``Partition Quality") and within the full RPC module (``Final Quality"). As shown in Table~\ref{tab:region}, RPC consistently improves results over direct voting for all partitioning methods, and regional context significantly outperforms point-wise consistency. Notably, while the quality of unsupervised partitions varies considerably, the performance gap between them narrows when using RPC. This suggests that the adaptive threshold mechanism in RPC effectively mitigates the negative impact of inaccuracies in the partitioning by filtering potentially unreliable pseudo-labels derived from inconsistent regions. Consequently, our method exhibits robustness to the specific choice of unsupervised region partitioning strategies. 
Among these, WyPR provides the best overall performance and is used by default, but other methods also achieve competitive results when combined with RPC. The remaining gap to the ground truth partition performance indicates potential for future gains with improved partitioning algorithms.
}

\paragraph{{Experiments with Different 2D and 3D Backbones.}} 
{Table~\ref{tab:backbones_2d_3d} presents the mIoU performance of our method on the ScanNet training set with different 2D and 3D backbones. 
When the 3D backbone is fixed as Mink U-Net 18A, using a larger 2D backbone significantly improves the 2D mIoU but has minimal impact on the final 3D pseudo-label mIoU. 
When the 2D backbone is fixed as ResNet50, using larger 3D backbones slightly enhances both 2D and 3D mIoU. 
However, the performance variations due to different backbone networks are less pronounced compared to the impact of the core components proposed in our method (CMG and RPC). 
This suggests that in the weakly supervised setting using only scene-level annotations, enhancing the quality and coverage of the supervisory signals is potentially more crucial for performance improvement than merely increasing the backbone network's capacity. 
}

\begin{table}[t!]
    \begin{minipage}{.495\textwidth}
        \centering
        \small
        \caption{{
        Ablation study on distillation losses for the cross-modal feature guidance module. Performance is measured by the mIoU of generated pseudo-labels on the ScanNet training set.}
        }
        \label{tab:type}
        \vspace{-2mm}
        \scalebox{0.90}{
            \begin{tabular}{cc|cc}
                \toprule
                3D-to-2D & 2D-to-3D & 3D & 2D \\
                \midrule
                - & - & 25.1 & 15.2 \\
                \midrule
                Affinity & MSE & 32.6 & 26.2 \\
                Affinity & CAM & 32.7 & 25.3 \\
                Affinity & Affinity & 34.2 & 21.1 \\
                Affinity & Contrastive & 35.5 & 22.0 \\
                Contrastive & Affinity & 35.1 & 23.9 \\
                Contrastive & Contrastive & 36.8 & 24.3 \\
                \bottomrule
            \end{tabular}
        }
    \end{minipage}
    
    \begin{minipage}{.495\textwidth}
        \color{revise}
        \centering
        \small
        \caption{
            {
                Ablation study on the impact of different region partitioning methods on pseudo-label quality (mIoU \%) on the ScanNet training set.
            }
        }
        \label{tab:region}
        \vspace{-2mm}
        \scalebox{0.9}{
            \begin{tabular}{l|c|c}
                \toprule[1pt]
                Partitioning & Partition Quality & Final Quality \\
                Method & (mIoU w/o RPC) & (mIoU w/ RPC) \\
                \midrule
                Point-wise & 36.0 & 44.5\\
                VCCS~\cite{VCCS} & 37.0 & 45.4 \\
                RG~\cite{adams1994seeded} & 39.3 & 47.9 \\
                VCCS + RG & 39.6 & 48.1 \\
                WyPR~\cite{ren20213d} & \textbf{41.5} & \textbf{48.2} \\ 
                \midrule
                GT & 66.8 & 69.2 \\
                \bottomrule[1pt]
            \end{tabular}
        }
        \vspace{2mm}
    \end{minipage}

    \begin{minipage}{.495\textwidth}
        \color{revise}
        \centering
        \small
        \caption{
        {
        The mIoU performance of pseudo labels generated on ScanNet training set with different backbone combinations. }
        }
        \label{tab:backbones_2d_3d}
        \vspace{-2mm}
        \scalebox{0.9}{
            \begin{tabular}{cc | cc}
                \toprule[1pt] 
                3D Backbone & 2D Backbone & 3D mIoU & 2D mIoU\\
                \midrule
                Mink U-Net 18A & ResNet 34 & 47.8 & 32.0\\
                Mink U-Net 18A & ResNet 50 & 48.2 & 35.4\\
                Mink U-Net 18A & ResNet 101 & \textbf{48.4} & \textbf{36.7}\\
                \midrule
                Mink U-Net 14A & ResNet 50 & 46.3 & 34.4\\
                Mink U-Net 18A & ResNet 50 & 48.2 & 35.4\\
                Mink U-Net 34C & ResNet 50 & \textbf{49.5} & \textbf{36.2}\\
                \bottomrule[1pt] 
            \end{tabular}
        }
        \vspace{2mm}
    \end{minipage}

    \vspace{-5mm}
\end{table}

\section{Conclusion}
In this paper, we present an efficient framework that leverages region-point semantic consistency and dense RGB information to improve the quality of pseudo-labels under scene-level annotation.
The cross-modal feature guidance is performed with point-wise contrastive distillation, which enhances point cloud feature representation.
Also, we develop a region-point semantic consistency regularization to align the noisy point-level semantic output to a stable and more accurate regional semantic output that is produced by region-averaging and filtering with a dynamic category threshold.
The experiment results on ScanNet v2 and S3DIS datasets demonstrate the effectiveness of our framework.

\section*{Acknowledgements}
This research is supported by the NSFC Grants under the contracts No. 62325111 and No.U22B2011. 

\ifCLASSOPTIONcaptionsoff
  \newpage
\fi

\bibliographystyle{IEEEtran}
\bibliography{main}

\appendices
\section{More Ablation Studies}
\paragraph{Analysis of Various 2D Image Views.} Experiments were conducted with a varying number of 2D image views.
Results in Table~\ref{tab:view_numbers} indicate that cross-modal feature guidance module achieves optimal 3D segmentation performance with 6 2D image views.
It is hypothesized that an excessive number of image views might hinder the point cloud branch's learning from weak labels, whereas an insufficient number of views may fail to provide adequate cross-modal guidance.

\paragraph{Analysis of Pre-reained 2D Backbones.} Experiments were then conducted with various 2D image backbones, including ResNet50 pre-trained on ImageNet~\cite{deng2009imagenet} with supervised learning and unsupervised learning with MoCo~\cite{he2020momentum} and DINO~\cite{caron2021emerging}, ResNet 50 without pre-trained weight, and frozen ResNet 50 with pre-trained weight.
Results of the first four columns in Table~\ref{tab:backbones} reveal that various 2D backbones with pre-trained weights achieve comparable performance, while the 2D backbone without pre-trained weights achieves the worst performance. 
This comparison shows that both supervised and unsupervised pre-training 2D networks can greatly improve the performance of weakly supervised semantic segmentation in 3D and 2D, owing to the introduction of additional cross-modal knowledge from pre-trained 2D networks.
In previous works on cross-modal contrastive distillation, such as PPKT~\cite{liu2021learning} and SLidR~\cite{sautier2022image}, the pre-trained 2D backbones are frozen to only provide feature guidance. We also conducted experiment with such setting. 
However, as shown in the last column of Table~\ref{tab:backbones}, the frozen 2D backbone underperforms the trained 2D backbone (first column) by 5.9\% mIoU, indicating the significance of 2D network learning from scene-level annotations and 3D features.

\paragraph{Analysis of Region-point Semantic Consistency.} To explore the impact of threshold $\eta$ in the region-point semantic consistency module, experiments were conducted using different $\eta$ values with three consistency constraints: ``Point-Point Consistency with Fixed threshold'', ``Region-Point Consistency with Fixed threshold'', and ``Region-Point Consistency with Adaptive threshold''.
Results with $\eta$ values ranging from 0.0 to 0.9 are presented in Table~\ref{tab:thresholds}.
Results of the first two rows reveal that region-point semantic consistency surpasses point-point semantic consistency across various fixed threshold.
And results of the last two rows reveal that region-point semantic consistency with an adaptive threshold uniformly exceeds region-point semantic consistency with a fixed threshold.
The comparison of first group confirms the effectiveness of the proposed region-point semantic consistency, whereas the comparison of second group underscores the significance of an adaptive threshold for various categories.

\begin{table}[t!]
    \begin{minipage}{.495\textwidth}
        \centering
        \small
        \caption{
        {
        Ablation study on the number of 2D image views for the cross-modal feature guidance module. Performance is measured by the mIoU of generated pseudo-labels on the ScanNet training set. }
        }
        \label{tab:view_numbers}
        \vspace{-2mm}
        \scalebox{0.95}{
            \begin{tabular}{c|ccccc}
                \toprule[1pt] 
                Number of views & 0 & 3 & 6 & 9 & 12  \\
                \midrule
                3D mIoU & 25.1 & 32.8 & 36.8 & 35.9 & 33.1  \\
                2D mIoU & 15.2 & 22.6 & 24.3 & 23.8 & 24.6  \\
                \bottomrule[1pt]
            \end{tabular}
        }
        \vspace{4mm}
    \end{minipage}
    \begin{minipage}{.495\textwidth}
        \centering
        \small
        \caption{
        {
        Ablation study on 2D backbones for the cross-modal feature guidance module. Performance is measured by the mIoU of generated pseudo-labels on the ScanNet training set. R.50 and R.50 w/o denote ResNet50 with and without ImageNet pre-trained weights, respectively.}
        }
        \label{tab:backbones}
        \vspace{-2mm}
        \scalebox{0.95}{
            \begin{tabular}{c|ccccc}
                \toprule[1pt] 
                Backbone & R.50 & DINO & MoCo & R.50 w/o & Frozen R.50\\
                \midrule
                3D mIoU & 48.2 & 49.9 & 46.5 & 40.0 & 42.3\\
                2D mIoU & 35.4 & 31.6 & 34.2 & 31.7 & -\\
                \bottomrule[1pt]
            \end{tabular}
        }
        \vspace{4mm}
    \end{minipage}
    \begin{minipage}{.495\textwidth}
        \centering
        \small
        \caption{
        {
        Ablation study on different consistency strategies and thresholds. Performance is measured by the mIoU of generated pseudo-labels on the ScanNet training set. ``P w/ F'', ``R w/ F'', and ``R w/ A'' denote ``Point-Point Consistency with Fixed Threshold'', ``Region-Point Consistency with Fixed Threshold'', and ``Region-Point Consistency with Adaptive Threshold'', respectively.}
        }
        \label{tab:thresholds}
        \vspace{-2mm}
        \scalebox{0.80}{
            \begin{tabular}{c|ccccccccccc}
                \toprule[1pt] 
                $\eta$ & 0.0 & 0.1 & 0.2 & 0.3 & 0.4 & 0.5 & 0.6 & 0.7 & 0.8 & 0.9 \\
                \midrule
                P w/ F & 42.4 & 42.0 & 42.4 & 42.4 & 42.7 & 43.8 & 43.9 & 44.3 & 44.5 & 44.5 \\
                R w/ F & 45.9 & 45.8 & 44.1 & 44.5 & 44.9 & 45.9 & 45.9 & 46.3 & 46.6 & 45.4 \\
                R w/ A & 45.9 & 46.2 & 44.9 & 45.3 & 45.7 & 47.6 & 47.3 & 47.6 & 48.2 & 47.5 \\
                \bottomrule[1pt]
            \end{tabular}
        }
    \end{minipage}
\end{table}

\begin{table*}[t!]
    \begin{minipage}{.995\textwidth}
        \color{revise}
        \centering
        \small
        \caption{{Statistics of per-category point proportions (\%) and scene occurrences (out of 28,130 scenes) in the nuScenes training set. }
        }
        \label{tab:nuscenes_sta}
        \vspace{-2mm}
        \scalebox{0.99}{
            \begin{tabular}{l|cccccccccccccccc}
                \toprule
                Statistic & \rotatebox{90}{barrier} & \rotatebox{90}{bicycle} & \rotatebox{90}{bus} & \rotatebox{90}{car} & \rotatebox{90}{construction} & \rotatebox{90}{motorcycle} & \rotatebox{90}{pedestrian} & \rotatebox{90}{traffic cone} & \rotatebox{90}{trailer} & \rotatebox{90}{truck} & \rotatebox{90}{drivable} & \rotatebox{90}{other flat} & \rotatebox{90}{sidewalk} & \rotatebox{90}{terrain} & \rotatebox{90}{manmade} & \rotatebox{90}{vegetation} \\
                \midrule
                Point Prop. & 1.11 & 0.02 & 0.54 & 4.60 & 0.19 & 0.05 & 0.27 & 0.09 & 0.65 & 1.91 & 37.57 & 0.95 & 8.29 & 8.16 & 21.10 & 14.49 \\
                Scene Occur. & 10k & 6k & 9k & 27k & 8k & 6k & 23k & 12k & 8k & 20k & 28k & 12k & 28k & 24k & 28k & 27k \\ 
                \bottomrule
            \end{tabular}
        }
    \end{minipage}
\end{table*}

\section{Additional Experimental Results}
{
This section presents additional experimental results to further validate our method's applicability and contextualize its performance. 
Noting that prior works in scene-level weakly supervised segmentation typically focused on indoor benchmarks, we first explore applying this setting to raw outdoor datasets and identify the associated challenges. 
Subsequently, we investigate integrating our framework with powerful pre-trained 2D foundation models. 
This integration not only proves effective in addressing the aforementioned outdoor dataset challenges but also provides comparative analyses against state-of-the-art 3D Open-Vocabulary segmentation methods and strong projection-based baselines utilizing these 2D foundation models, which further demonstrate the effectiveness of our method.
}

\paragraph{{Results and Challenges on Outdoor Datasets.}} 
{Applying baseline method using, which uses only scene-level tags (derived from object presence/absence) to outdoor datasets like nuScenes~\cite{caesar2020nuscenes} and SemanticKITTI~\cite{behley2019semantickitti} proved challenging, achieving a mIoU around 10\% as shown in Table~\ref{tab:outdoor_ablation}.
We identified two primary reasons for this difficulty, illustrated by statistics in Table~\ref{tab:nuscenes_sta} from the nuScenes dataset:
\begin{itemize}
    \item[1.]  \textbf{Extreme Class Imbalance:} Outdoor scenes exhibit extreme imbalance in point distribution across categories. For instance, `drivable surface', `manmade', and `vegetation' constitute over 70\% of all points, while rare classes like `bicycle' or `motorcycle' represent less than 0.1\%. This makes it difficult for scene-level labels (which indicate presence/absence) to provide meaningful supervision for rare classes during pseudo-label generation.
    \item[2.]  \textbf{High Scene Label Co-occurrence:} Many common outdoor categories (e.g., `car', `drivable surface', `sidewalk', `manmade', `vegetation') co-occur in almost every scene (present in more than 95\% of 28,130 scenes). This high degree of co-occurrence makes the scene-level labels less discriminative for distinguishing points belonging to these common classes based solely on scene tags.
\end{itemize}
To achieve meaningful results, we followed recent trends~\cite{3dvlg} and utilized initial pseudo-labels generated by projecting segmentation maps from pre-trained 2D Open-Vocabulary Semantic Segmentation (OVSS) models (OpenSeg~\cite{ghiasi2022scaling} and LSeg~\cite{li2022languagedriven}). 
We then applied our proposed method to improve these initial labels.
Table~\ref{tab:outdoor_ablation} presents the final segmentation performance on the validation sets.
The results show that simply projecting OVSS labels provides a strong baseline. Applying our modules consistently improves performance on both nuScenes and SemanticKITTI across different OVSS initializations. This demonstrates the effectiveness of our approach in improving noisy pseudo-labels, even in challenging outdoor environments when provided with richer initial supervision beyond just scene tags. }

\begin{table}[t!]
    \begin{minipage}{.495\textwidth}
        \color{revise}
        \centering
        \small
        \setlength{\tabcolsep}{3.5pt}
        \caption{
        {
        Performance comparison (mIoU \%) on nuScenes and SemanticKITTI validation sets. ``Baseline" uses only scene labels. ``OpenSeg/LSeg" uses initial pseudo-labels from projected 2D OVSS maps. 
        }
        }
        \label{tab:outdoor_ablation}
        \vspace{-2mm}
        \scalebox{0.9}{
            \begin{tabular}{l|c|cc}
                \toprule[1pt] 
                    Method & OVSS & nuScenes  &  SemanticKITTI  \\
                \midrule
                    Baseline & - & 10.3 & 9.8 \\
                \midrule
                    Baseline & OpenSeg & 41.9 & 42.3 \\
                    Ours & OpenSeg & \textbf{47.8} & \textbf{48.2} \\
                \midrule
                    Baseline & LSeg & 36.8 & 43.7 \\
                    Ours &  LSeg & \textbf{41.4} & \textbf{49.5} \\
                \bottomrule[1pt]
            \end{tabular}
        }
        \vspace{2mm}
    \end{minipage}
    
    \begin{minipage}{.495\textwidth}
        \color{revise}
        \centering
        \small
        \setlength{\tabcolsep}{3.5pt}
        \caption{
        {
        Performance comparison (mIoU \%) with Open-Vocabulary (OV) and related Weakly-Supervised (WS) methods on ScanNet (SC), nuScenes (NS), and SemanticKITTI (SK) validation sets.
        }
        }
        \label{tab:ovss_sc_nu_sk}
        \vspace{-2mm}
        \scalebox{0.95}{
            \begin{tabular}{l|c c| ccc}
                \toprule[1pt]
                Method & Setting &  OVSS & SC & NS & SK\\
                \midrule
                \multicolumn{5}{l}{\textit{Fully-Supervised 3D Segmentation}} \\
                \midrule
                Mink UNet [CVPR'19]~\cite{choy20194d}  & Full & - & 72.2 & 73.3 & 63.8 \\
                \midrule
                \multicolumn{5}{l}{\textit{Open-Vocabulary 3D Segmentation}} \\
                \midrule
                OpenScene [CVPR'23]~\cite{peng2023openscene} & OV & OpenSeg & 47.5 & 42.1 & - \\
                OpenScene [CVPR'23]~\cite{peng2023openscene}  & OV & LSeg & 54.2 & 36.7 & - \\
                GGSD [CVPR'24]~\cite{GGSD}  & OV & - & 56.5 & 46.1 & -\\
                OVD [CVPR'24]~\cite{ov3d}  & OV & - & 57.3 & 44.6 & -\\
                \midrule
                \multicolumn{5}{l}{\textit{Weakly-Supervised 3D Segmentation (using OV priors)}} \\
                \midrule
                3DSS-VLG [ECCV'24]~\cite{3dvlg} & WS & OpenSeg & 49.7 & - & -\\
                Baseline & WS & OpenSeg & 47.1 & 41.9 & 42.3 \\ 
                \textbf{Ours} & WS & OpenSeg & \textbf{51.6} & \textbf{47.8} & \textbf{48.2}\\ 
                Baseline & WS & LSeg & 54.1 & 36.8 & 43.7\\ 
                \textbf{Ours} & WS & LSeg & \textbf{58.7} & \textbf{41.4} & \textbf{49.5}\\ 
                \bottomrule[1pt]
            \end{tabular}
        }
        \vspace{2mm}
    \end{minipage}
    
    \begin{minipage}{.495\textwidth}
        \color{revise}
        \centering
        \begin{small}
        \caption{
        {Comparison of mIoU (\%) on ScanNet validation set using OVSS models and SAM.}
        }
        \label{tab:sam_ovss}
        \vspace{-2mm}
        \resizebox{0.8\textwidth}{!}{
            \begin{tabular}{l | cc | c}
                \toprule[1pt]
                Method & OVSS & SAM & ScanNet mIoU\\
                \midrule
                Baseline & OpenSeg & & 47.1 \\
                Baseline & OpenSeg & \checkmark & 48.5 \\ 
                Ours     & OpenSeg & & 51.6 \\
                Ours     & OpenSeg & \checkmark & \textbf{52.4} \\ 
                \midrule
                Baseline & LSeg & &  54.1 \\
                Baseline & LSeg & \checkmark & 55.8 \\ 
                Ours & LSeg & & 58.7 \\
                Ours & LSeg & \checkmark & \textbf{59.6} \\ 
                \bottomrule[1pt]
            \end{tabular}
        }
        \end{small}
    \end{minipage}
    \vspace{-4mm}
\end{table}

\paragraph{{Comparison with Open-Vocabulary Methods.}}
{To contextualize our approach within the broader landscape, we compare it with recent 3D open-vocabulary (OV) segmentation methods on indoor reconstructed point cloud and outdoor LiDAR point cloud. 
Specifically, following previous 3D OV methods, we leverage 2D OV semantic segmentation models (e.g., OpenSeg~\cite{ghiasi2022scaling} and LSeg~\cite{li2022languagedriven}) to obtain 2D image pseudo-labels and then project these pseudo-labels to 3D point cloud based on the pixel-point correspondence. 
Table~\ref{tab:ovss_sc_nu_sk} compares our results against state-of-the-art OV methods and a related weakly supervised method using OV priors (3DSS-VLG~\cite{3dvlg}).
Our method achieves highly competitive performance, surpassing several dedicated OV methods on both ScanNet and nuScenes validation sets. 
Notably, our method significantly outperforms the baseline only with projected OVSS pseudo-labels, demonstrating the value added by our proposed modules. 
Our method with OpenSeg (51.6\%) also surpasses 3DSS-VLG (49.7\%) on ScanNet with the same 2D OVSS model under the same setting.
This highlights the effectiveness of our framework in leveraging and improving upon strong OV priors within a weakly supervised refinement paradigm. }

\paragraph{{Comparison with SAM-based Projection.}}
{We also compare our method against a strong baseline that leverages the Segment Anything Model (SAM)~\cite{kirillov2023segment} combined with 2D OVSS models (OpenSeg/LSeg) for initial pseudo-label generation. Specifically, the pseudo-labels from OVSS models are refined by SAM's class-agnostic masks before projection to 3D point cloud. Table~\ref{tab:sam_ovss} presents the results on ScanNet validation set.
While using SAM provides a moderate improvement over the basic OVSS projection baseline (+1.4\% / +1.7\% mIoU), our method applied to the original OVSS projections significantly outperforms this SAM-enhanced baseline (+3.1\% / +2.9\% mIoU). 
Applying our refinement on top of the SAM-enhanced pseudo-labels yields the best results, indicating that our approach complements SAM-based improvements and provides substantial gains through the proposed modules.}

{
These additional experiments highlight both the challenges inherent in applying scene-level weak supervision to diverse datasets and the significant potential unlocked by integrating priors from modern 2D foundation models, suggesting that future advancements in weakly supervised point cloud segmentation will likely benefit significantly from integration with large pre-trained foundation models, representing a promising direction for future research.
}

\section{Per-class Quantitative Results}
In Table 1 of the main paper, the mean IoU results for the ScanNet and S3DIS datasets are presented. 
Here, detailed IoU results for each category are provided. 
Tables~\ref{tab:scannet_per_cls},~\ref{tab:scannet_test}, and~\ref{tab:s3dis_per_cls} present the per-class IoU results of our method alongside those of several comparative methods on the ScanNet validation set, ScanNet test set, and S3DIS Area 5 set, respectively. 
The per-class IoU results for comparative methods are sourced directly from their respective papers.
It is worth noting that our method outperforms existing methods in 13 out of 20 categories on the ScanNet validation set and 10 out of 13 categories on the S3DIS Area 5 set.
Moreover, on the ScanNet test set, our method outperforms the MIT method in 17 out of 20 categories.

\paragraph{{Analysis of Challenging Categories.}} 
{However, our method, relying solely on scene-level tags, struggles with certain categories, such as `picture' and `counter'.
Analysis of the confusion matrix (Table~\ref{tab:scannet_confusion}) reveals that `picture' points are often misclassified as `wall', and `counter' points as `cabinet' due to factors like spatial co-occurrence or semantic ambiguity. 
However, when integrating priors from 2D Open Vocabulary Semantic Segmentation models (OpenSeg and LSeg) and improving the initial pseudo-labels from 2D OVSS models with our method, performance on these challenging categories improves dramatically. Table~\ref{tab:scannet_train_pseudo} compares the per-category pseudo-label IoU on the ScanNet training set. Notably, the IoU for `picture' increases from 0.3\% to $\sim$28\%, and for `counter' from 0.0\% to $\sim$49\% when integrating OVSS priors into our method. This highlights the ability of our framework to leverage and enhance stronger initial supervision signals to overcome the limitations of scene tags for ambiguous classes.}

\begin{table*}[h!]
    \begin{minipage}{.995\textwidth}
        \centering
        \small
        \caption{Quantitative results (mIoU) of several point-cloud segmentation methods with scene-level supervision setting on the ScanNet validation set. ``B.S.'' denotes bookshelf; ``S.C.'' stands for shower curtain and ``cnt'' denotes counter. 
        }
        \label{tab:scannet_per_cls}
        \vspace{-2mm}
        \scalebox{0.9}{
        \setlength{\tabcolsep}{1pt}
            \begin{tabular}{l | c c c c c c c c c c c c c c c c c c c c | c}
                \toprule
                Method & wall & floor & cabinet & bed & chair & sofa & table & door & window & B.S. & picture & cnt & desk & curtain & fridge & S.C. & toilet & sink & bathtub & other & mIoU\\
                
                \midrule
                MIL-trans & 52.1 & 50.6 & 8.3 & 46.3 & 27.9 & 39.7 & 20.9 & 15.8 & 26.8 & 40.2 & 8.1 & 21.1 & 22.0 & 45.9 & 4.5 & 16.6 & 15.2 & 32.4 & 21.2 & 8.0 & 26.2\\   
                
                WYPR & 52.0 & 77.1 & 6.6 & 54.3 & 35.2 & 40.9 & 29.6 & 9.3 & 28.7 & 33.3 & 4.8 & 26.6 & 27.9 & 69.4 & 8.1 & 27.9 & 24.1 & 25.4 & 32.3 & 8.7 & 31.1 \\        
                
                MIT & 57.3 & 89.7 & 24.1 & 54.9 & 31.5 & 62.8 & 42.5 & 19.8 & 27.4 & 45.1 & 1.1 & 31.4 & 41.7 & 41.4 & 17.6 & 25.0 & 34.5 & 8.3 & 44.4 & 15.6 & 35.8 \\
            
                Xia et al. & 58.7 & 87.3 & 13.3 & 60.2 & 52.3 & 61.9&36.2 & 12.3 & 22.4 & 72.1& 1.5 & 5.7  &36.4& 53.3 & 7.3 & 38.6 & 61.0 & 12.1 &  53.5 & 16.2& 38.1 \\
            
                Ours & 56.5 & 93.5 & 24.6 & 70.4 & 77.5 & 70.7 & 50.3 & 22.3 & 34.3 & 69.8& 0.0 & 0.0  &36.7& 51.3 & 1.0 & 54.2 & 87.4 & 34.0 &  79.2 & 25.4 & 46.9 \\
                \bottomrule
            \end{tabular}
        }
        \vspace{2mm}
    \end{minipage}
    \begin{minipage}{.995\textwidth}
        \centering
        \small
        \caption{Quantitative results (mIoU) of our method and MIT with scene-level supervision setting on the test set from official ScanNet benchmark server. ``B.S.'' denotes bookshelf; ``S.C.'' stands for shower curtain and ``cnt'' denotes counter.  
        }
        \label{tab:scannet_test}
        \vspace{-2mm}
        \scalebox{0.9}{
        \setlength{\tabcolsep}{1pt}
            \begin{tabular}{l | c c c c c c c c c c c c c c c c c c c c | c}
                \toprule
                Method & wall & floor & cabinet & bed & chair & sofa & table & door & window & B.S. & picture & cnt & desk & curtain & fridge & S.C. & toilet & sink & bathtub & other & mIoU\\
                \midrule
                MIT & 42.2 & 82.1 & 16.3 & 55.8 & 30.6 & 57.6 & 35.9 & 19.3 & 27.0 & 39.0 & 1.4 & 25.3 & 27.7 & 31.3 & 21.3 & 17.8 & 47.8 & 7.9 & 29.8 & 18.8 & 31.7 \\
                
                Ours & 52.2 & 93.7 & 23.5 & 72.3 & 70.9 & 65.0 & 41.4 & 22.4 & 42.2 & 83.6 & 0.0 & 0.0 & 38.4 & 66.8 & 1.4 & 47.1 & 87.2 & 28.1 & 83.6 & 27.2 & 46.8 \\
            
                \bottomrule
            \end{tabular}
        }
        \vspace{2mm}
    \end{minipage}
    \begin{minipage}{.995\textwidth}
        \centering
        \small
        \caption{Quantitative results (mIoU) of several point-cloud segmentation methods with scene-level supervision setting on the S3DIS Area 5 dataset.
        }
        \label{tab:s3dis_per_cls}
        \vspace{-2mm}
        \scalebox{0.9}{
        \setlength{\tabcolsep}{2pt}
            \begin{tabular}{l | c c c c c c c c c c c c c | c}
                \toprule
                Method& ceiling & floor & wall & beam & column & window & door & table & chair & bookcase & sofa & board & clutter & mIoU\\
                
                \midrule
                MIL-Trans & 24.9 & 4.7 & 40.0 & 0.0 & 1.3 & 2.2 & 1.8 & 16.8 & 5.6 & 33.0 & 32.1 & 0.1 & 5.8 & 12.9 \\
            
                Xia et al. &79.1&86.6&51.8&0.0& 0.3& 0.5& 7.6& 30.6& 26.4& 5.6& 45.5& 0.0 & 0.7 &25.7 \\
            
                MIT & 80.8 & 81.0 & 81.8 & 0.0 & 0.9 & 0.2 & 27.6  & 19.5 & 26.7 & 15.5 & 16.8 & 0.0 & 9.9 & 27.7\\
            
                Ours & 82.3 & 95.8 & 72.4 & 0.0 & 16.1 & 4.5 & 48.4 & 58.4 & 80.4 & 69.0 & 57.7 & 0.0 & 31.8 & 47.4\\
                \bottomrule
            \end{tabular}
        }
        \vspace{2mm}
    \end{minipage}
    \begin{minipage}{.995\textwidth}
        \color{revise}
        \centering
        \small
        \caption{{Per-category pseudo-label mIoU (\%) on the ScanNet training set for our method and our method with OVSS priors. }
        }
        \label{tab:scannet_train_pseudo}
        \vspace{-2mm}
        \resizebox{0.995\textwidth}{!}{
            \setlength{\tabcolsep}{1.5pt}
            \begin{tabular}{l| c c c c c c c c c c c c c c c c c c c c | c}
                \toprule
                Method & wall & floor & cabinet & bed & chair & sofa & table & door & window & B.S. & picture & cnt & desk & curtain & fridge & S.C. & toilet & sink & bathtub & other & mIoU\\
                \midrule
                Ours & 51.4 & 93.1 & 28.9 & 69.3 & 73.2 & 72.3 & 50.4 & 27.1 & 39.6 & 70.7 & 0.3 & 0.0 & 48.3 & 64.8 & 2.5 & 68.8 & 66.1 & 32.3 & 72.0 & 31.4 & 48.2 \\
                Ours $+$ OpenSeg & 65.4 & 89.8 & 45.9 & 80.4 & 64.7 & 67.6 & 49.4 & 39.0 & 57.3 & 74.1 & 28.4 & 48.0 & 44.9 & 71.3 & 59.2 & 73.1 & 69.0 & 29.9 & 84.6 & 18.7 & 58.0 \\
                Ours $+$ LSeg & 77.3 & 92.9 & 55.9 & 83.5 & 80.5 & 73.9 & 65.7 & 53.8 & 65.4 & 75.6 & 26.7 & 49.2 & 54.7 & 77.1 & 61.4 & 63.3 & 82.4 & 52.1 & 82.4 & 26.9 & 65.0 \\
                \bottomrule
            \end{tabular}
        }
        \vspace{2mm}
    \end{minipage}
    \begin{minipage}{.995\textwidth}
        \color{revise}
        \centering
        \begin{small}
        \caption{{Confusion matrix (rows: GT label, columns: Predicted pseudo-label) on ScanNet training set. Numbers indicate point counts. Gray cells highlight major confusion sources for `picture' and `counter'.}
        }
        \label{tab:scannet_confusion}
        \vspace{-2mm}
        \resizebox{0.995\textwidth}{!}{
        \setlength{\tabcolsep}{1.5pt}
            \begin{tabular}{l|rrrrrrrrrrrrrrrrrrrr}
                \toprule
                 & wall & floor & cabinet & bed & chair & sofa & table & door & window & B.S. & picture & cnt & desk & curtain & fridge & S.C. & toilet & sink & bathtub & other \\
                \midrule
                wall & 23029577 & 343151 & 383727 & 144286 & 475063 & 114620 & 426934 & 10147332 & 3523436 & 249416 & 7691 & 235 & 767031 & 228313 & 874089 & 40755 & 47872 & 163094 & 17284 & 838190 \\
                floor & 78506 & 30943895 & 74422 & 3250 & 104192 & 39793 & 224369 & 136835 & 8776 & 23833 & 18378 & 0 & 109227 & 11246 & 22 & 299 & 18495 & 0 & 1587 & 132819 \\
                cabinet & 269395 & 53844 & 2094052 & 4645 & 28142 & 8873 & 52521 & 687971 & 322723 & 186185 & 12743 & 25 & 204694 & 7355 & 864302 & 0 & 1422 & 169538 & 0 & 678361 \\
                bed & 101280 & 37078 & 4669 & 2732612 & 4931 & 18664 & 8297 & 16215 & 34565 & 2734 & 474184 & 0 & 178988 & 2737 & 27591 & 0 & 0 & 0 & 0 & 117935 \\
                chair & 50878 & 238493 & 517 & 1389 & 8401340 & 73772 & 988381 & 21973 & 29975 & 15316 & 614 & 0 & 81415 & 4594 & 54 & 0 & 0 & 0 & 0 & 103327 \\
                sofa & 13276 & 45717 & 211 & 1908 & 84813 & 2780438 & 397725 & 462 & 4038 & 1652 & 2680 & 0 & 7937 & 1053 & 22 & 0 & 0 & 0 & 0 & 59217 \\
                table & 32704 & 215710 & 20795 & 1200 & 247019 & 74270 & 3815358 & 11081 & 26323 & 622 & 0 & 0 & 194661 & 2167 & 730 & 0 & 44 & 1140 & 0 & 277243 \\
                door & 757560 & 62121 & 19843 & 1499 & 29503 & 268 & 24739 & 5050282 & 225123 & 13187 & 0 & 0 & 3330 & 65791 & 3132 & 2847 & 1680 & 956 & 918 & 32147 \\
                window & 594696 & 57947 & 8342 & 1470 & 69086 & 27152 & 73097 & 290414 & 4194015 & 6677 & 82 & 0 & 21696 & 34930 & 29135 & 0 & 0 & 2423 & 0 & 14885 \\
                B.S. & 149591 & 35719 & 21367 & 675 & 20783 & 154 & 29789 & 22012 & 155723 & 2741209 & 0 & 0 & 54713 & 60 & 9 & 0 & 0 & 11 & 0 & 61019 \\
                picture & \cellcolor{black!20}{546814} & 288 & 1541 & 352 & 4379 & 550 & 83 & 54527 & 41229 & 4241 & 3771 & 0 & 2367 & 705 & 3541 & 0 & 0 & 236 & 0 & 13753 \\
                cnt & 12379 & 352 & \cellcolor{black!20}{470707} & 12 & 894 & 313 & 9420 & 362 & 11898 & 0 & 0 & 14 & 3039 & 0 & 44831 & 0 & 0 & 84470 & 0 & 28961 \\
                desk & 21552 & 48345 & 6963 & 5123 & 89333 & 503 & 49901 & 3923 & 54202 & 620 & 225 & 0 & 2178457 & 852 & 0 & 0 & 0 & 0 & 0 & 215492 \\
                curtain & 182678 & 16273 & 558 & 5240 & 20882 & 8192 & 9870 & 134314 & 345791 & 4126 & 143 & 0 & 11223 & 2194482 & 45128 & 10802 & 419 & 15 & 562 & 21458 \\
                fridge & 12520 & 6766 & 329070 & 0 & 550 & 970 & 1688 & 183337 & 6975 & 0 & 0 & 0 & 10415 & 0 & 66185 & 0 & 0 & 350 & 0 & 103048 \\
                S.C. & 5 & 489 & 18 & 0 & 0 & 0 & 0 & 83051 & 0 & 0 & 0 & 0 & 0 & 200 & 0 & 341515 & 4799 & 15 & 5857 & 1561 \\
                toilet & 646 & 12401 & 101 & 0 & 504 & 0 & 0 & 11315 & 135 & 0 & 0 & 0 & 0 & 0 & 0 & 0 & 392503 & 80 & 0 & 17891 \\
                sink & 1326 & 520 & 55773 & 0 & 5522 & 0 & 620 & 354 & 81 & 0 & 0 & 5 & 0 & 0 & 3351 & 0 & 220 & 256255 & 0 & 35077 \\
                bathtub & 4809 & 1226 & 82 & 0 & 0 & 0 & 0 & 19288 & 4024 & 0 & 0 & 0 & 0 & 0 & 0 & 4135 & 72262 & 134 & 361014 & 8060 \\
                other & 195237 & 116902 & 200179 & 12364 & 280703 & 75459 & 348689 & 528461 & 374704 & 74136 & 1803 & 12053 & 185261 & 16274 & 32874 & 56 & 10902 & 10927 & 33 & 2397700 \\
                \bottomrule
            \end{tabular}
        }
        \end{small}
        \vspace{2mm}
    \end{minipage}
\end{table*}

\section{Qualitative Results}
\begin{figure*}[h!]
	\centering
	\small
	\resizebox{0.99\linewidth}{!}{
            \begin{tabular}{cccc}
            	Input & Ground truth & Baseline & Ours \\
            	\includegraphics[width=0.24\linewidth]{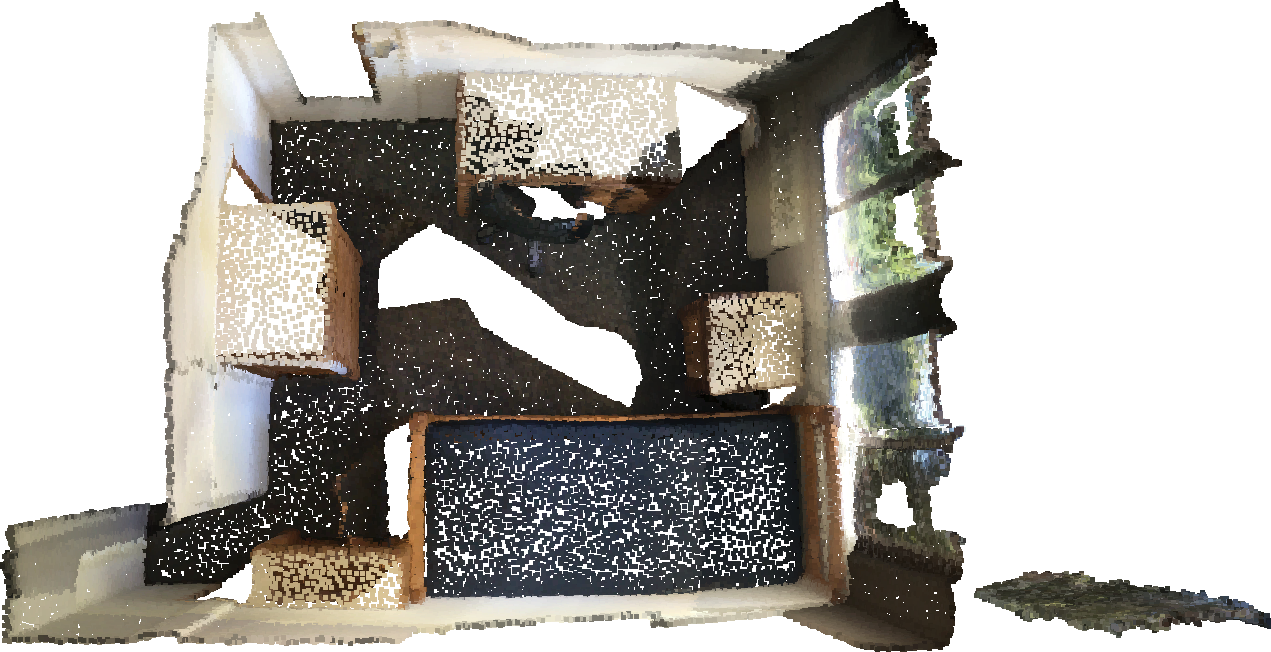}&
            	\includegraphics[width=0.24\linewidth]{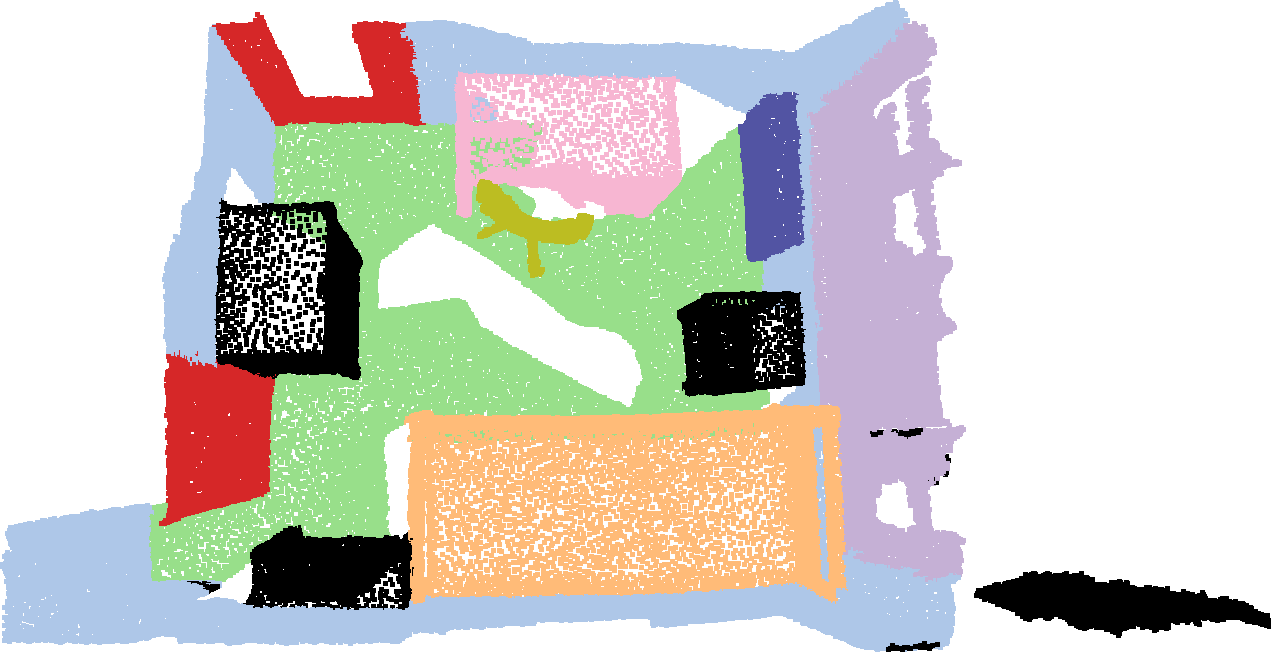}&
            	\includegraphics[width=0.24\linewidth]{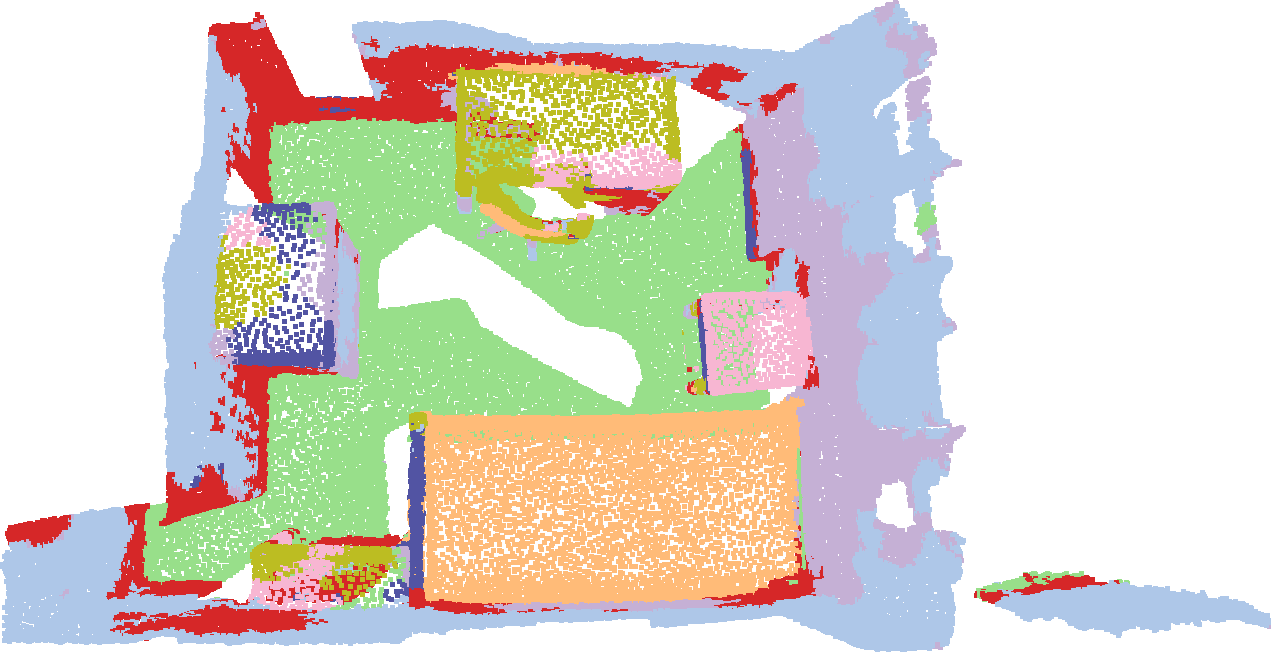}&
            	\includegraphics[width=0.24\linewidth]{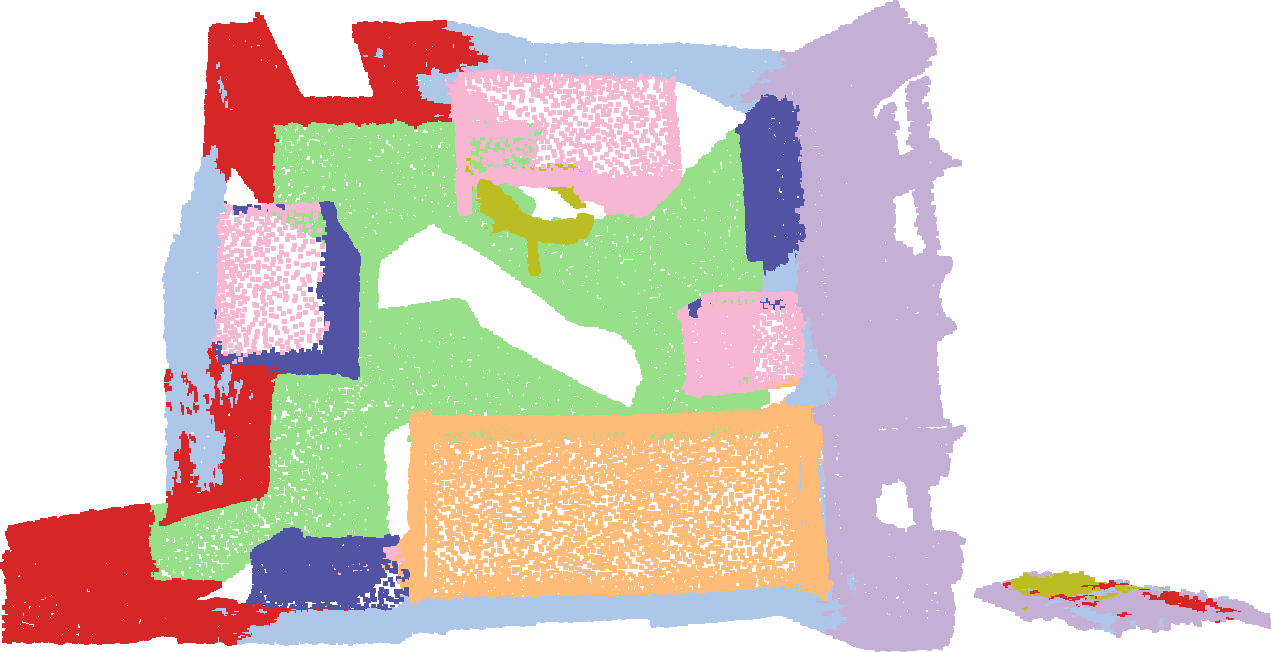}\\

            	\includegraphics[width=0.24\linewidth]{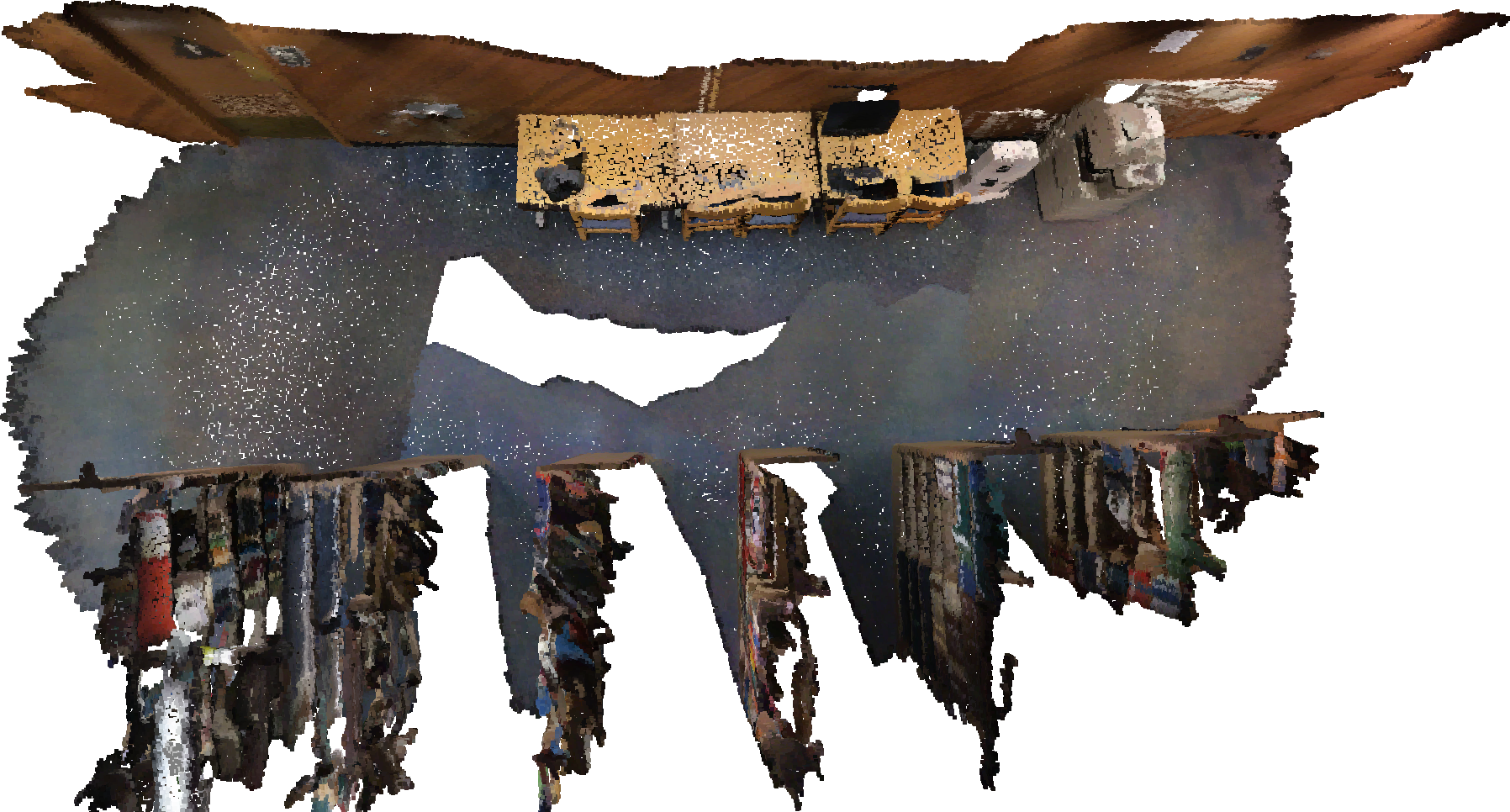}&
            	\includegraphics[width=0.24\linewidth]{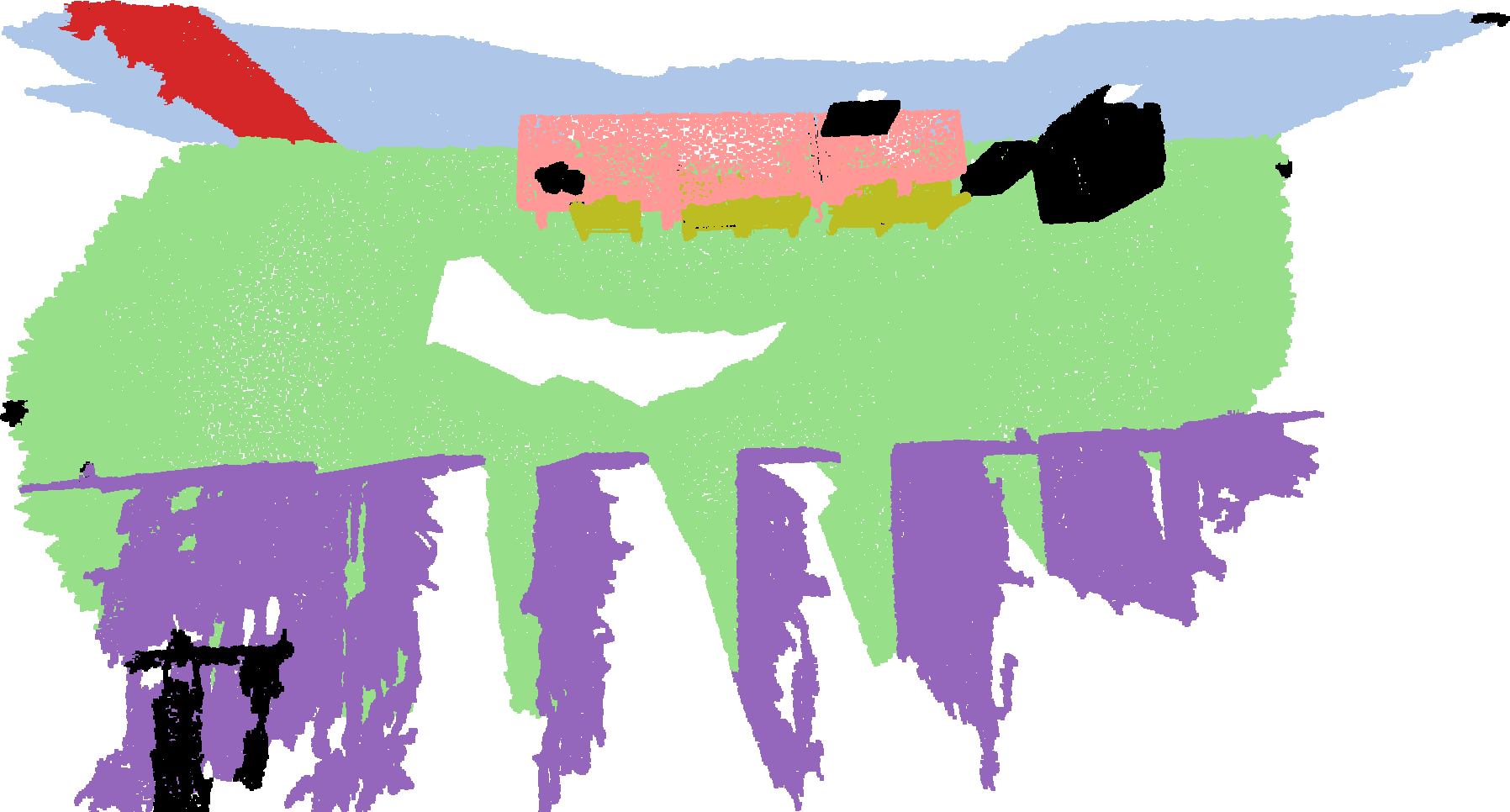}&
            	\includegraphics[width=0.24\linewidth]{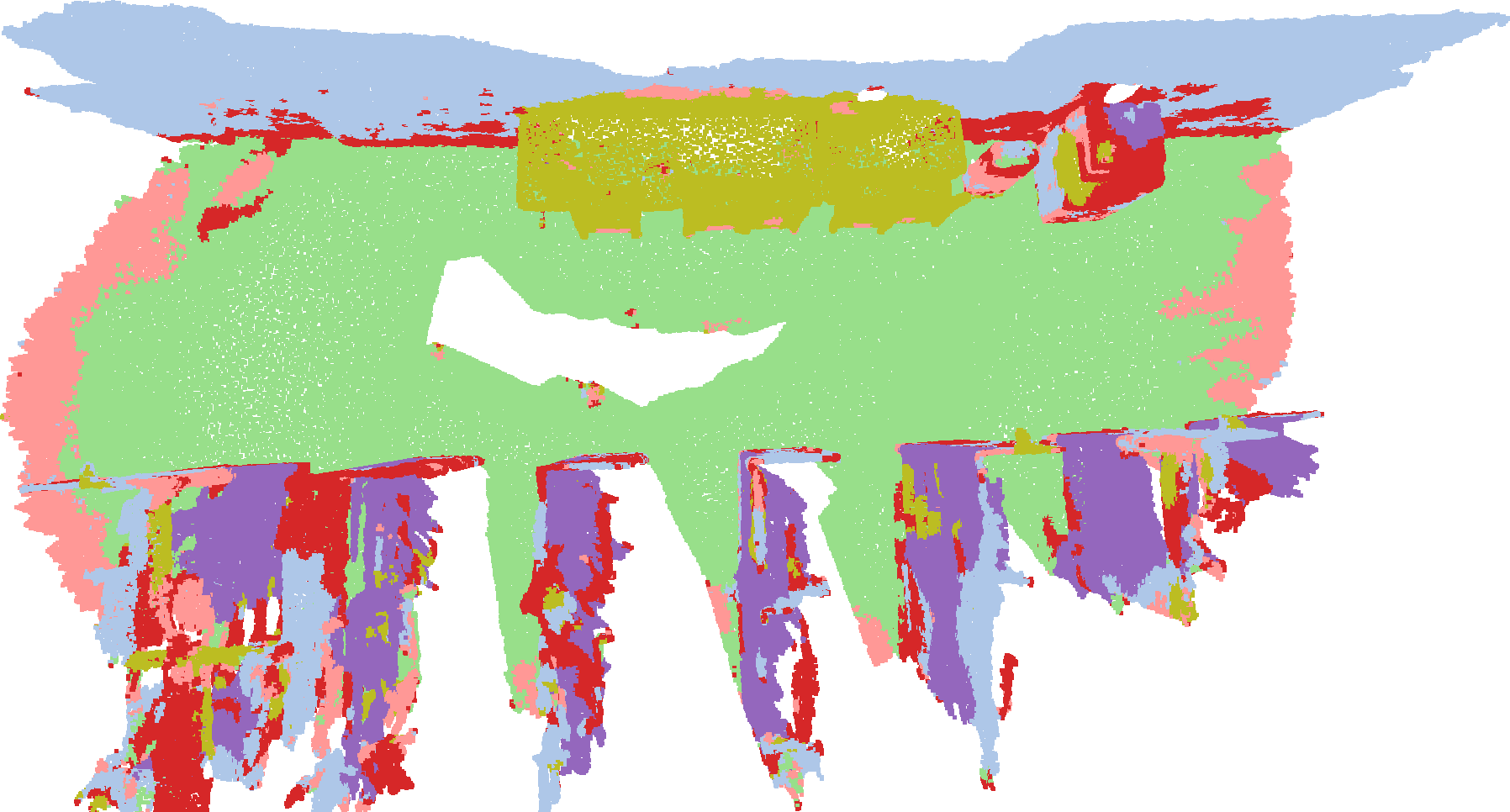}&
            	\includegraphics[width=0.24\linewidth]{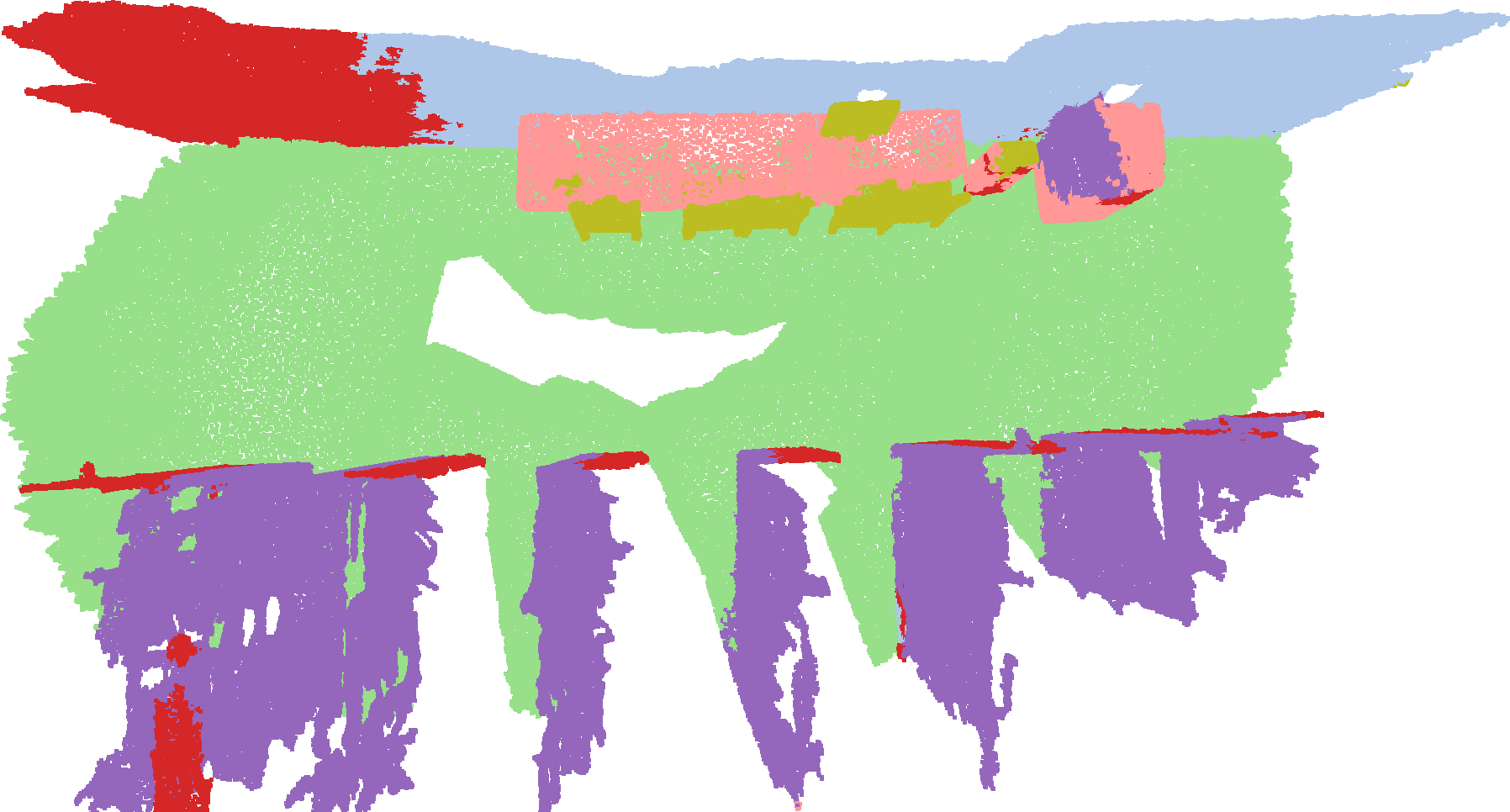}\\

            	\includegraphics[width=0.24\linewidth]{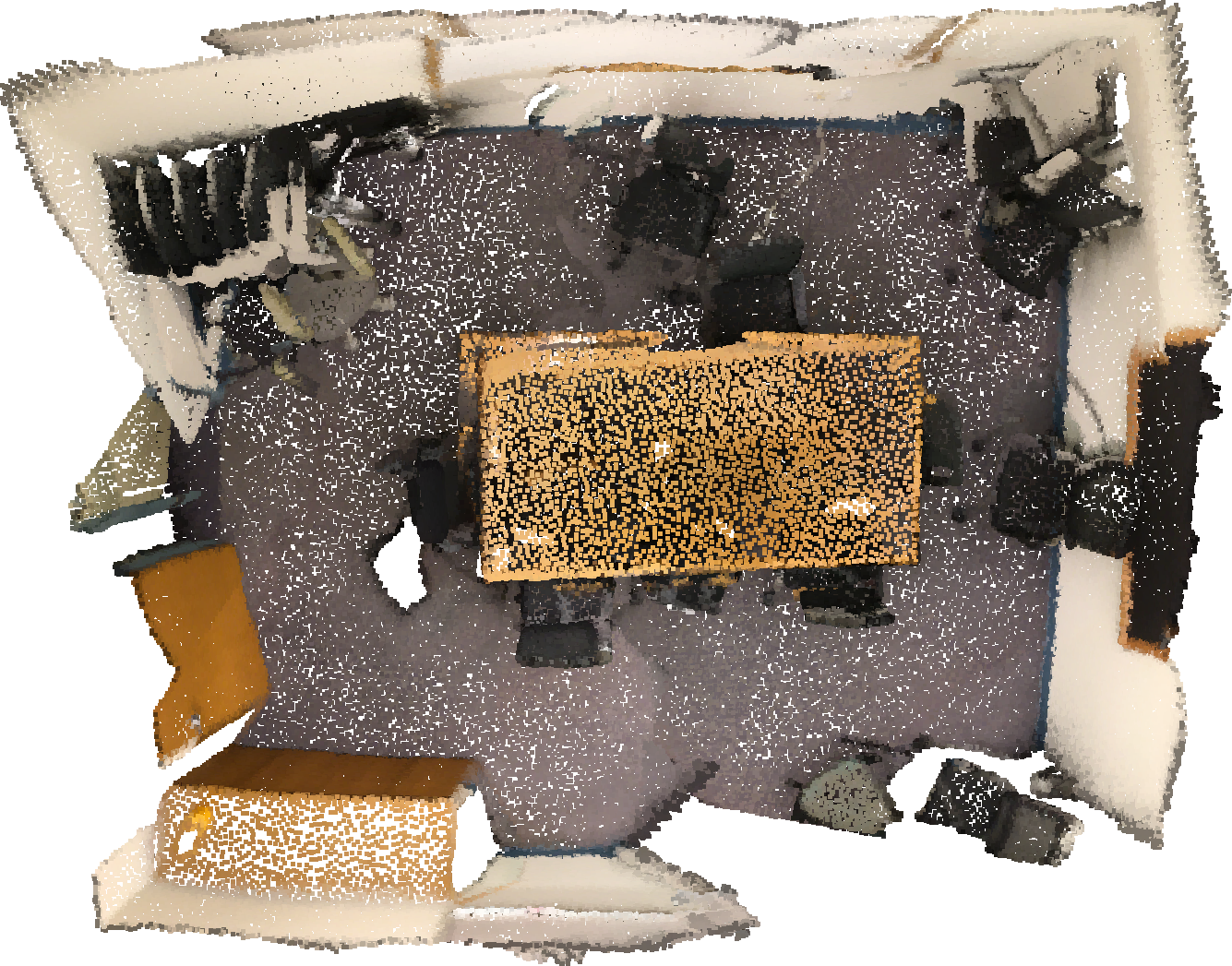}&
            	\includegraphics[width=0.24\linewidth]{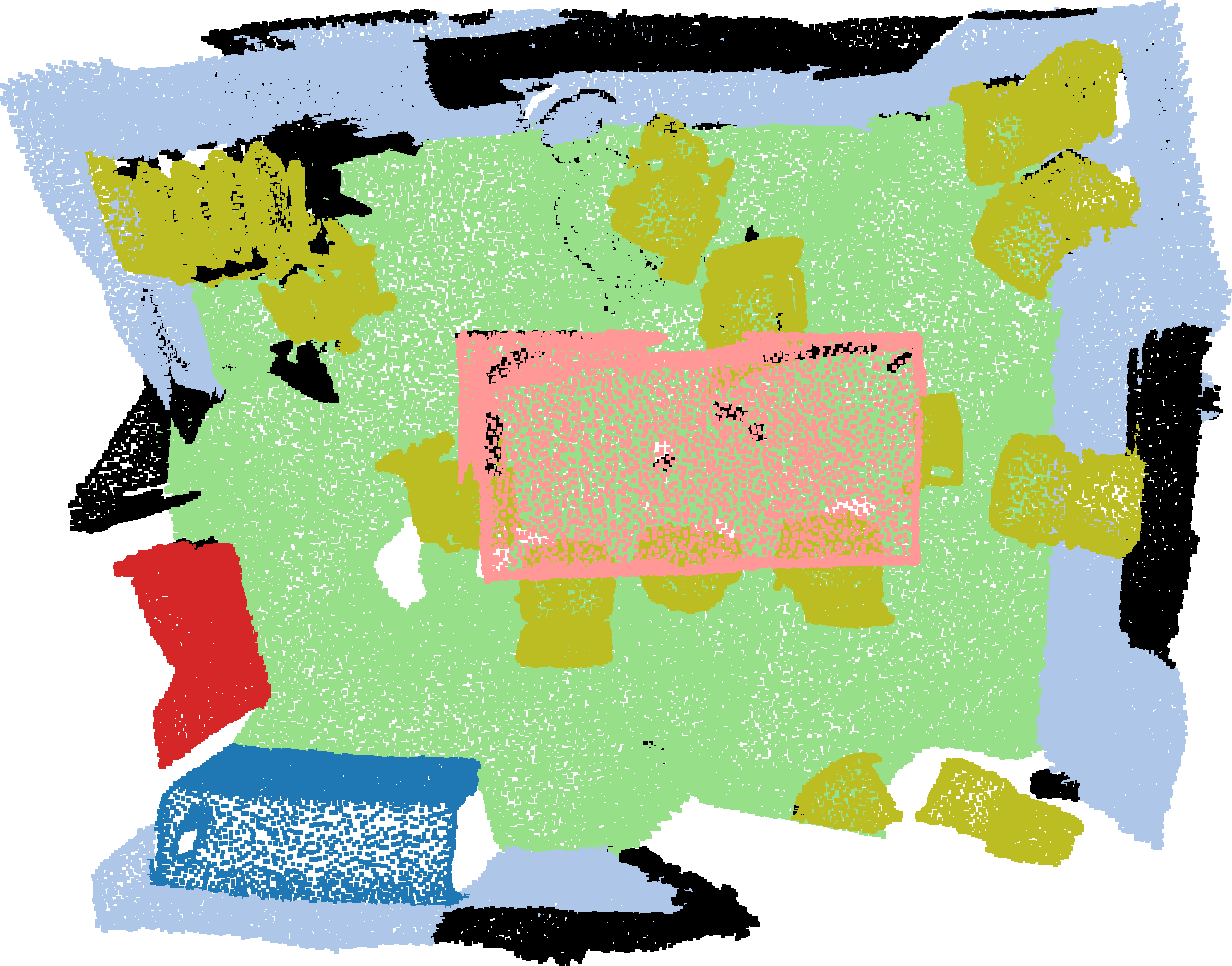}&
            	\includegraphics[width=0.24\linewidth]{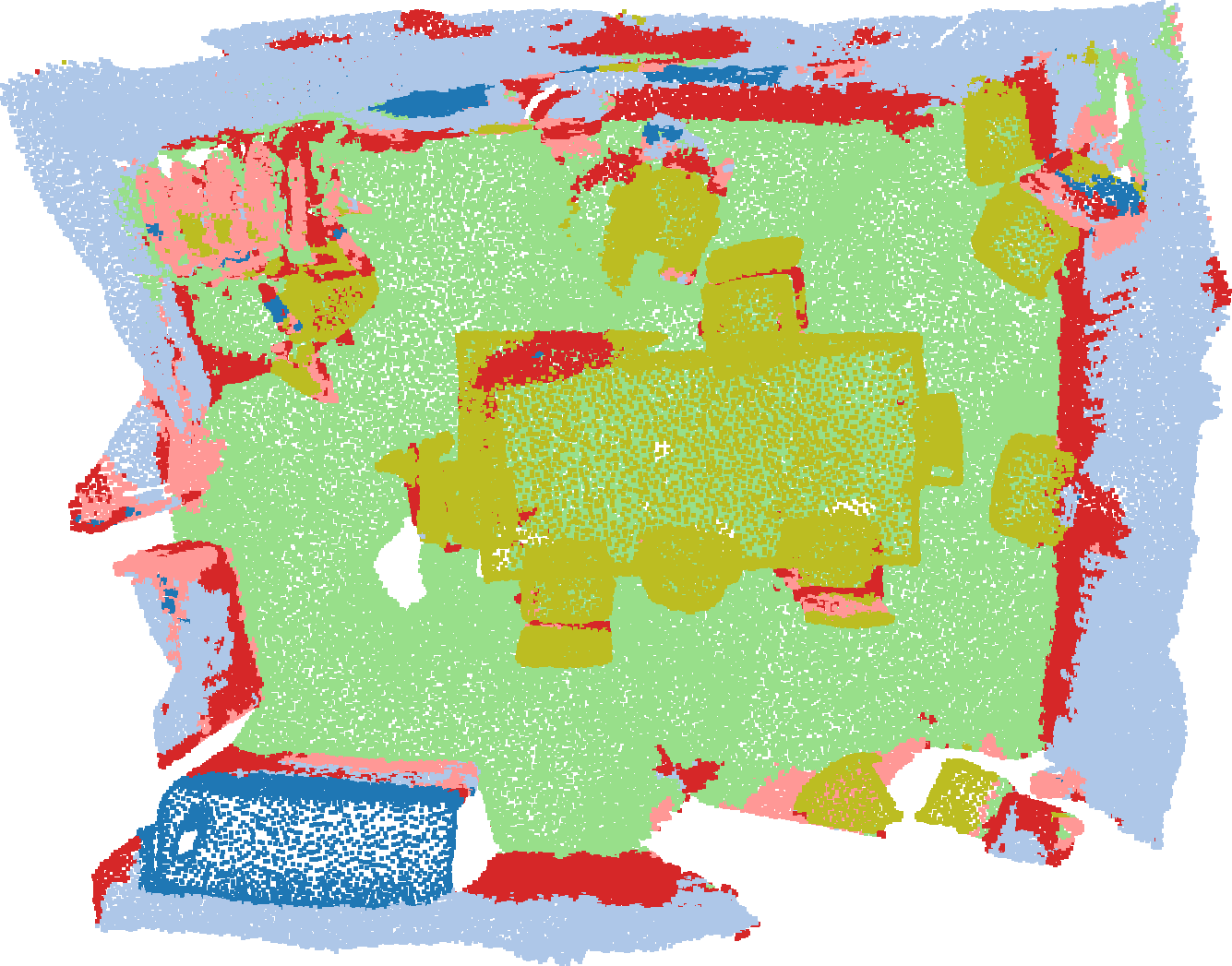}&
            	\includegraphics[width=0.24\linewidth]{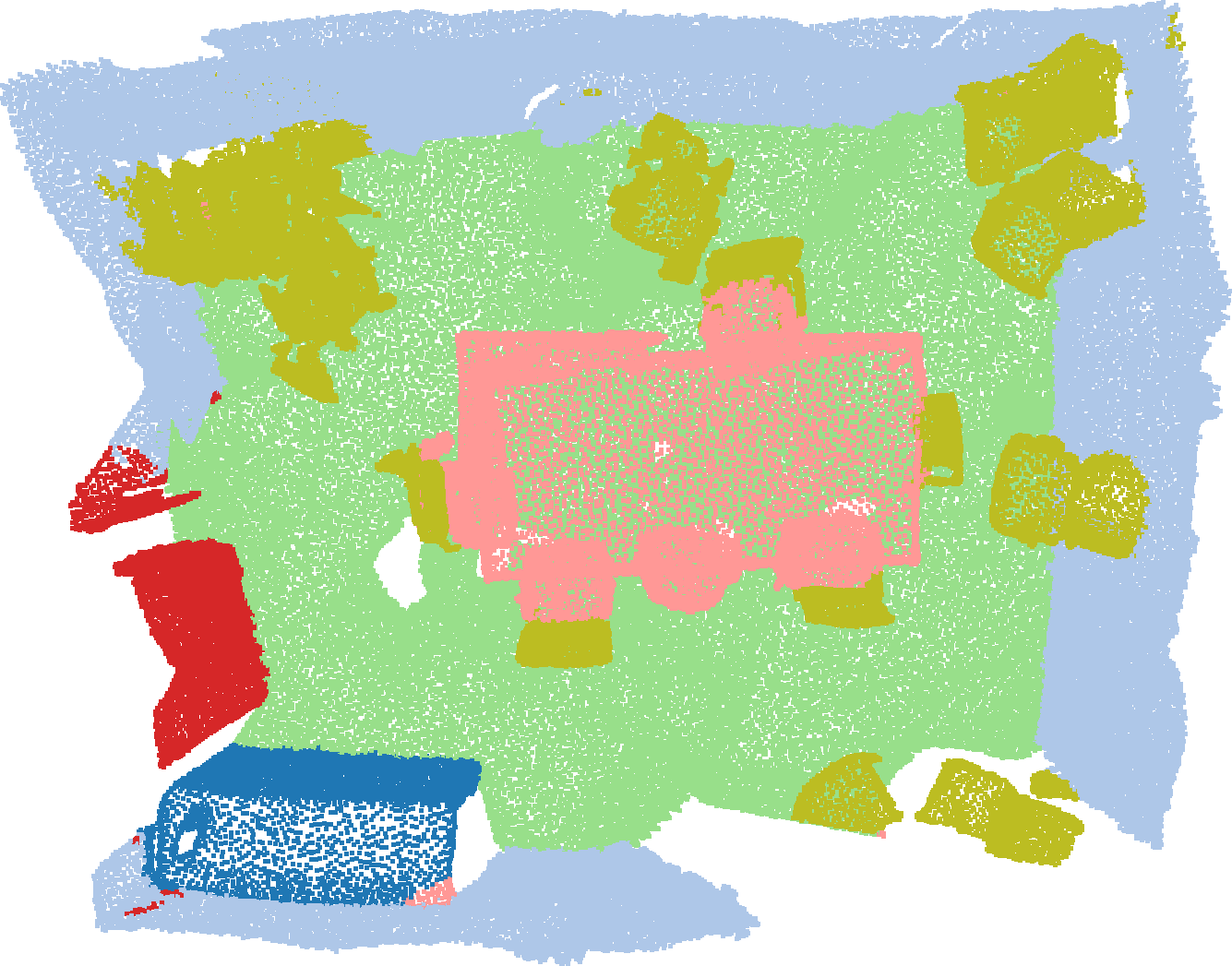}\\

            	\includegraphics[width=0.24\linewidth]{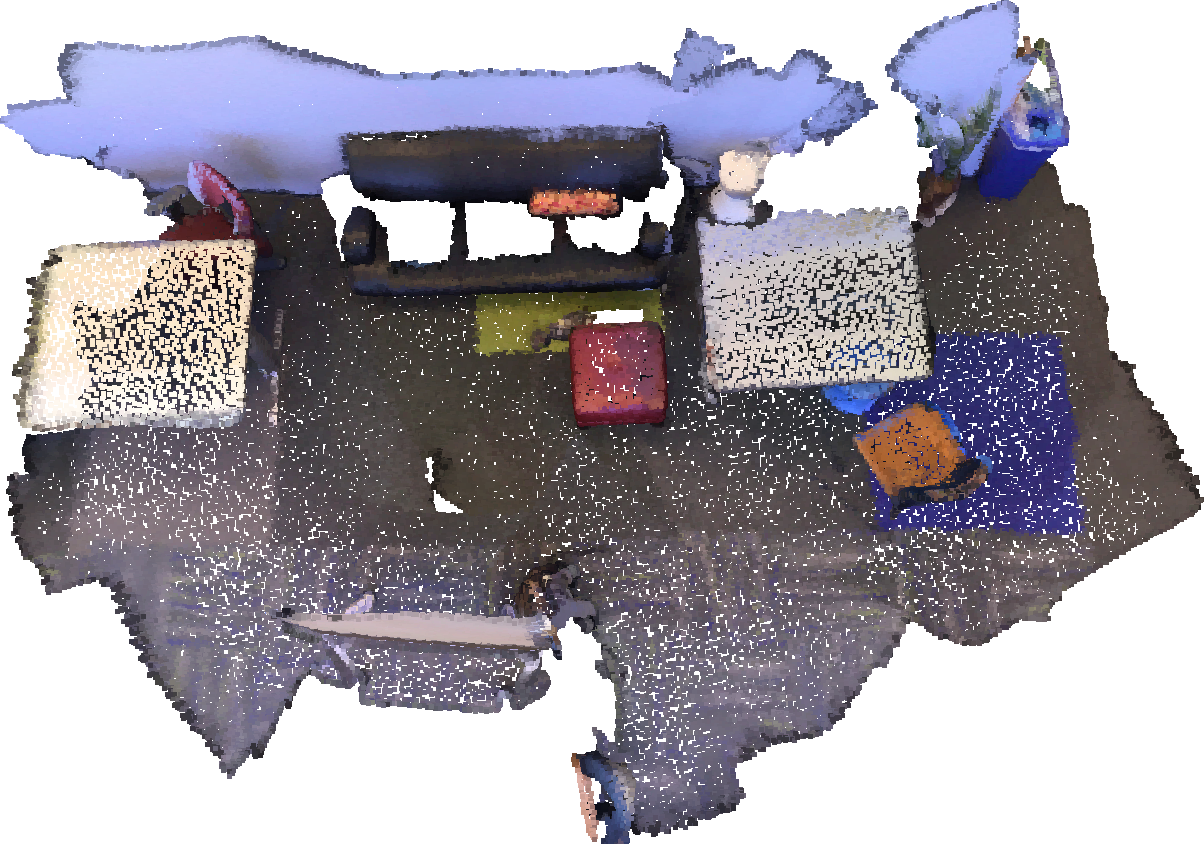}&
            	\includegraphics[width=0.24\linewidth]{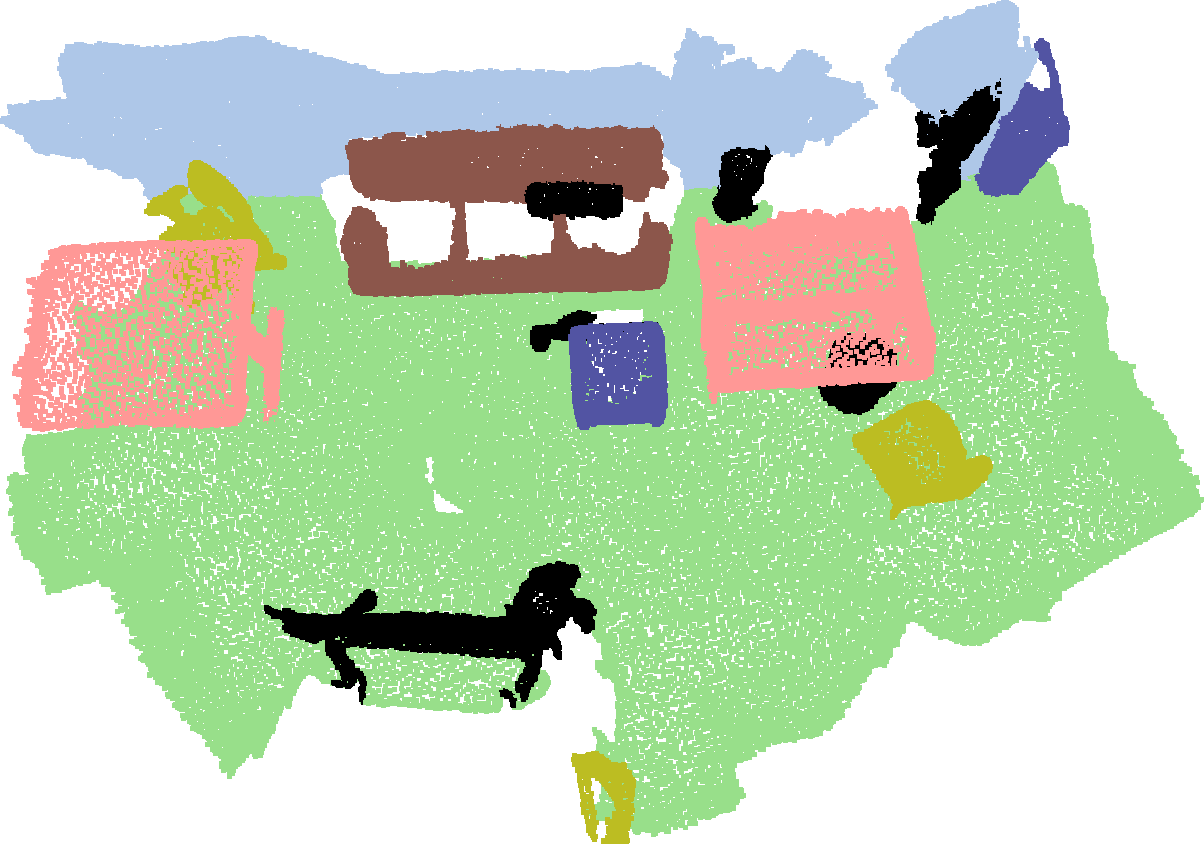}&
            	\includegraphics[width=0.24\linewidth]{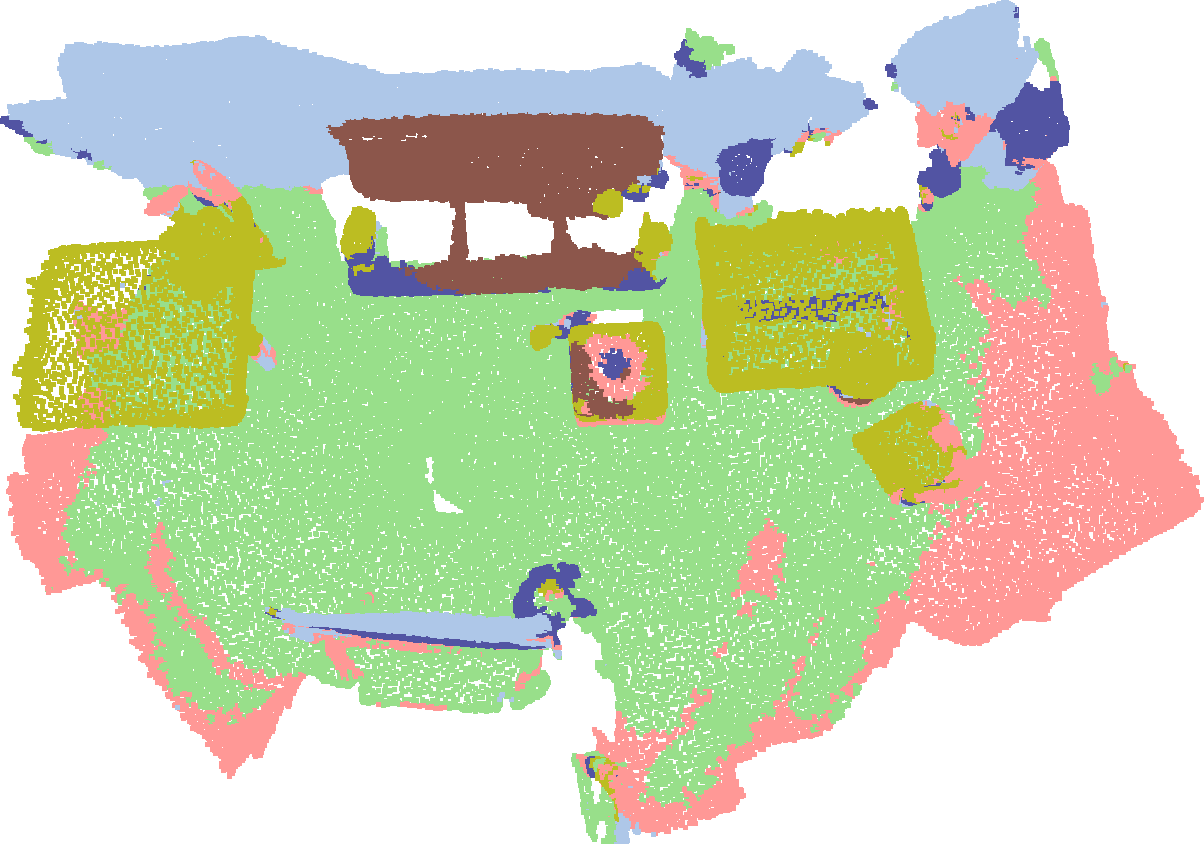}&
            	\includegraphics[width=0.24\linewidth]{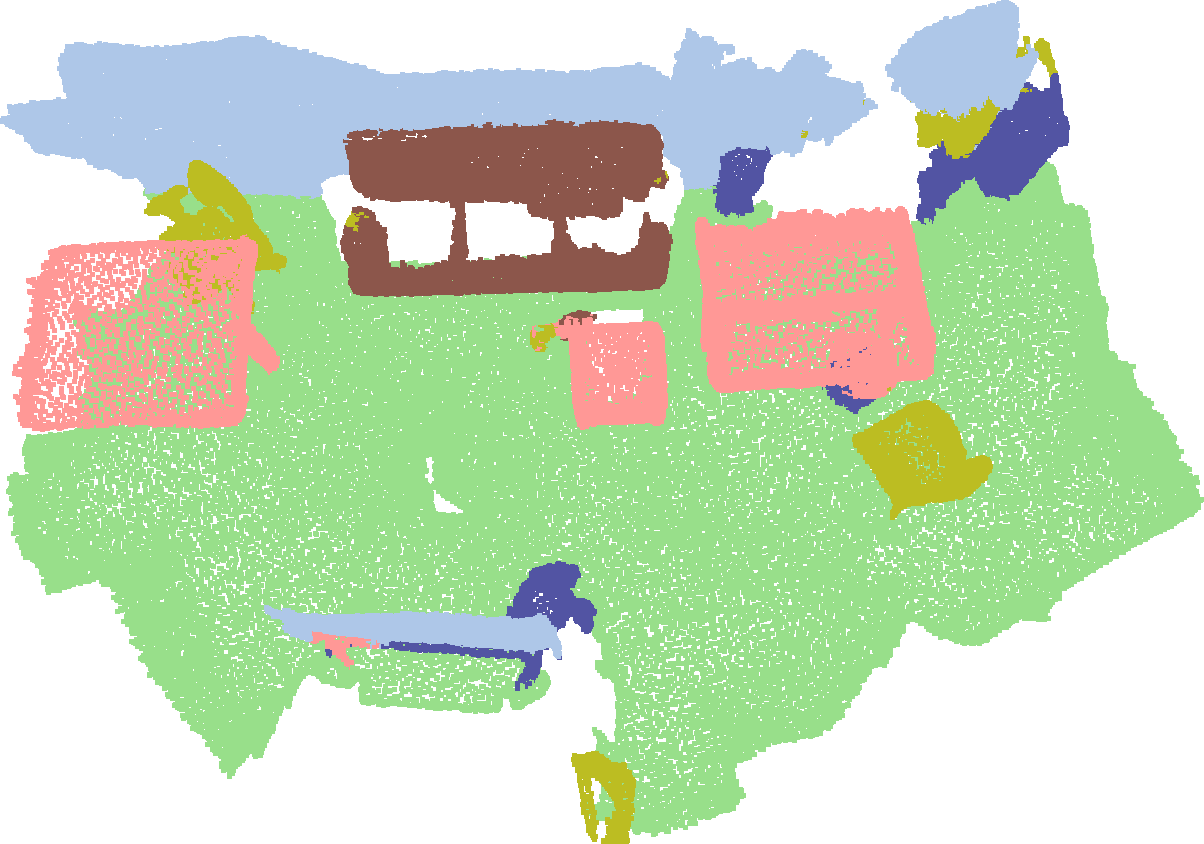}\\

                \multicolumn{4}{c}{\includegraphics[width=0.85\linewidth]{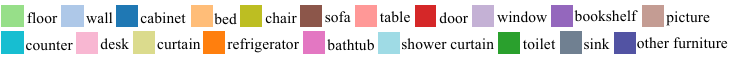}}\\

            	\includegraphics[width=0.24\linewidth]{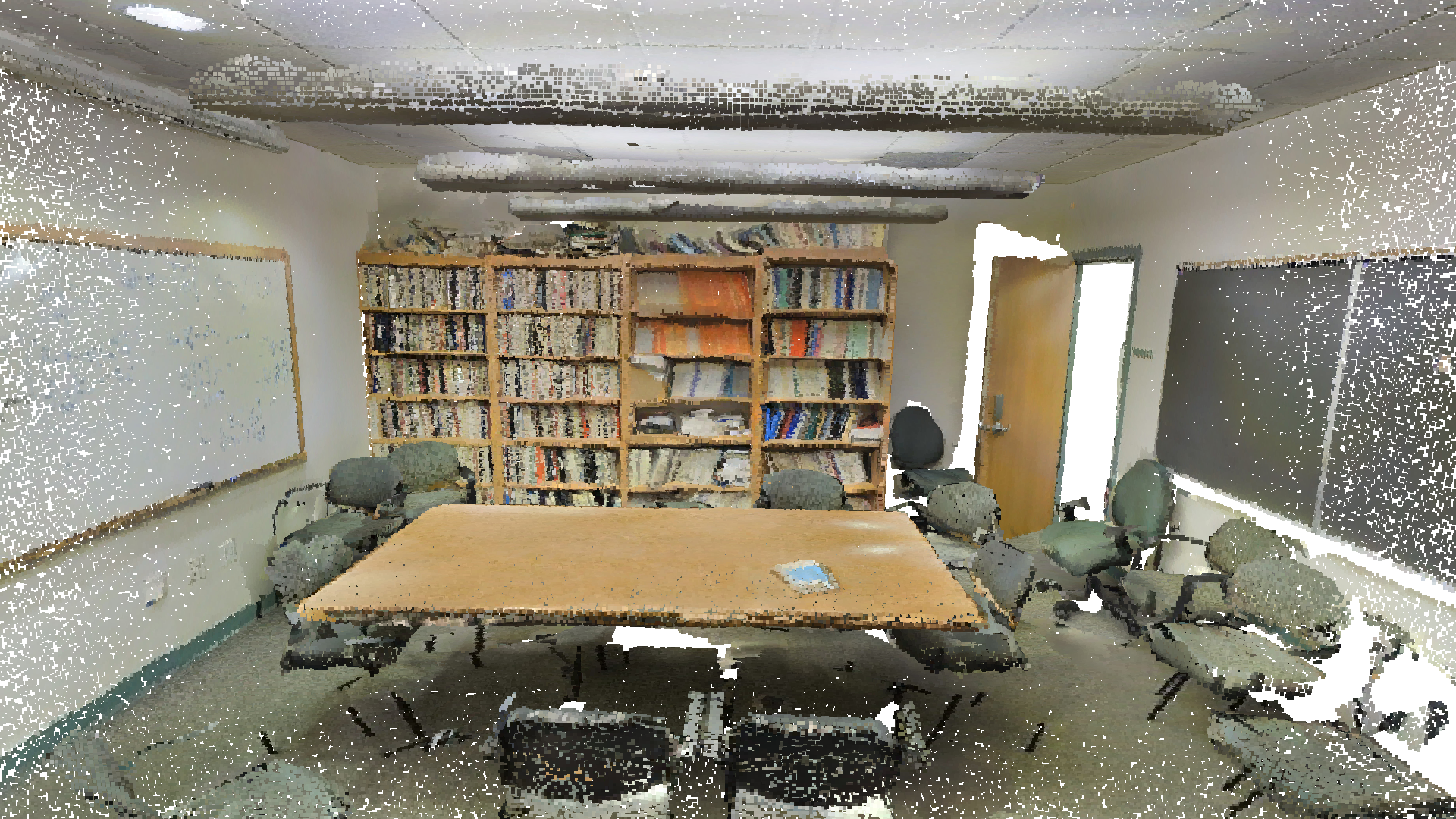}&
            	\includegraphics[width=0.24\linewidth]{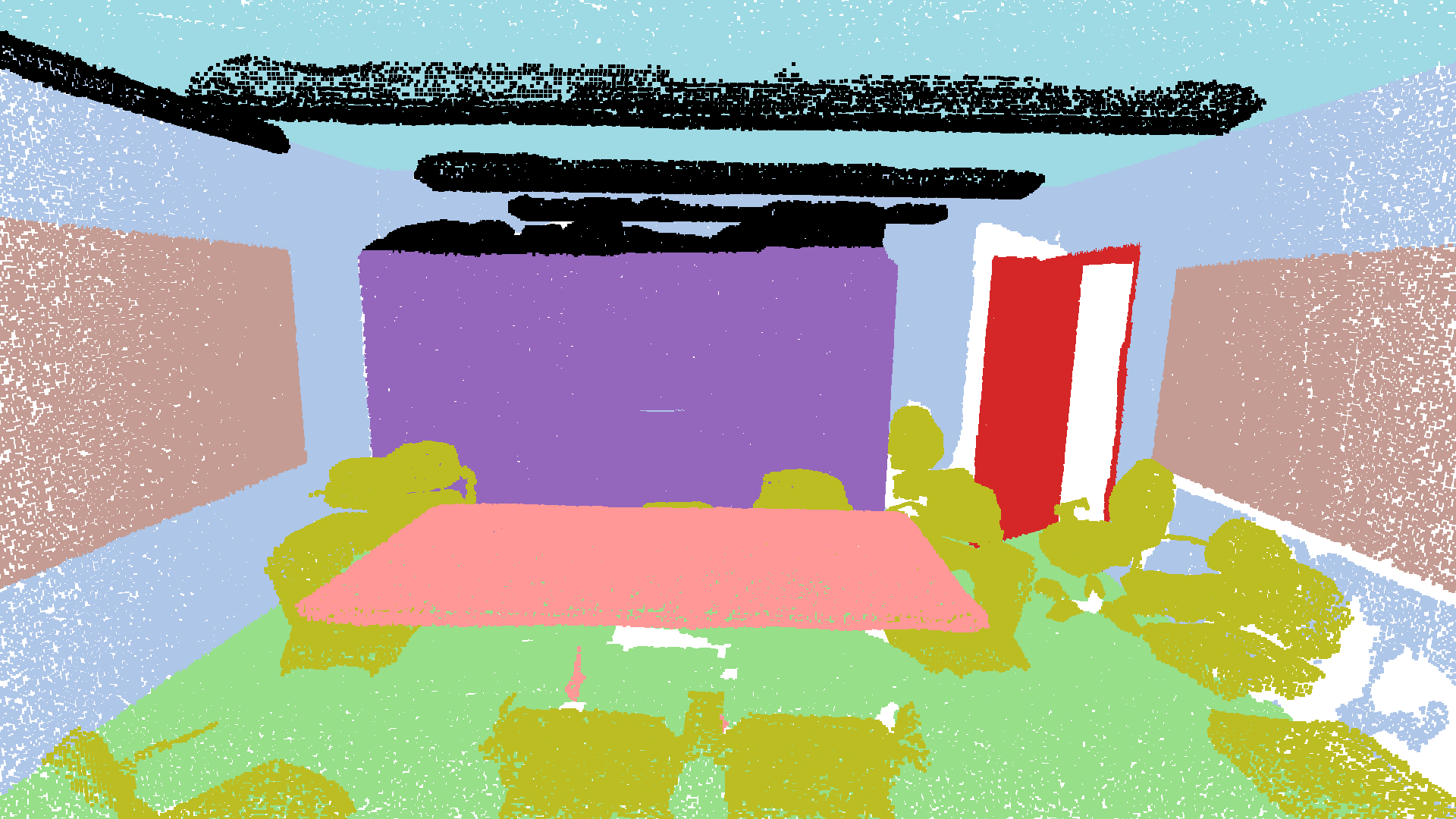}&
            	\includegraphics[width=0.24\linewidth]{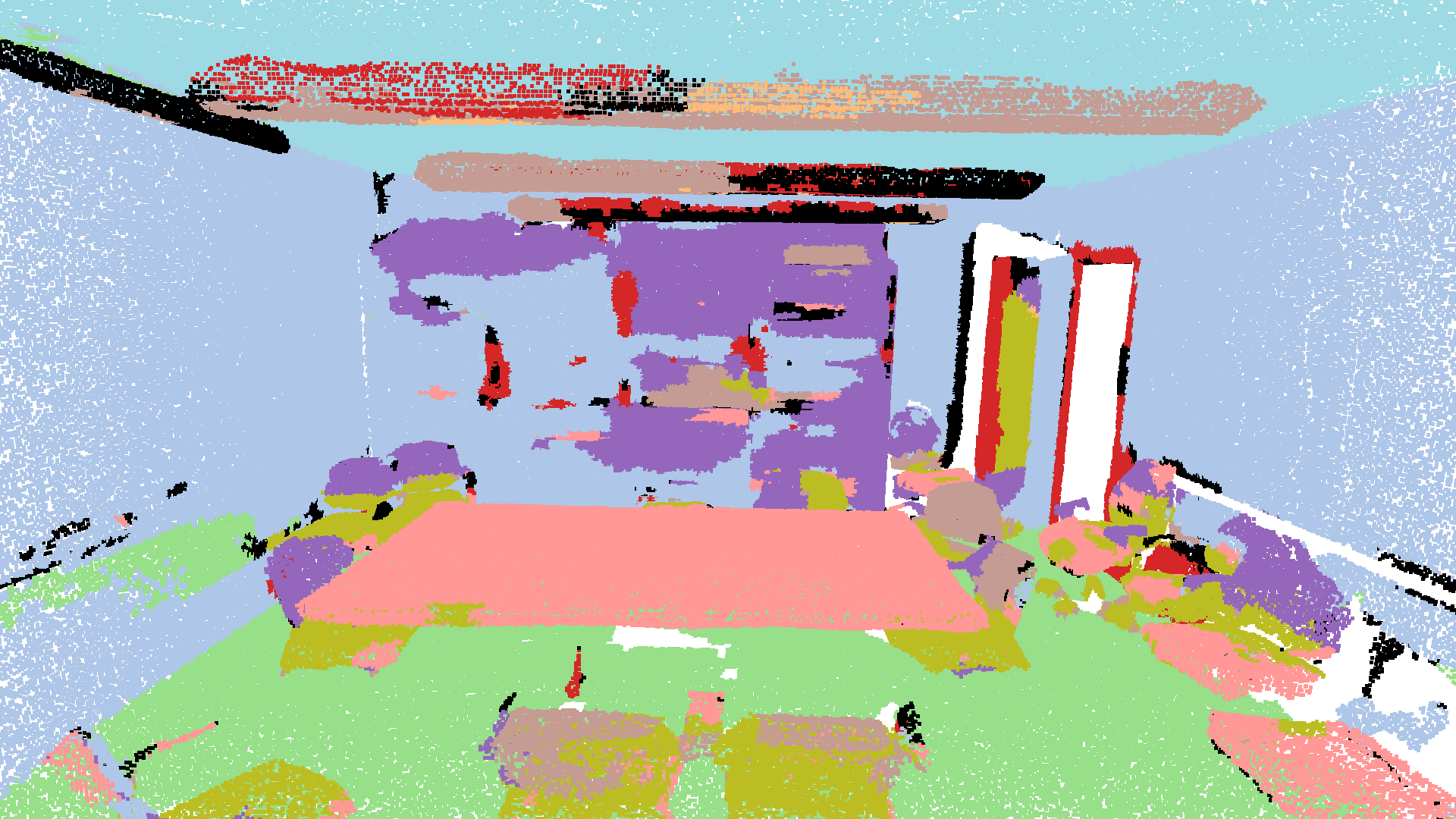}&
            	\includegraphics[width=0.24\linewidth]{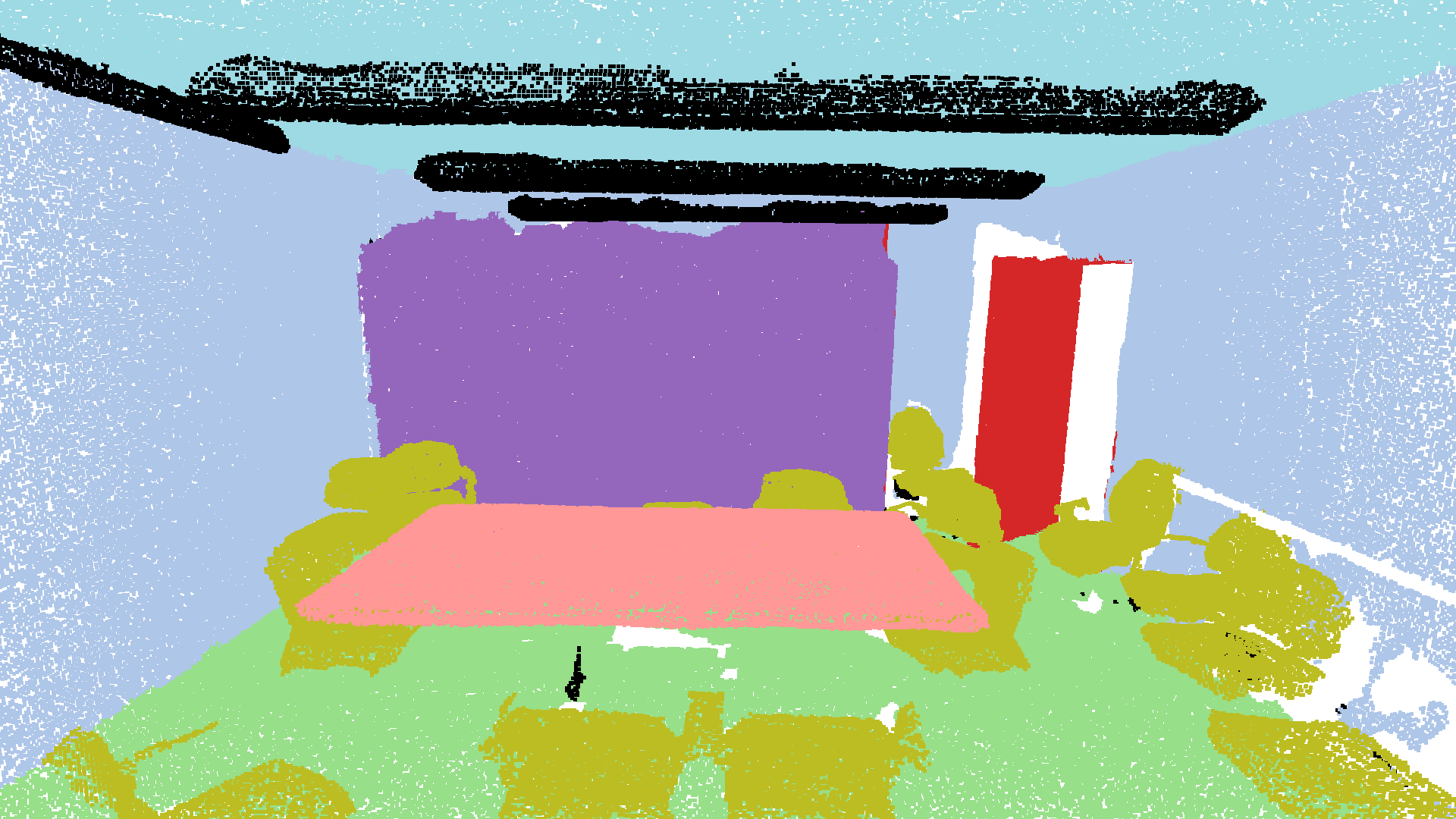}\\
             
            	\includegraphics[width=0.24\linewidth]{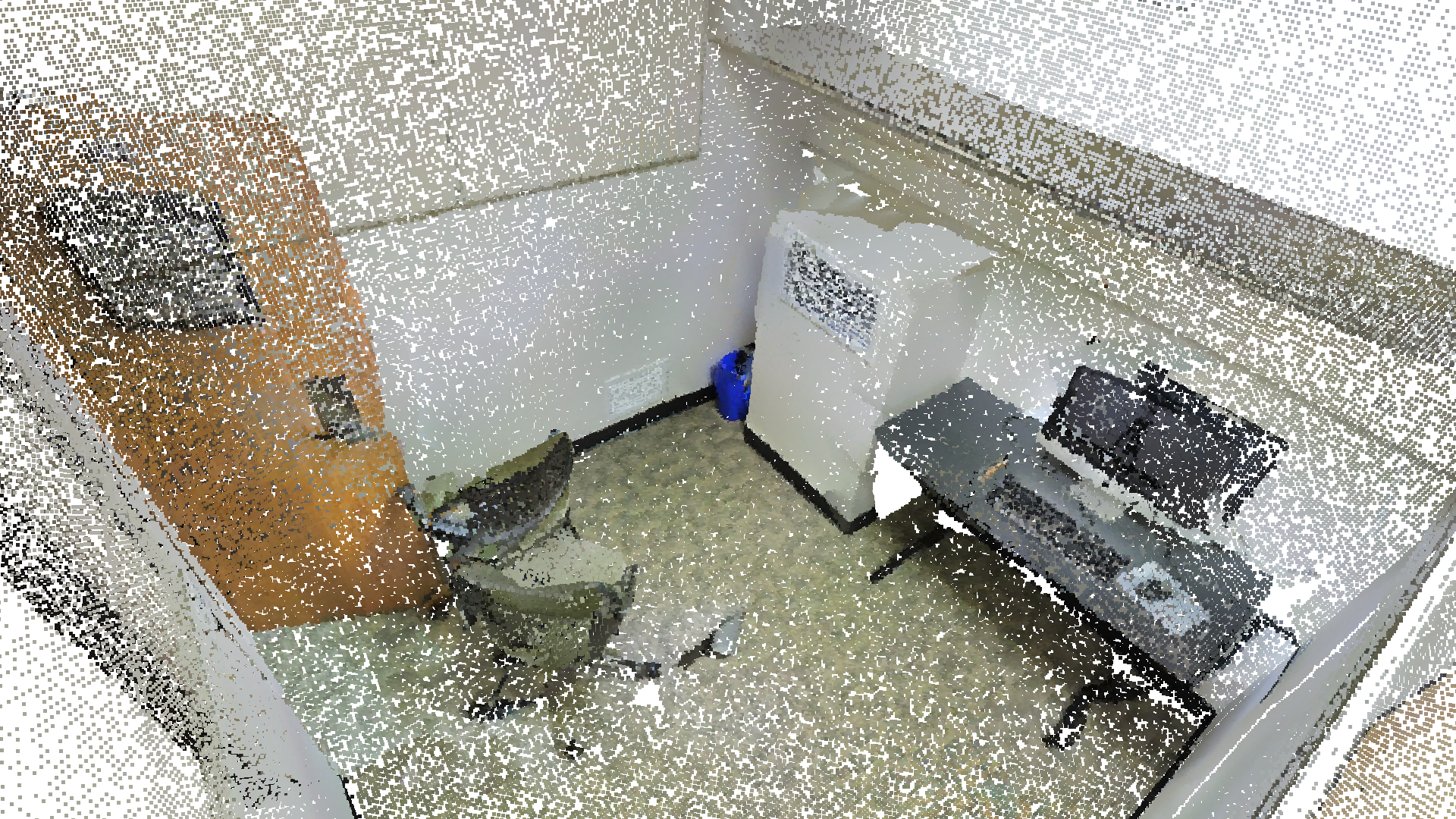}&
            	\includegraphics[width=0.24\linewidth]{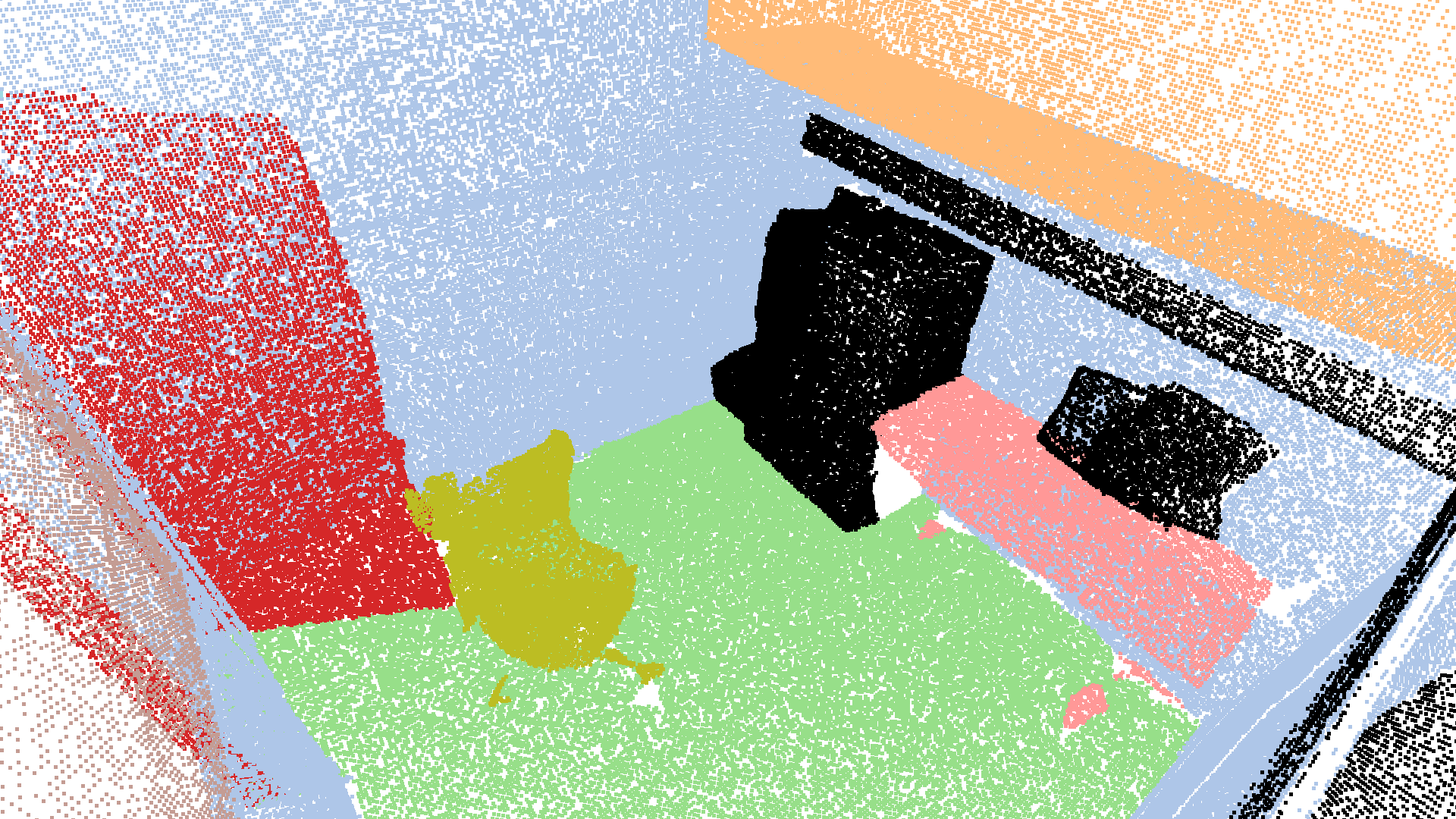}&
            	\includegraphics[width=0.24\linewidth]{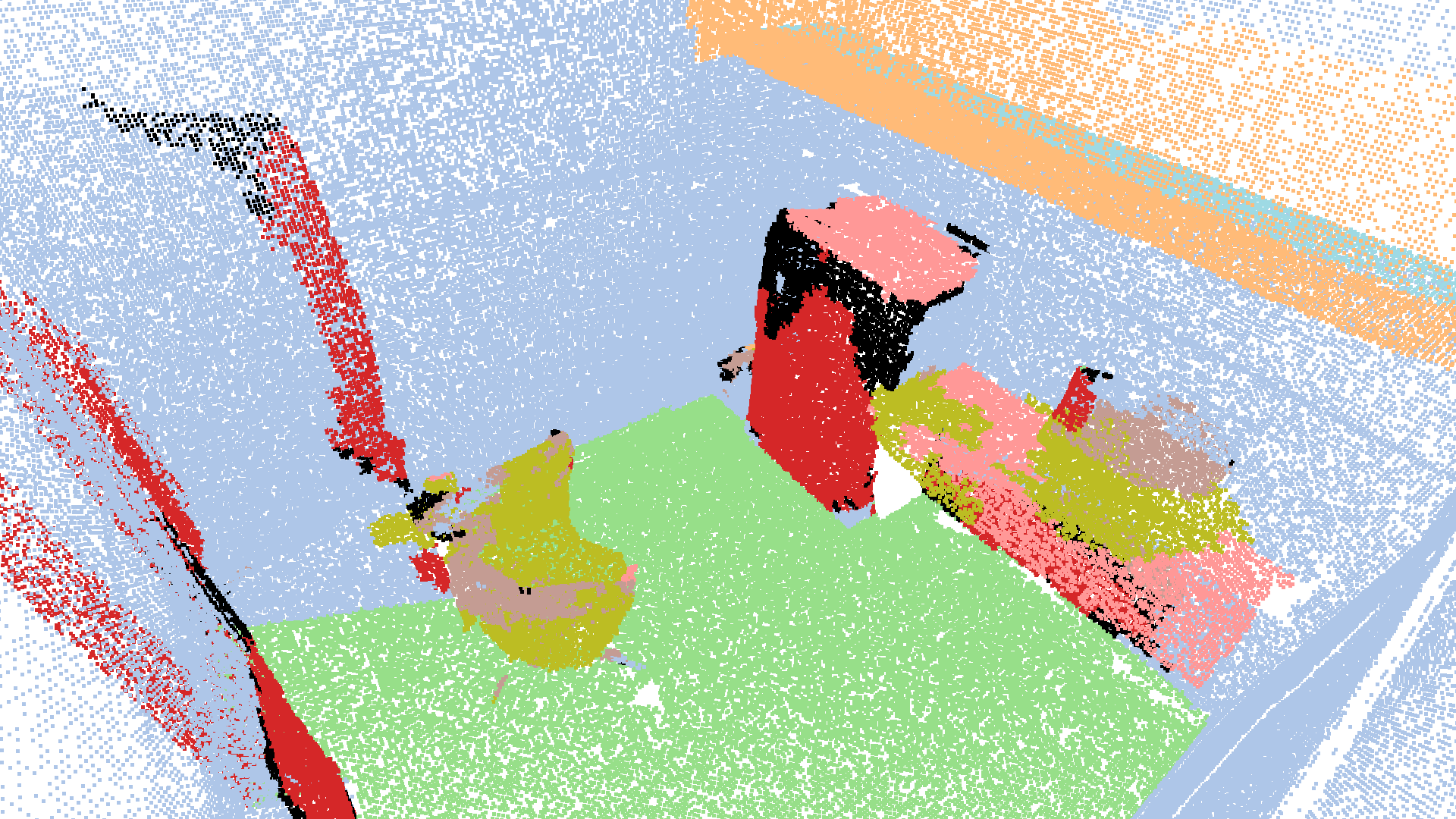}&
            	\includegraphics[width=0.24\linewidth]{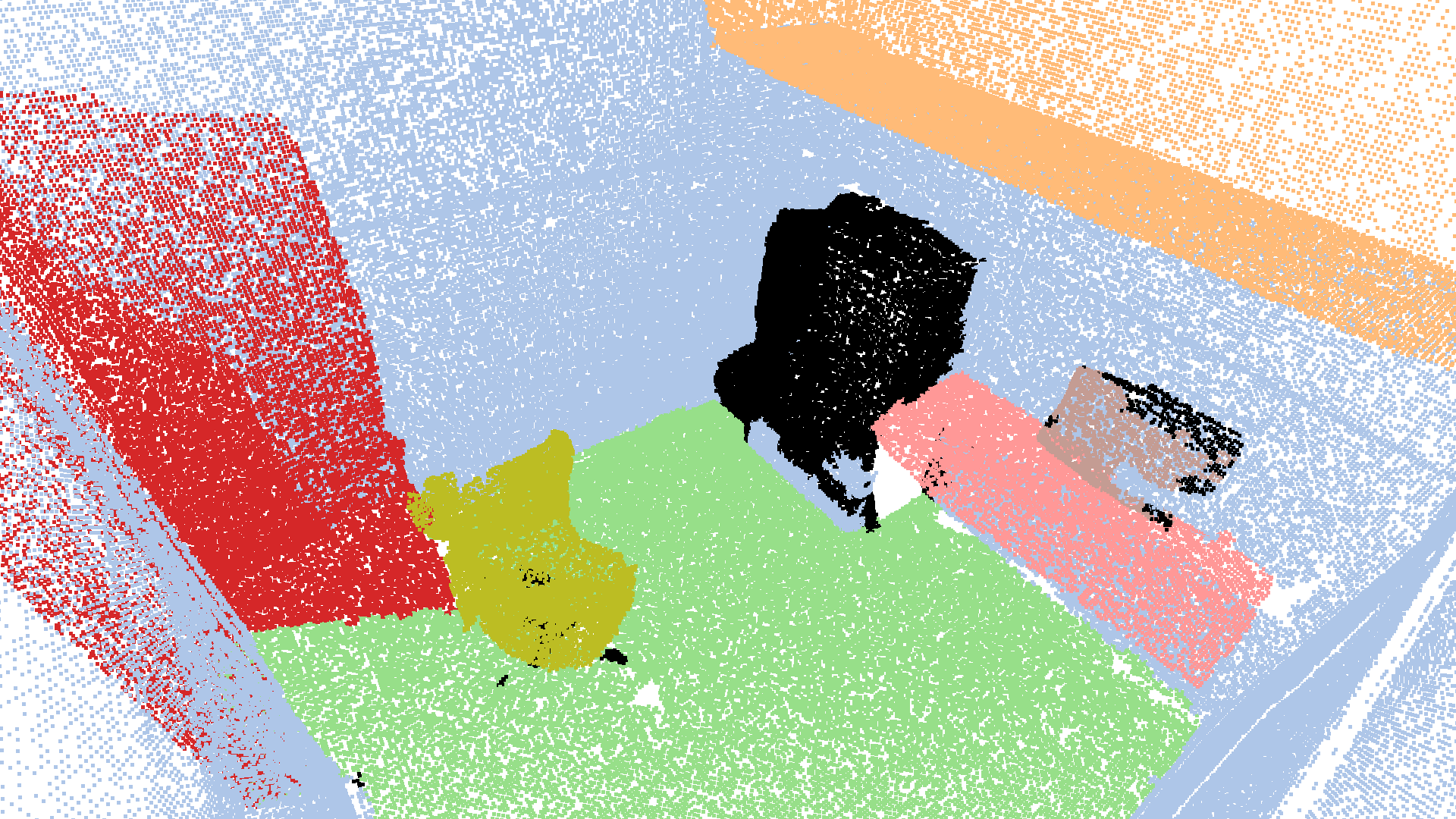}\\

                    \includegraphics[width=0.24\linewidth]{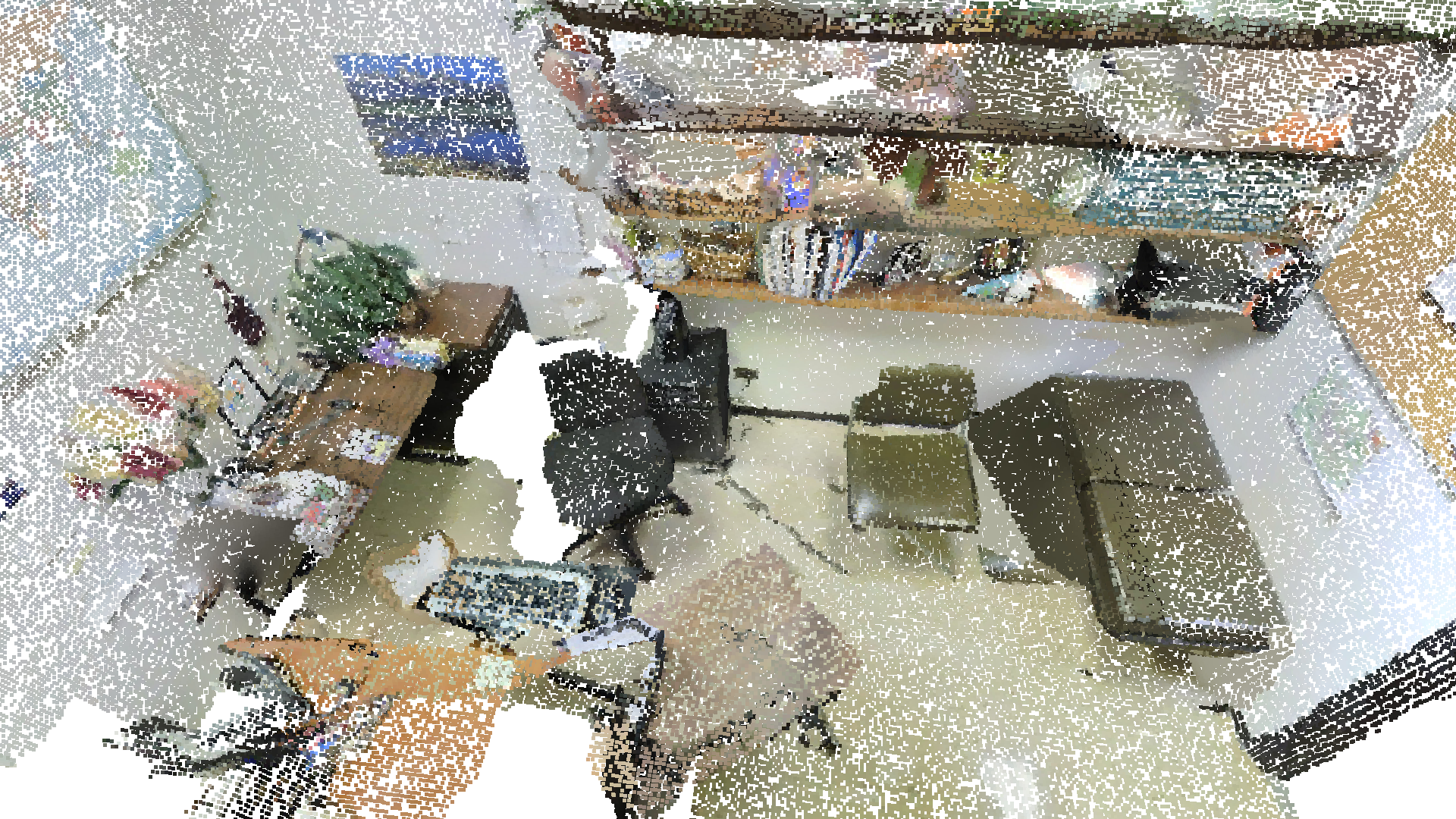}&
            	\includegraphics[width=0.24\linewidth]{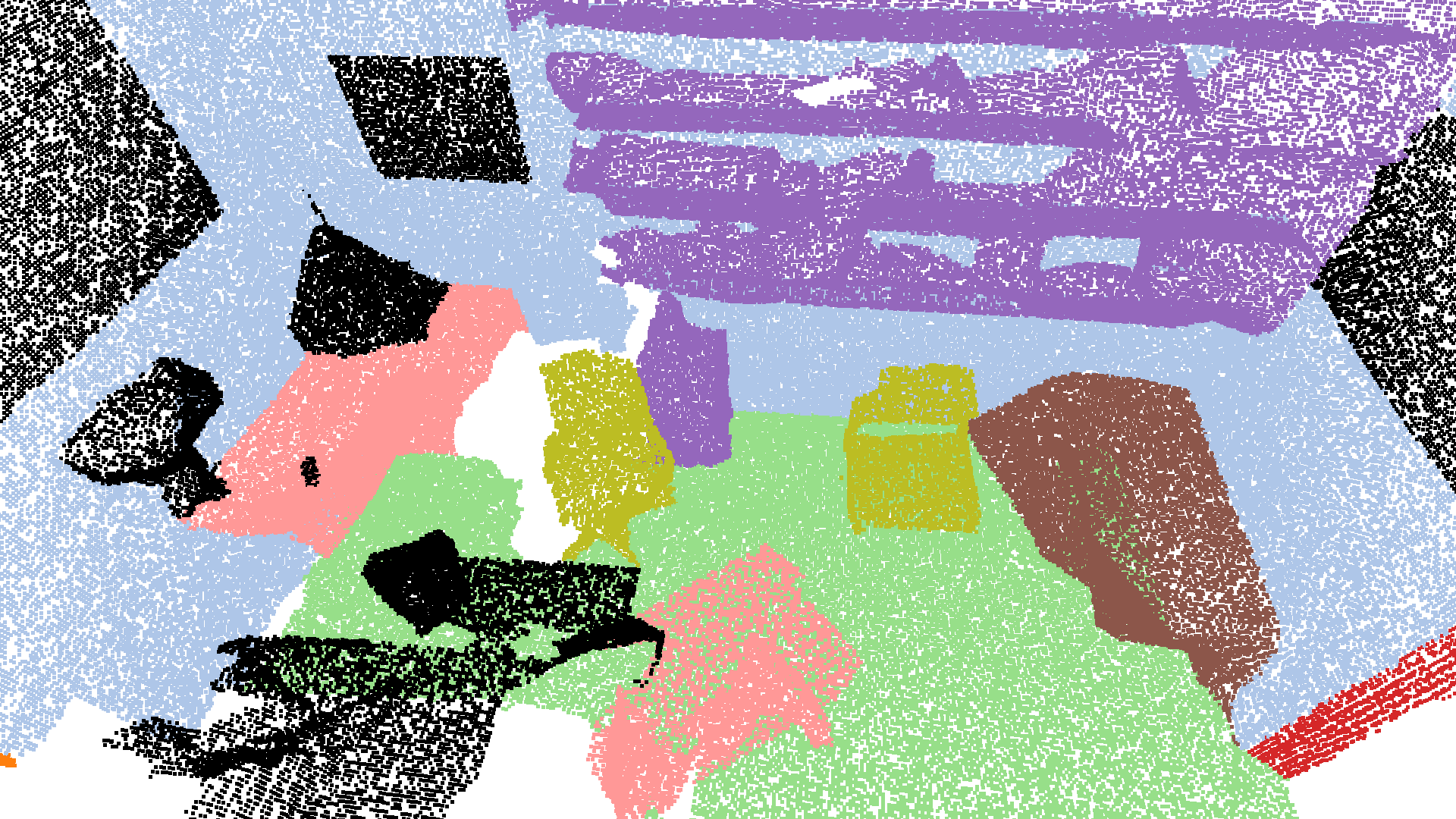}&
            	\includegraphics[width=0.24\linewidth]{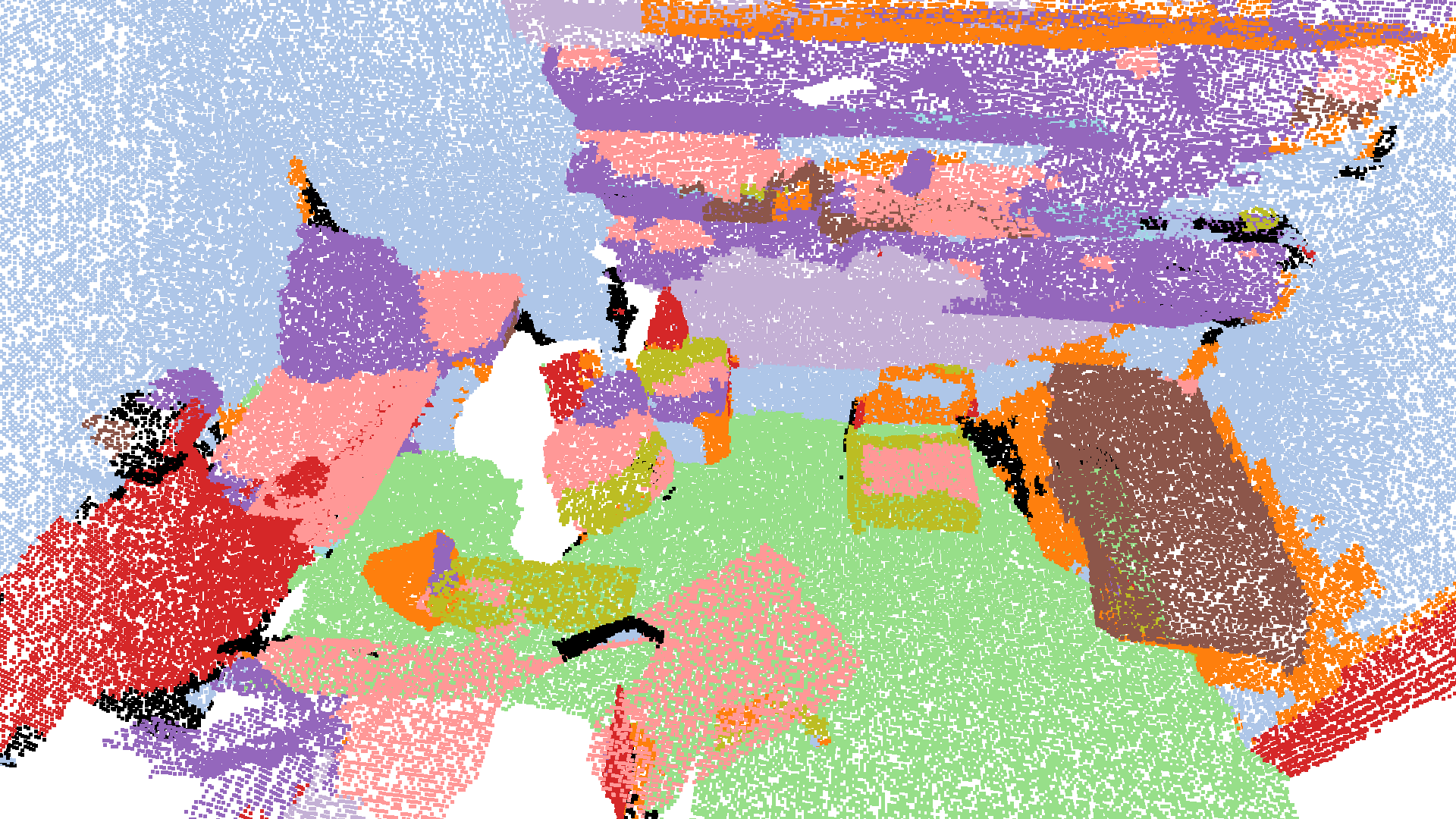}&
            	\includegraphics[width=0.24\linewidth]{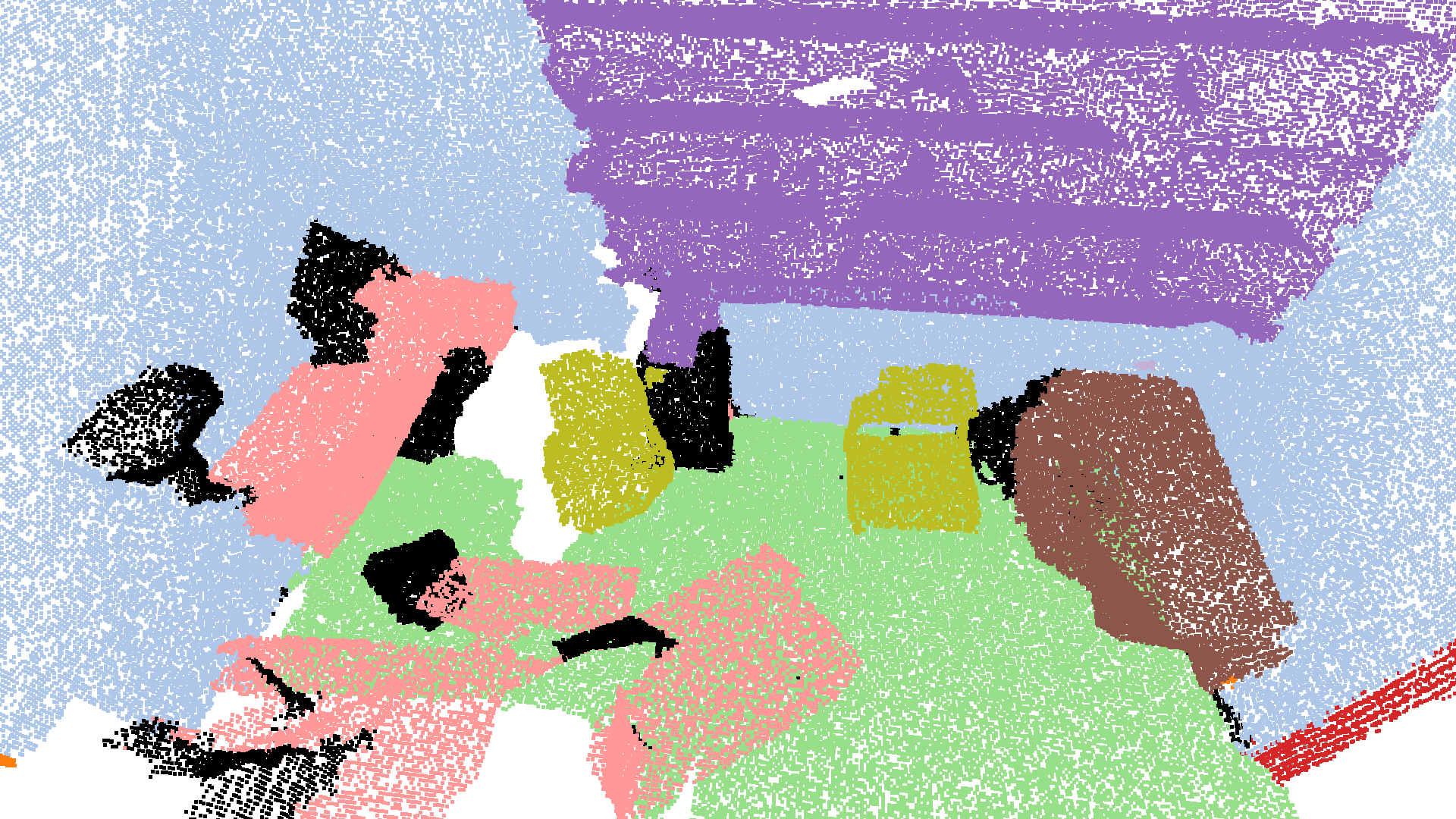}\\

                    \includegraphics[width=0.24\linewidth]{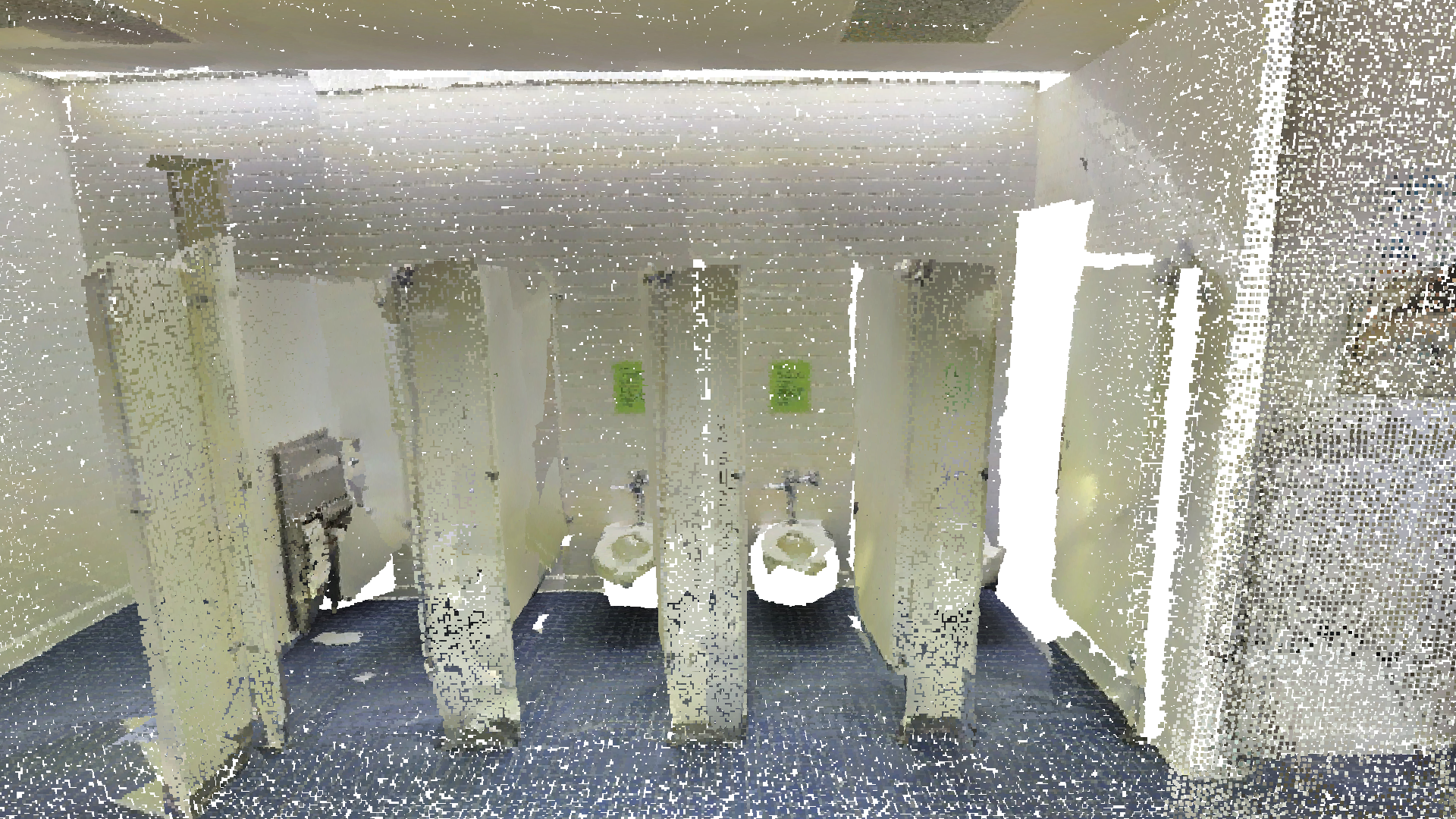}&
            	\includegraphics[width=0.24\linewidth]{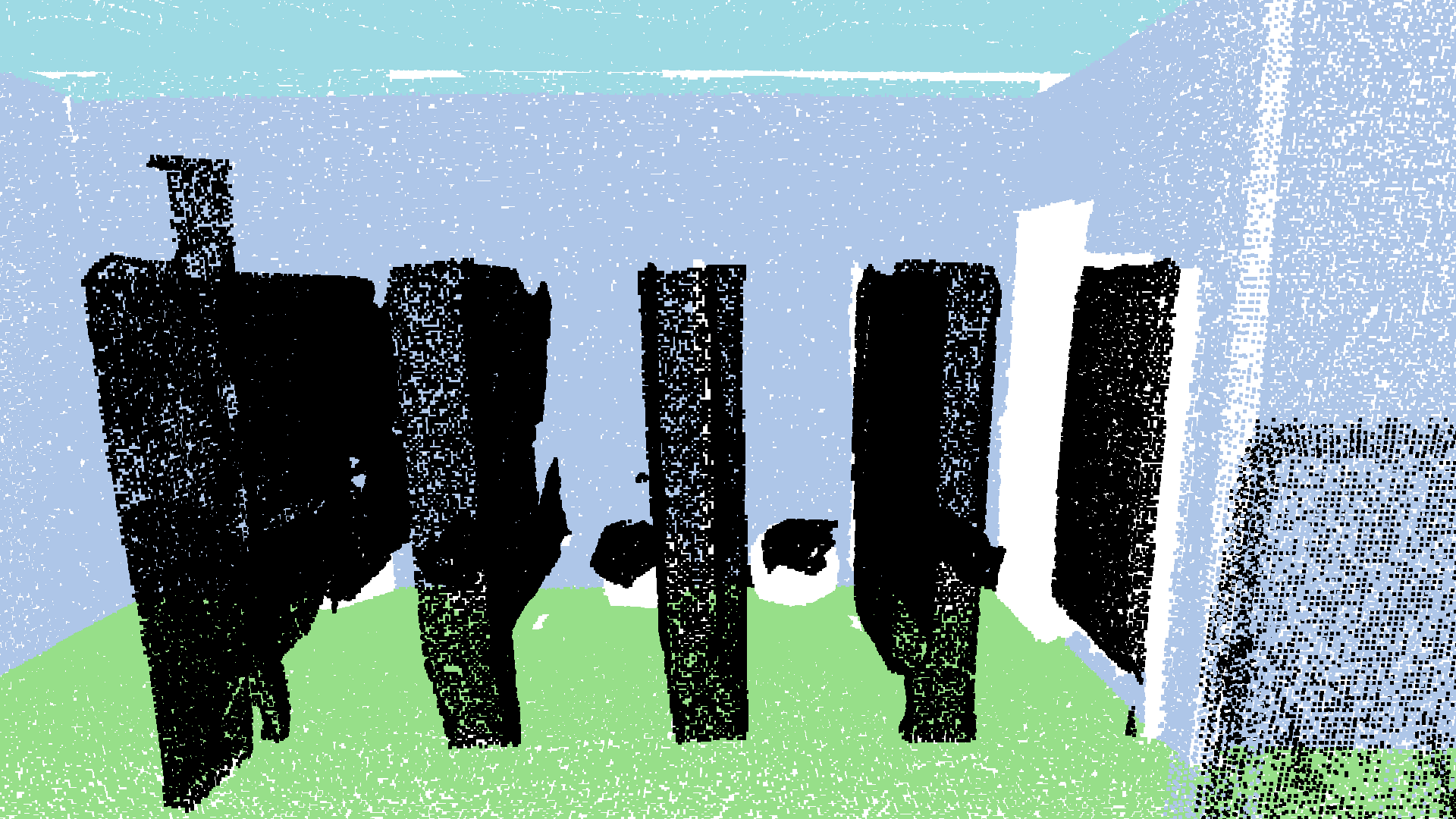}&
            	\includegraphics[width=0.24\linewidth]{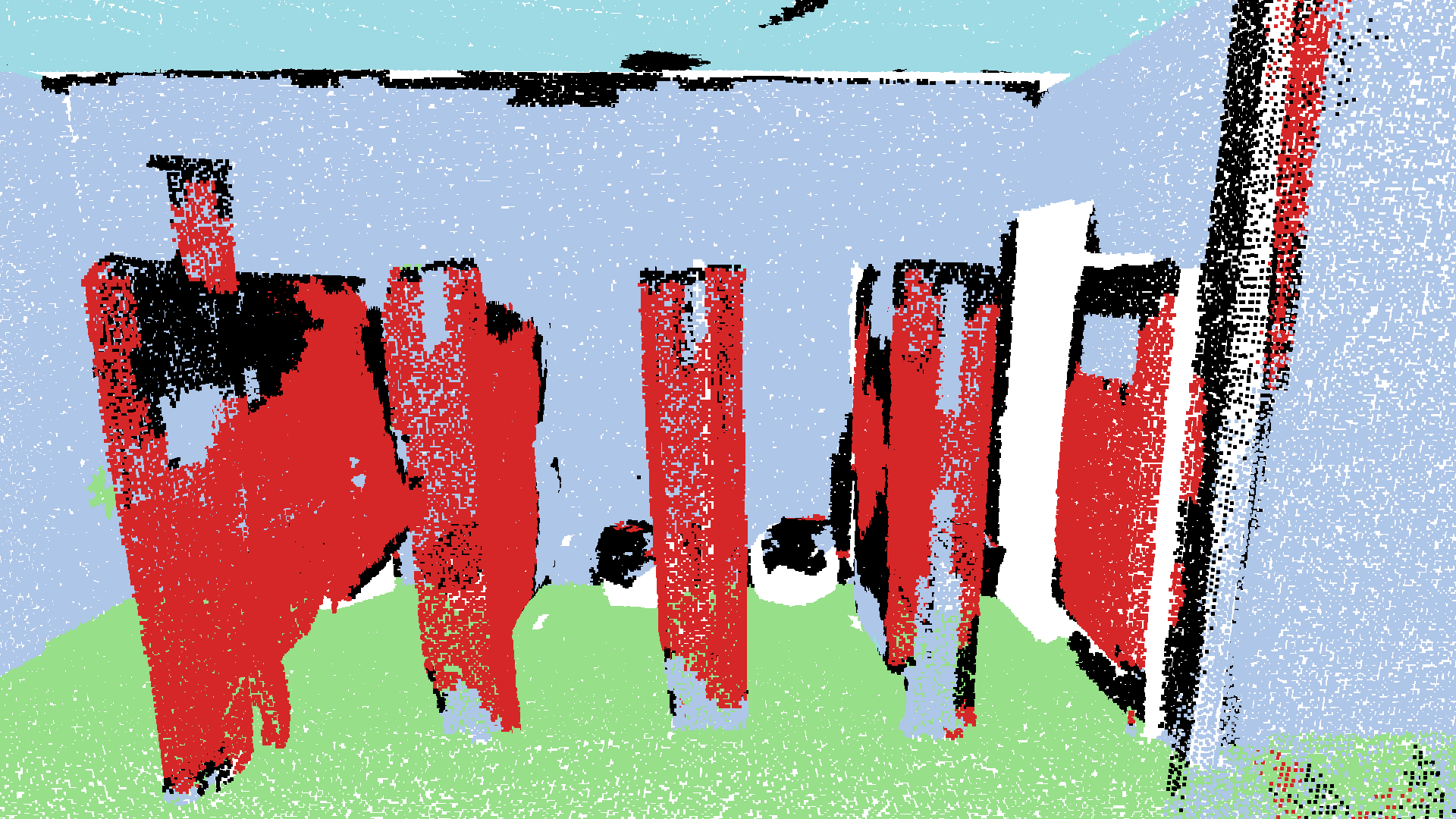}&
            	\includegraphics[width=0.24\linewidth]{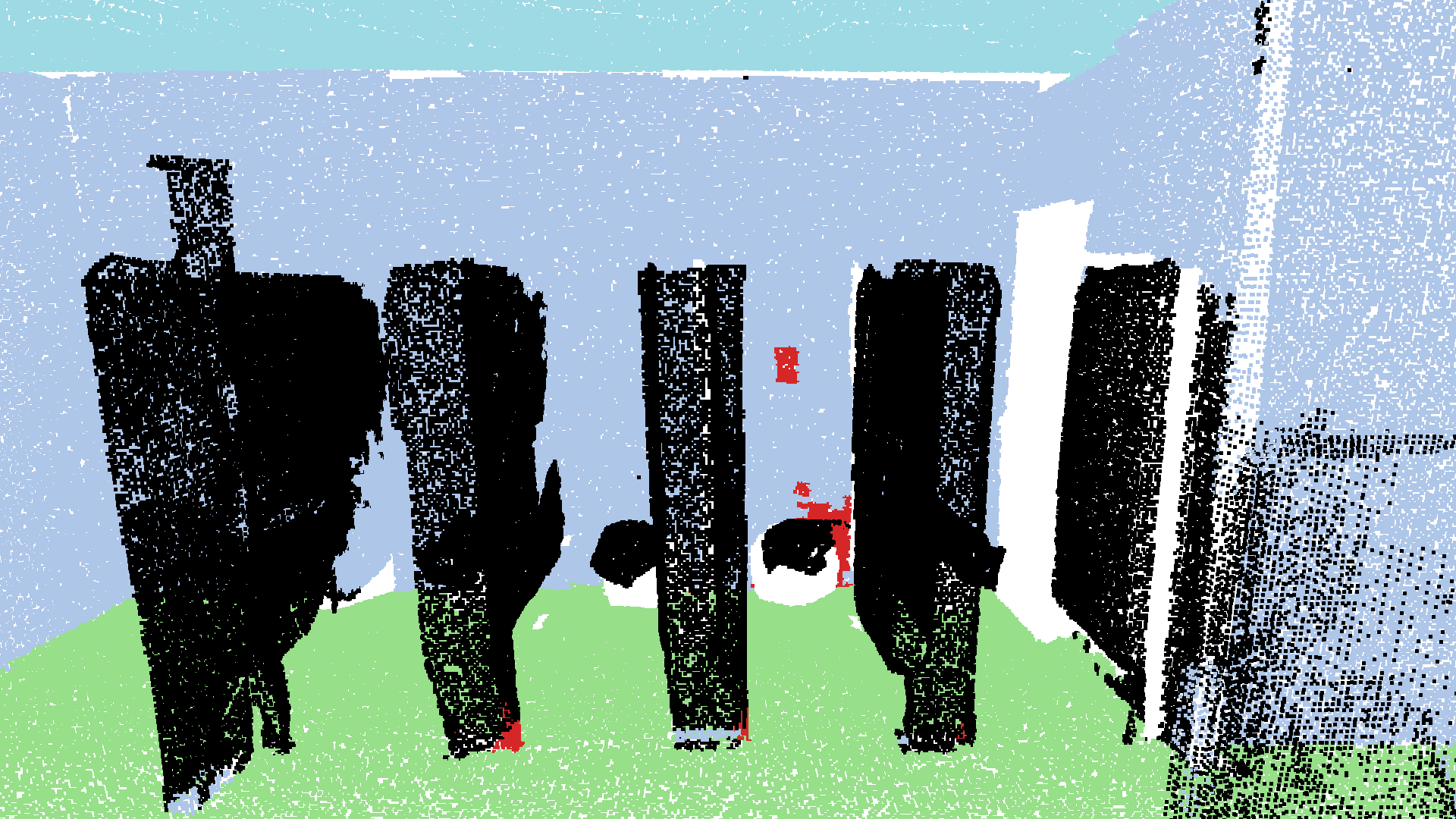}\\
                \multicolumn{4}{c}{\includegraphics[width=0.85\linewidth]{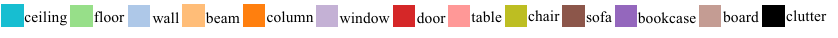}}\\
            \end{tabular}
        }
        \vspace{-2mm}
	\caption{Visualization of semantic segmentation results on the ScanNet and S3DIS training sets.}
	\label{fig:scannet_train}
\end{figure*}

Fig.~\ref{fig:scannet_train} shows additional visualization results of the generated pseudo labels on the ScanNet and S3DIS training sets under scene-level annotation.

\end{document}